\documentclass[10pt,twocolumn,letterpaper]{article}

\usepackage{iccv}              
\usepackage[accsupp]{axessibility}

\newcommand{\framework}{\texttt{BATCLIP}}
\newcommand{\cmark}{\ding{52}}%
\newcommand{\xmark}{\ding{55}}%
\definecolor{light_gray}{gray}{0.9}
\definecolor{iccvblue}{rgb}{0.21,0.49,0.74}
\usepackage[pagebackref,breaklinks,colorlinks,allcolors=iccvblue]{hyperref}
\usepackage{url}
\usepackage{graphicx}
\usepackage{multirow}
\usepackage{dsfont}
\usepackage{amsmath}
\usepackage{amssymb}
\usepackage{booktabs}
\usepackage{color}
\usepackage{colortbl}
\usepackage{xcolor}
\usepackage{enumitem}
\usepackage{multicol}
\usepackage{multirow}
\usepackage{colortbl}
\usepackage{pifont}
\usepackage{makecell}
\usepackage{wrapfig}
\usepackage{subcaption}
\usepackage{graphicx}
\usepackage{subcaption} 
\usepackage{color}
\usepackage{colortbl}
\usepackage{xcolor}
\usepackage{tikz}
\usepackage{arydshln}
\usepackage{pifont}

\def\paperID{12728} 
\def\confName{ICCV}
\def\confYear{2025}

\title{\framework: Bimodal Online Test-Time Adaptation for CLIP}

\author{Sarthak Kumar Maharana\textsuperscript{1}, Baoming Zhang\textsuperscript{1}, Leonid Karlinsky\textsuperscript{2}, Rogerio Feris\textsuperscript{2}, Yunhui Guo\textsuperscript{1}\\
\textsuperscript{1}The University of Texas at Dallas
\textsuperscript{2}MIT-IBM Watson AI Lab \\
\texttt{\{sarthak.maharana, yunhui.guo\}@utdallas.edu}}

\begin{document}
\maketitle

\begin{abstract}
Although open-vocabulary classification models like Contrastive Language Image Pretraining (CLIP) have demonstrated strong zero-shot learning capabilities, their robustness to common image corruptions remains poorly understood. Through extensive experiments, we show that zero-shot CLIP lacks robustness to common image corruptions 
during test-time, necessitating the adaptation of CLIP to unlabeled corrupted images using test-time adaptation (TTA). However, we found that existing TTA methods have severe limitations in adapting CLIP due to their unimodal nature. To address these limitations, we propose \framework, a bimodal \textbf{online} TTA method designed to improve CLIP's robustness to common image corruptions. The key insight of our approach is not only to adapt the visual encoders for improving image features but also to strengthen the alignment between image and text features by promoting a stronger association between the image class prototype, computed using pseudo-labels, and the corresponding text feature. We evaluate our approach on benchmark image corruption datasets and achieve state-of-the-art results in online TTA for CLIP. Furthermore, we evaluate our proposed TTA approach on various domain generalization datasets to demonstrate its generalization capabilities. Our code is available at \url{https://github.com/sarthaxxxxx/BATCLIP}.

\end{abstract}    
\vspace{-0.6cm}
\section{Introduction}
\label{sec:intro}
\vspace{-3pt}

\begin{figure*}[!tb]
\small
\centering
\setlength{\tabcolsep}{.9pt}
\begin{tabular}{c c cc }
\begin{subfigure}[b]{0.245\textwidth}
    \centering
    \includegraphics[width=\linewidth, trim = 0mm 0mm 0mm 0mm, clip]{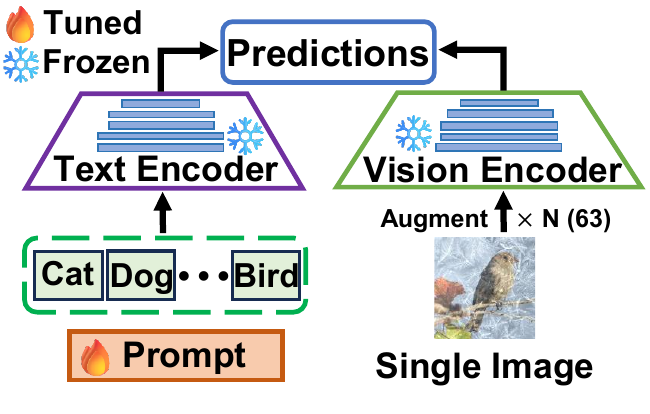}
    \caption{ TPT \cite{shu2022test}}
\end{subfigure} &
\begin{subfigure}[b]{0.245\textwidth}
    \centering
    \includegraphics[width=\linewidth, trim = 0mm 0mm 0mm 1mm, clip]{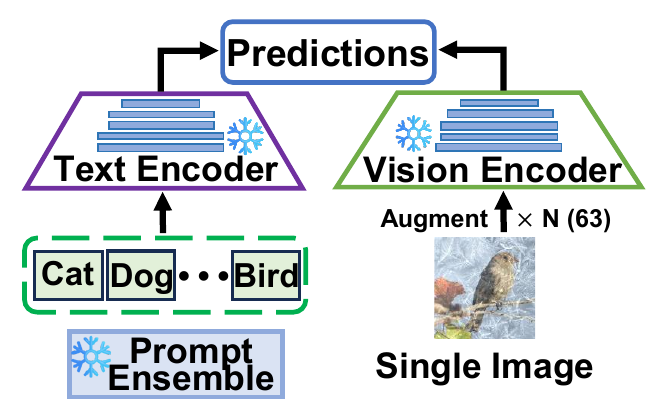}
    \caption{ VTE \cite{dobler2024lost}} 
\end{subfigure} &
\begin{subfigure}[b]{0.245\textwidth}
    \centering
    \includegraphics[width=\linewidth, trim = 0mm 0mm 0mm 0mm, clip]{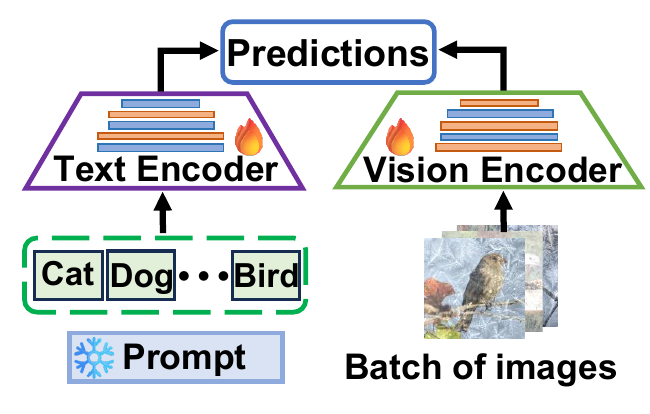}
    \caption{ \framework\ (Ours)}
\end{subfigure} &

\begin{subfigure}[b]{0.245\textwidth}
    \centering
    \includegraphics[width=\linewidth]{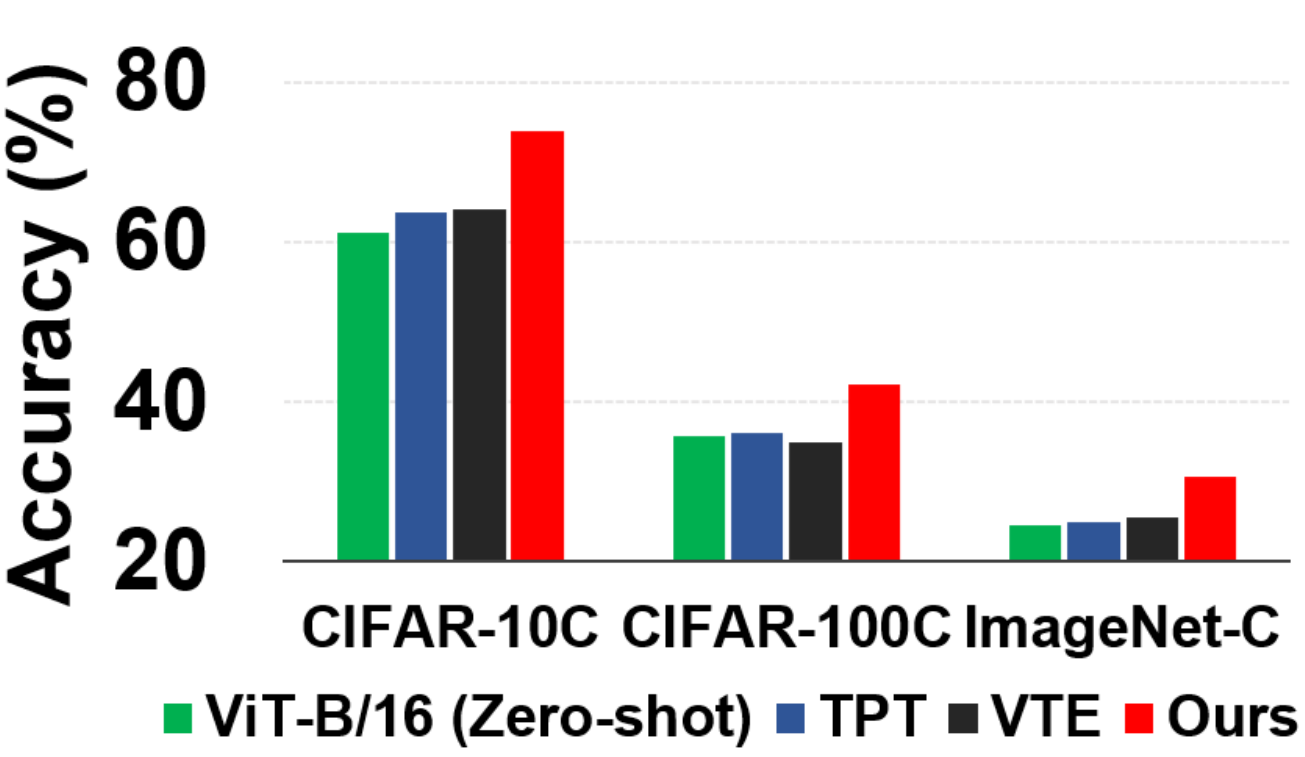}
    \caption{Performance comparison}
    \label{fig:1d}
\end{subfigure} \\
\end{tabular}
\caption{Comparison of \framework\ with other \textbf{online} TTA approaches using CLIP. a) TPT \cite{shu2022test} optimizes text prompts only for a single test image, making it \textit{unimodal}. b) VTE \cite{dobler2024lost} considers an ensemble of prompts without model updates. Both methods consider the generation of multiple augmentations of the test image. c) \framework\ is a \textit{bimodal} approach, that adapts the LayerNorm parameters of the vision and text encoders, maximizing alignment between class prototypes and text features while increasing the inter-class separability of prototypes.}
\vspace{-16pt}
\label{fig:compare}
\end{figure*}

The emergence of large pre-trained vision-language models (VLMs), such as CLIP \cite{radford2021learning}, has led to their widespread adoption in various visual recognition tasks, including segmentation \cite{li2022language, luo2023segclip}, detection \cite{bangalath2022bridging, lin2023gridclip}, classification \cite{zhou2022learning, zhou2022conditional}, and image generation \cite{vinker2022clipasso, ramesh2022hierarchical, rombach2022high}. Thanks to supervision from massive corpora of paired language and image data, VLMs like CLIP demonstrate strong zero-shot capabilities for these downstream tasks.


Despite CLIP's successes in such important applications, its robustness when faced with corrupted images remains largely underexplored. Our motivation stems from the fact that the vision perception system of humans exhibits a level of robustness that real-world vision systems are yet to achieve. For example, models deployed for safety-critical applications like autonomous driving \cite{arnold2019survey}, could face rapid distributional shifts of blurriness, pixel changes, snowy nights, or other weather conditions \cite{sakaridis2021acdc}. In particular, our findings on the zero-shot performance of CLIP with a ResNet-101 \cite{he2016deep} vision backbone reveals that the accuracy on the test set of CIFAR100 \cite{krizhevsky2009learning} with \textit{Gaussian} noise of severity level 5, plummets to 10.79\% from 49\% on the clean set. Similar trends are observed with ViT-B/16, -B/32, and -L/14 \cite{dosovitskiy2020image} as backbones. 
The performance degradation caused by image corruption can have significant consequences in real-world scenarios, particularly in safety-critical applications like self-driving cars. 

One effective approach to improving model performance under distribution shift is test-time adaptation (TTA) \cite{wang2020tent, dobler2023robust}, where a pre-trained model is directly adapted to unlabeled test batches, without access to the source dataset. Moreover, given the abundance of unlabeled data available in the wild, there is a growing need for \emph{online} adaptation to streaming data, with a \textit{single} forward pass, ensuring privacy, real-time performance \cite{mai2022online}, and preventing \textit{catastrophic forgetting} \cite{goodfellow2013empirical} of source knowledge.
Several online TTA methods proposed in the literature have been effective in mitigating domain shifts \cite{wang2020tent, schneider2020improving, rusak2021if, niu2023towards, sun2020test, chen2022contrastive}. In this paper, we use the terms distributions and domains interchangeably.


While there have been few works on using CLIP for \textbf{online} TTA \cite{shu2022test, dobler2024lost, karmanov2024efficient}, they come with certain limitations. For example, test-time prompt tuning (TPT) \cite{shu2022test} tunes the text prompts on the text encoder alone and generates multiple random augmented views, for each test image. The text prompts, initialized to pre-trained values, are optimized by minimizing the marginal entropy of the confident model predictions. The prompts are reset after adaptation to each image. However, such a method is expensive and slow due to performing multiple forward passes through the vision encoder of CLIP, for each image. Also, it relies on hand-crafted prompts for initialization, making it impractical at test-time. Another recent approach, Vision-Text-Space Ensemble (VTE) \cite{dobler2024lost}, uses an ensemble of different prompts as input to CLIP's text encoder while keeping both the encoders frozen. Since the vision encoder is frozen, it struggles to effectively adapt to images with severe noise.


While TPT effectively improves CLIP's test generalization by dynamically tuning the text prompts, it remains primarily a \textit{unimodal} approach. This limits the capacity of a multimodal model like CLIP to fully leverage its multimodal nature for adaptation. Specifically, it prevents the encoders from jointly adjusting their features, resulting in suboptimal alignment between the visual and text modalities after adaptation. For instance, when a test image includes common corruptions, the text prompts adapt, but the vision encoder features remain fixed. As a result, the learned prompts lack awareness of the test image distribution, leading to a less effective adaptation. 


To address these core limitations, we propose, \framework, a \textit{\underline{b}imodal} \underline{a}daptation approach of \underline{CLIP} for \textbf{online} \underline{T}TA specifically, where both the visual and text encoders are jointly adapted by exploiting CLIP's shared feature space of images and text. The overall objective is to achieve a strong alignment between image features and text features to enable more effective multimodal learning and adaptation. The adaptation procedure is two-fold: 1) Inspired by \cite{yuan2023robust, zhao2023tuning} for efficient fine-tuning, we only adapt the norm layers, i.e., \textit{LayerNorm} parameters of both encoders. However, such a model update does not consider the alignment of the encoder features. Therefore, to improve the alignment between class-specific visual and text features, we introduce a projection matching loss that maximizes the projection of the visual class prototypes with their corresponding text features. 2) To learn more discriminative visual features, we enhance the cosine distance between class prototypes, computed using pseudo-labels, to promote a more distinct separation in the image feature space. Our method is general-purpose, with no reliance on multiple prompt templates \cite{osowiechi2024watt, hakim2024clipartt, mishra2024words} or their ensembles, unlike VTE. We leverage batches of test samples for TTA, rather than focusing on single-image test adaptation, which makes our method fast. In the Supplementary, we show several examples of classification results, as a comparison. Figure \ref{fig:1d} shows that with a \textit{bimodal} \textbf{online} test-time adaptation (TTA) setup for CLIP, the proposed approach achieves a significant improvement compared to TPT and VTE.
Our main contributions are as follows:
\begin{itemize}
    \item We start with an in-depth analysis of CLIP's zero-shot performance across different visual backbones on an established benchmark on common image corruptions \cite{hendrycks2019benchmarking}, at test-time, for varying severity levels. We found that while CLIP shows strong performance on clean images, its performance drops notably on corrupted images.
    
    \item To address the \textit{unimodal} limitations highlighted earlier, we propose \framework, a \textit{bimodal} \textbf{online} test-time adaptation method for CLIP encoders, designed to enhance alignment by maximizing the projection of class-wise prototypes onto their corresponding text features. Simultaneously, we increase the cosine distance between class prototypes to encourage learning of more discriminative features, making the test adaptation process more flexible and robust. 
    
    \item We conduct extensive experiments benchmarking the proposed method against established TTA baselines and others using CLIP on CIFAR-10C, CIFAR-100C, and ImageNet-C \cite{hendrycks2019benchmarking}. To generalize our method beyond common corruptions, we evaluate on various domain generalization datasets \cite{gulrajani2020search}-OfficeHome \cite{venkateswara2017deep}, PACS \cite{li2017deeper}, VLCS \cite{fang2013unbiased}, and Terra Incognita \cite{beery2018recognition}. Overall, our approach achieves SOTA performance for CLIP adaptation at test-time on various domain shifts (see Supplementary).


    
\end{itemize}

\vspace{-7pt}

\section{Related Works}
\label{sec:related}
\vspace{-2pt}
\noindent \textbf{Online Test-Time Adaptation (TTA).} The objective of online TTA is to adapt a pre-trained model, with no access to the source data, to incoming batches of unlabelled test data of a specific domain \cite{wang2020tent, sun2020test,  schneider2020improving, rusak2021if, niu2023towards, chen2022contrastive, zhang2022memo, choi2022improving}. 
In this approach, the model is reset to its pre-trained state after adapting to each target domain. As the updates are performed online, TENT \cite{wang2020tent} updates the affine parameters of the normalization layers and minimizes the entropy \cite{shannon1948mathematical} of the model predictions. RoTTA \cite{yuan2023robust} proposes a robust feature normalization method, leveraging a memory bank to track test data distribution and a teacher-student framework for adaptive reweighting and updates. RPL \cite{rusak2021if} argues that self-learning during adaptation via entropy minimization and pseudo-labels is beneficial. They propose the usage of a generalized cross-entropy loss for adaptation. SAR \cite{niu2023towards} filters noisy test samples that cause a performance drop identified from the gradient space. However, none of these works utilize CLIP for TTA.

\noindent \textbf{TTA using CLIP.} Lately, VLMs like CLIP have found extensive applications for \textit{online} TTA \cite{shu2022test, karmanov2024efficient, dobler2024lost, zhao2023test, ma2024swapprompt}. TPT \cite{shu2022test} was the first work to propose prompt tuning using CLIP at test-time in an online manner. However, this method is computationally intensive for generating multiple views, per image. A similar line of work was done in VTE \cite{dobler2024lost} where ensembles are created in the text and vision space without any CLIP parameter update. Ma et al. \cite{ma2024swapprompt} explore the idea of self-supervised contrastive learning for prompt learning. Another brewing line of TTA work focuses on realistic TTA. Stat$\mathit{A}$ \cite{zanella2025realistic} proposes a regularization anchor term, for zero-shot CLIP, handling variable number of effective classes within a test batch, often overlooked in existing TTA methods \cite{martin2024transductive, kalantidis2024label, zanella2025boosting} on other visual benchmarks \cite{zhou2022learning}.

Another set of TTA works using CLIP \cite{osowiechi2024watt, hakim2024clipartt, mishra2024words} leverage information from multiple prompt templates. WATT \cite{osowiechi2024watt} adapts CLIP's vision encoder, weight-averaging the adapted weights from multiple prompt templates, for each test batch, over several optimization steps. Similarly, CLIPArTT \cite{hakim2024clipartt} enhances text supervision from the support of pseudo-labels. Mishra et al. \cite{mishra2024words} pose pseudo-labeling as an Optimal Transport \cite{cuturi2013sinkhorn} problem by distilling knowledge from multiple text prototypes, for several iterations. Please note that the TTA setups here largely differ from our \textbf{online} TTA setup \cite{wang2020tent}. 


While promising, these approaches have key limitations. Firstly, multiple gradient descent steps on a test batch can result in large parameter shifts, leading to the loss of CLIP’s source knowledge i.e., \textit{catastrophic forgetting} \cite{goodfellow2013empirical}. No regularization strategies have been proposed to restrict the loss of CLIP's pretrained knowledge. Secondly, these methods are not favorable for real-time deployment where fast and on-the-fly test-adaptation must meet privacy and memory constraints. Thirdly, relying on multiple prompt templates, at test-time, is impractical. We discuss this in Section \ref{analysis}.

\vspace{-8pt}
\section{Zero-shot performance analysis of CLIP to common image corruptions}
\label{analysis}

While CLIP generalizes well to new concepts across vision-language modalities \cite{chen2021pre, han2021pre}, its performance under image corruptions is less explored. This section evaluates CLIP’s zero-shot capabilities in real-world scenarios with domain shifts caused by common corruptions, focusing on two primary areas: \textit{zero-shot performance and the need for adaptation to address domain shifts effectively.}

\noindent \textbf{Vision Backbones.} We evaluate the robustness with a ResNet-101 (RN101) vision backbone \cite{he2016deep} and three ViT backbones (ViT-B/16, ViT-B/32, and ViT-L/14) \cite{dosovitskiy2020image}. The models are tested on their zero-shot classification performance.

\noindent \textbf{Datasets.} For all the experiments in this paper, we utilize the CIFAR-10C, CIFAR-100C, and ImageNet-C \cite{hendrycks2019benchmarking} datasets, which are standard benchmark datasets to evaluate the robustness of vision models \cite{hendrycks2019benchmarking}. Each dataset contains 15 distinct corruption types as tasks (e.g., Gaussian noise, Shot noise, \ldots). Each corruption is applied at 5 different severity levels to the test sets of CIFAR10, CIFAR100 \cite{krizhevsky2009learning}, and ImageNet \cite{deng2009imagenet}, acting as source test sets, and allowing us to systematically evaluate the model's performance under increasing degrees of image degradation. 


\noindent \textbf{Online TTA Problem Setup.} Each corruption type of a certain severity level, posed as a task $\mathcal{T}_i$ with $\mathcal{B}$ test batches, is sequentially presented to CLIP's vision encoder for model predictions, with each test batch being revealed one at a time for a \textit{single} forward pass. Let a batch of images from task $\mathcal{T}_i$, at time step \textit{t}, be denoted as $x_i^t$. For the prompt template, unless explicitly mentioned, we always use the generic ``{\fontfamily{cmss}\selectfont a photo of a $<$CLS$>$.}" to generate \textit{C} text representations ($\mathcal{Z}$ = \{\textit{z}$_c$\}$_{c=1}^C$), where \textit{C} is the total number of classes. Let $f_{vis}$ and $f_{txt}$ denote CLIP's vision and text encoder, respectively. The visual feature of the $k^{th}$ image in batch $x_i^t$ is $v_{k,i}^t$ = $f_{vis}(x_{k,i}^t)$. The likelihood of it belonging to class \textit{c} is, 
\begin{equation}\label{eq:likelihood}
\begin{aligned}
    p(y=c|x_{k,i}^t) &= \frac{\text{exp}(\text{sim}(v_{k,i}^t, z_c)/\tau)}{\sum_{j}{}\text{exp}(\text{sim}(v_{k,i}^t, z_j)/\tau)} \\
    \text{sim}(v, z) &= \frac{v^T \cdot z}{||v||_2 \cdot ||z||_2}
\end{aligned}
\end{equation}
where $\text{sim}(\cdot)$ is the cosine similarity and $\tau$ is the softmax temperature from CLIP's pre-training stage. The text features $\mathcal{Z}$, in zero-shot evaluation, are always pre-computed. We also draw a comparison with CLIP performance on respective source test sets and follow this implementation \footnote{\href{https://github.com/LAION-AI/CLIP_benchmark}{https://github.com/LAION-AI/CLIP\_benchmark}}. Throughout this paper, the same TTA problem setup is used, unless mentioned otherwise. This TTA setup differs from the TTA setups of \cite{osowiechi2024watt, hakim2024clipartt, mishra2024words}.

\begin{figure}[!htb]
\Large
\centering
\setlength{\tabcolsep}{1pt}
\begin{tabular}{c c}
\includegraphics[width=\columnwidth, trim = 0mm 0mm 0mm 0mm, clip]{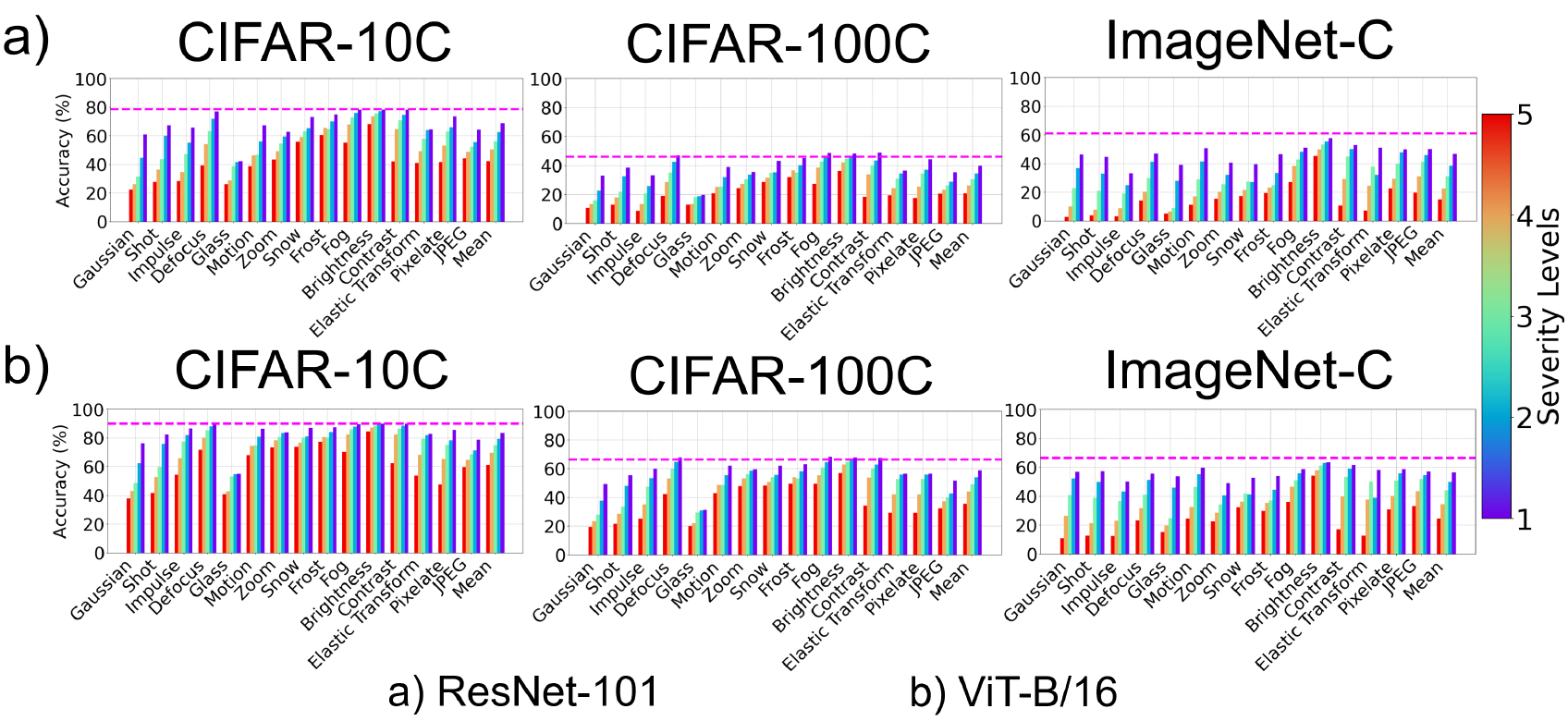}
\end{tabular}
\vspace{-8pt}
\caption{Task-wise mean accuracy (\%) of zero-shot CLIP across corruption severity levels. [Top]: ResNet-101 backbone. [Bottom]: ViT-B/16 backbone. \textcolor{magenta}{\textbf{Dashed lines}} show zero-shot CLIP performance on corresponding source test sets (clean).}
\label{fig:severity}
\vspace{-15pt}
\end{figure}

\vspace{-5pt}

\subsection{Sensitivity of CLIP to image corruption severity} 
\vspace{-3pt}
We analyze CLIP’s zero-shot performance by progressively increasing corruption severity for RN101 and ViT-B/16 vision backbones. We illustrate mean accuracy and overall performance in Figure \ref{fig:severity}. Despite CLIP’s robust multimodal feature space, accuracy drops significantly with an increase in corruption severity, regardless of the backbone. Specifically, for CIFAR-10C with a ViT-B/16 backbone, we observe accuracy as low as 37.92\%, at a severity level of 5, for \textit{Gaussian} noise. Similarly, for CIFAR-100C and ImageNet-C, the mean accuracy rates are as low as 35.79\% and 24.51\%  at a severity level of 5, respectively. Results on additional vision backbones are in the Supplementary.

\noindent \textbf{Analysis.} CLIP's zero-shot classification accuracy varies across models and datasets. Comparing the results to source test sets, for CIFAR10, RN101 achieves 78.8\%, improving to 90.1\% with ViT-B/16. On CIFAR100, accuracies are 46.1\% and 66.6\%, respectively. For ImageNet, RN101 scores 61.2\%, while ViT-B/16 reaches 67.7\%. 
More importantly, the key takeaway from Figure  \ref{fig:severity} is --- even a slight increase in severity to level 1, for a majority of the corruption tasks, leads to a noticeable drop in accuracy. One plausible explanation for CLIP's subpar performance is that the parameters of $f_{vis}$ were not optimized for such corruptions during pre-training. In zero-shot classification, the visual features from a given domain may lack the robustness and richness necessary to align well with their corresponding text features. This results in lower likelihoods and thus, higher misclassification rates. 

\begin{figure}[t!]
\small
\centering
\setlength{\tabcolsep}{.1pt}
\begin{tabular}{c c c}
\begin{subfigure}[b]{0.16\textwidth}
    \centering
    \includegraphics[width=\linewidth, trim = 0mm 5mm 0mm 0mm, clip]{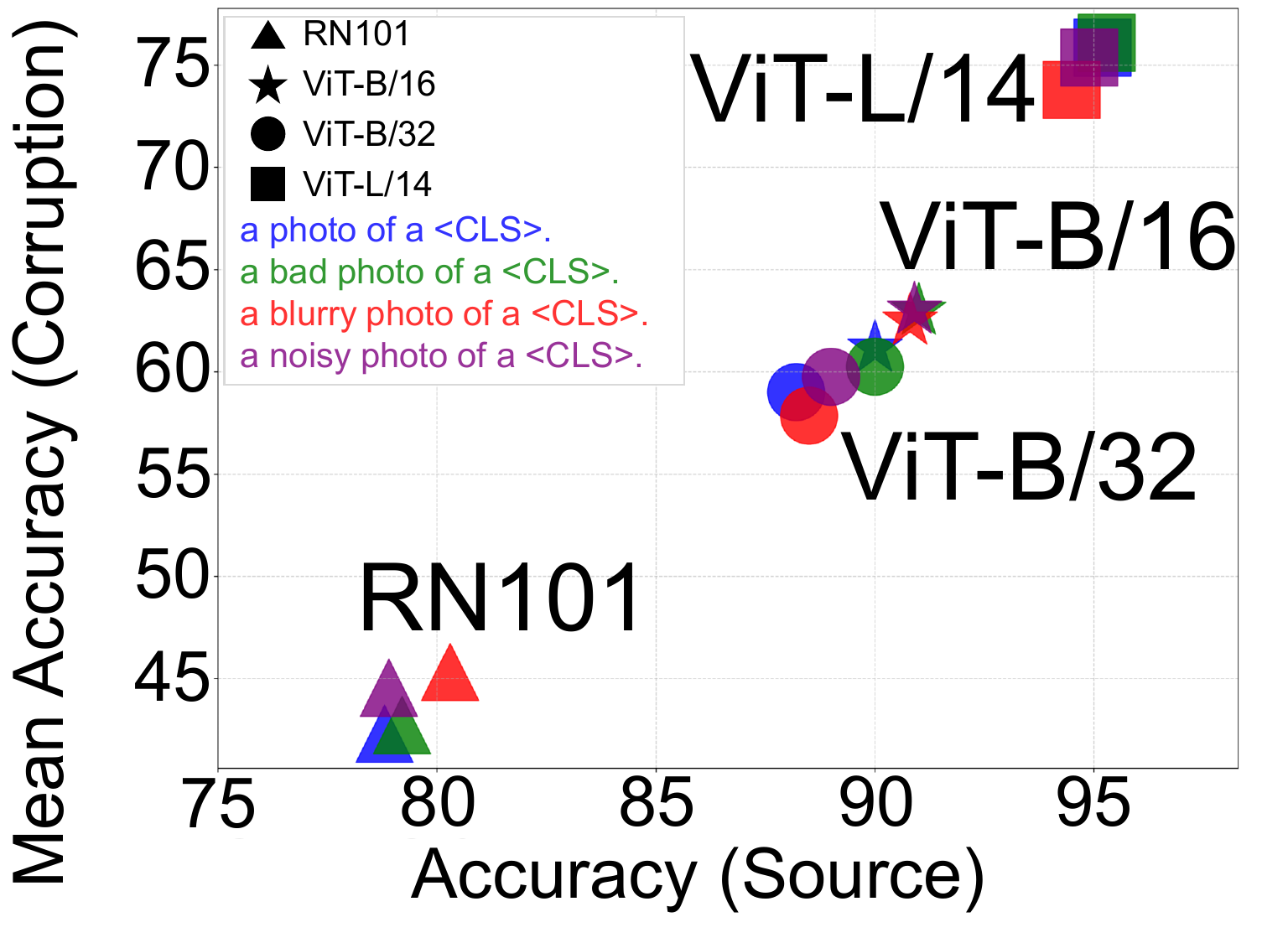}
    \caption{CIFAR-10C}
\end{subfigure} &
\begin{subfigure}[b]{0.16\textwidth}
    \centering
    \includegraphics[width=\linewidth, trim = 0mm 0mm 0mm 1mm, clip]{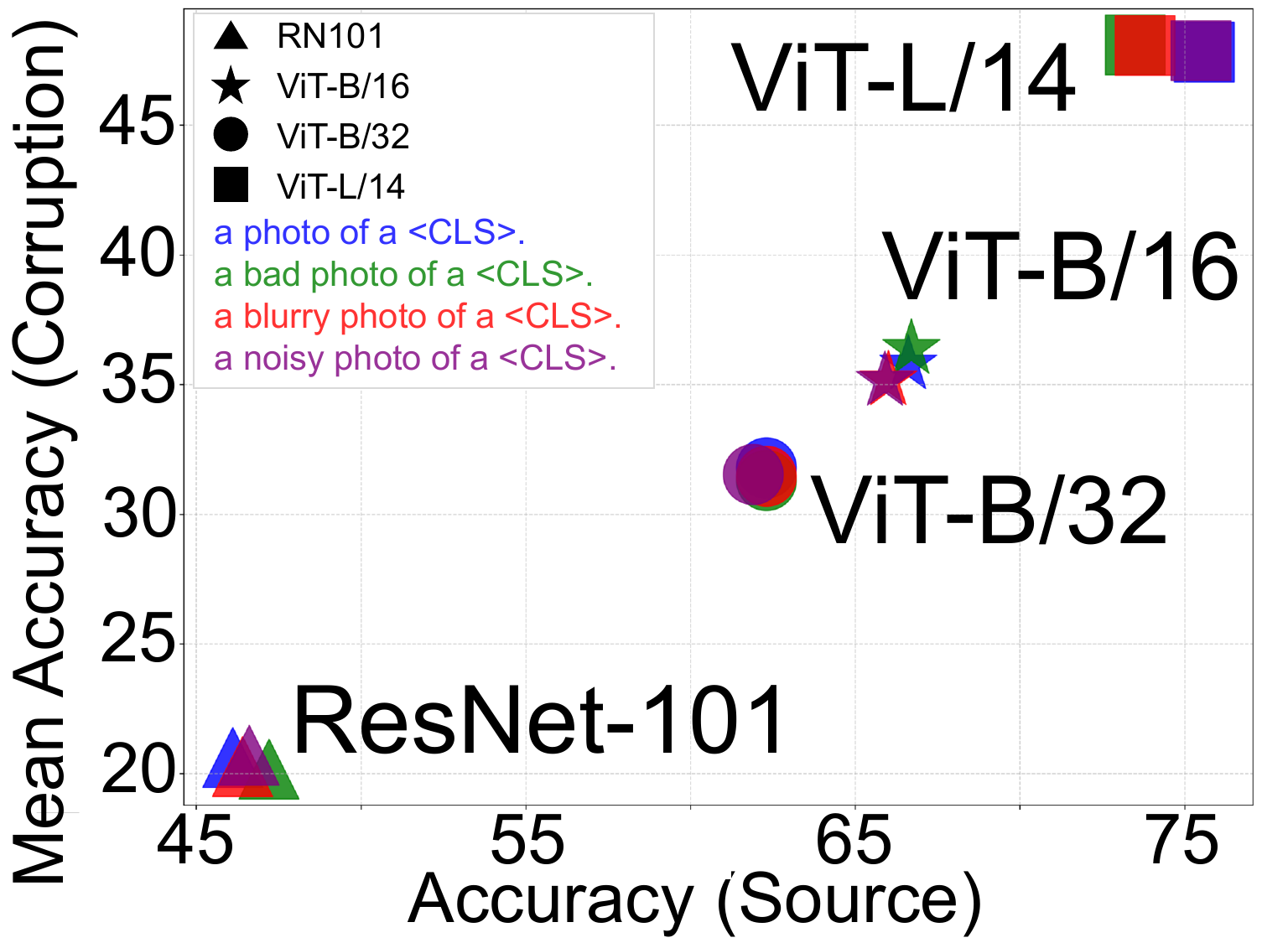}
    \caption{CIFAR-100C} 
\end{subfigure} &
\begin{subfigure}[b]{0.16\textwidth}
    \centering
    \includegraphics[width=\linewidth, trim = 0mm 5mm 0mm 0mm, clip]{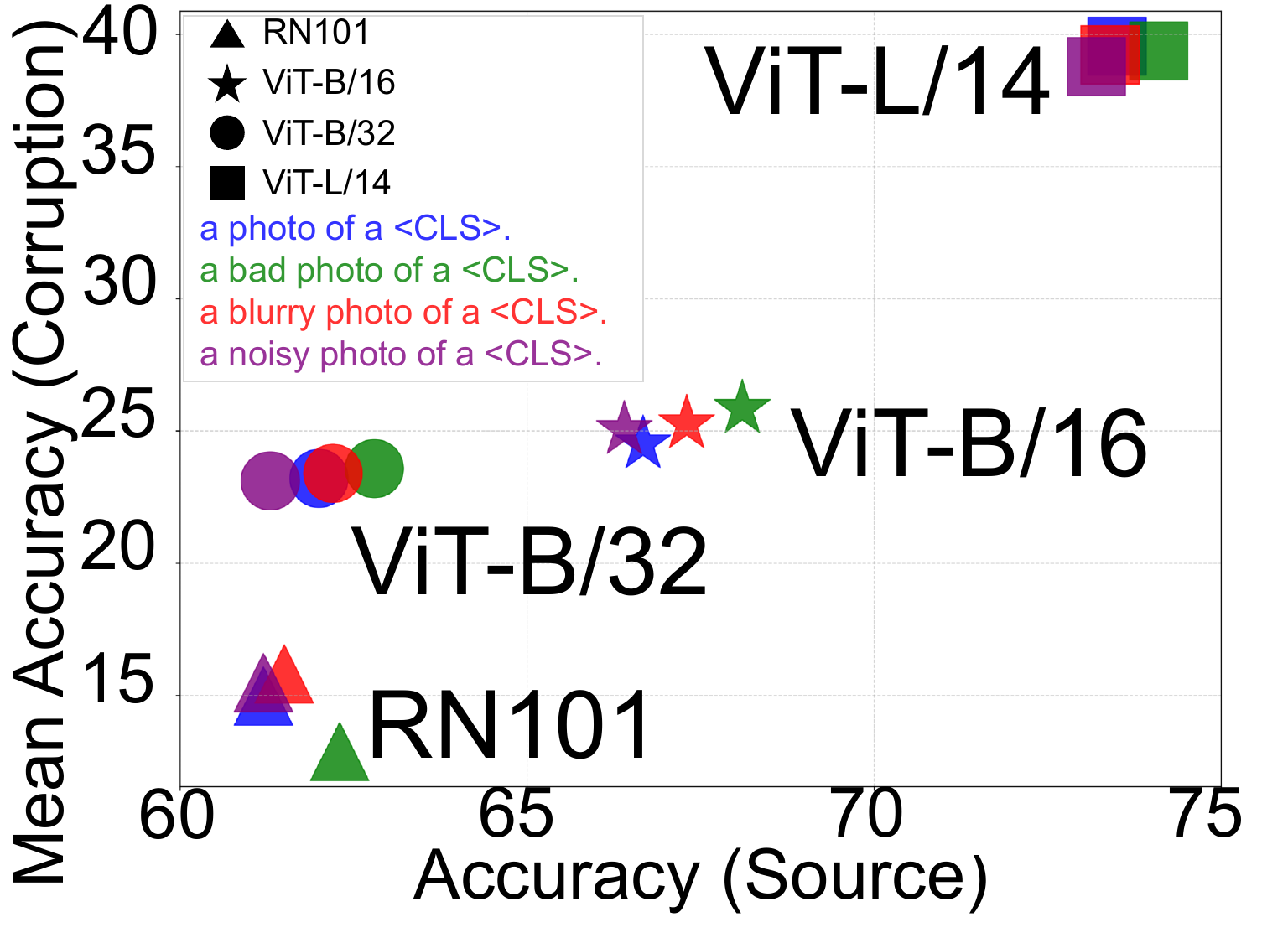}
    \caption{ImageNet-C}
\end{subfigure} \\
\end{tabular}
\vspace{-10pt}
\caption{Mean classification accuracy (in \%) across all the corruption types vs. accuracy on corresponding source test sets, evaluated using various prompt templates and across vision backbones.}
\vspace{-16pt}
\label{fig:prompts}
\end{figure}

\vspace{-2pt}
\subsection{Sensitivity of CLIP to prompt templates}
\vspace{-3pt}

In this analysis, we evaluate the impact of prompt engineering by providing ``relevant" prompt templates to $f_{txt}$ for TTA. Each prompt adds context to help CLIP extract more relevant text features. We report the mean classification accuracy (in \%) across RN101, ViT-B/16, ViT-B/32, and ViT-L/14 backbones at an image corruption severity level of 5 for all datasets, with results presented in Figure \ref{fig:prompts}. For each dataset, the x-axis is the source accuracy and y-axis is the mean accuracy, across 15 tasks. 

\noindent \textbf{Analysis.}  It is interesting to observe that, irrespective of the backbones used, we do not see any drastic changes in the mean accuracy, for different ``relevant" prompt templates. However, the major concern arises in the performance gap of each model and the zero-shot CLIP performance on the corresponding source test set, for the same prompt template. This discrepancy highlights the limited robustness of CLIP's text encoder $f_{txt}$ to prompt selection in the context of image corruption. A key reason for this is that, \textit{despite the use of ``relevant" prompts, the text and visual features remain largely independent and unaware of one another.}

As expected, RN101 performs worse than the ViT-based backbones, primarily due to its lack of global attention-based modeling inherent to transformers \cite{vaswani2017attention}. Therefore, for the remainder of the experiments in this paper, we focus on ViT-based backbones, specifically ViT-B/16. 

\noindent \textbf{Unsuitability of prompt template selection at test-time.} Additionally, at test-time, it is impractical to perform prompt engineering or optimize prompt vectors since 1) Choosing different prompt templates for generating text features is extremely tedious and time-consuming. As discussed in Section \ref{sec:related}, CLIP TTA works \cite{osowiechi2024watt, hakim2024clipartt, mishra2024words} use various prompt templates, making it unrealistic. 2) In real-time deployment involving prompt-tuning, such prompts cannot quickly estimate the distribution of incoming test batches. TPT \cite{shu2022test} optimizes pre-trained text prompts for each test image, turning out to be suboptimal since the prompts are optimized ignorant of the distribution of the test image. 
\vspace{-0.2cm}
\section{Proposed Methodology}
\label{proposed}

\begin{figure*}[!tb]
\small
\centering
\setlength{\tabcolsep}{5pt}
\begin{tabular}{c}
\includegraphics[trim={0cm 0 0cm 0},clip, width=0.8\textwidth]{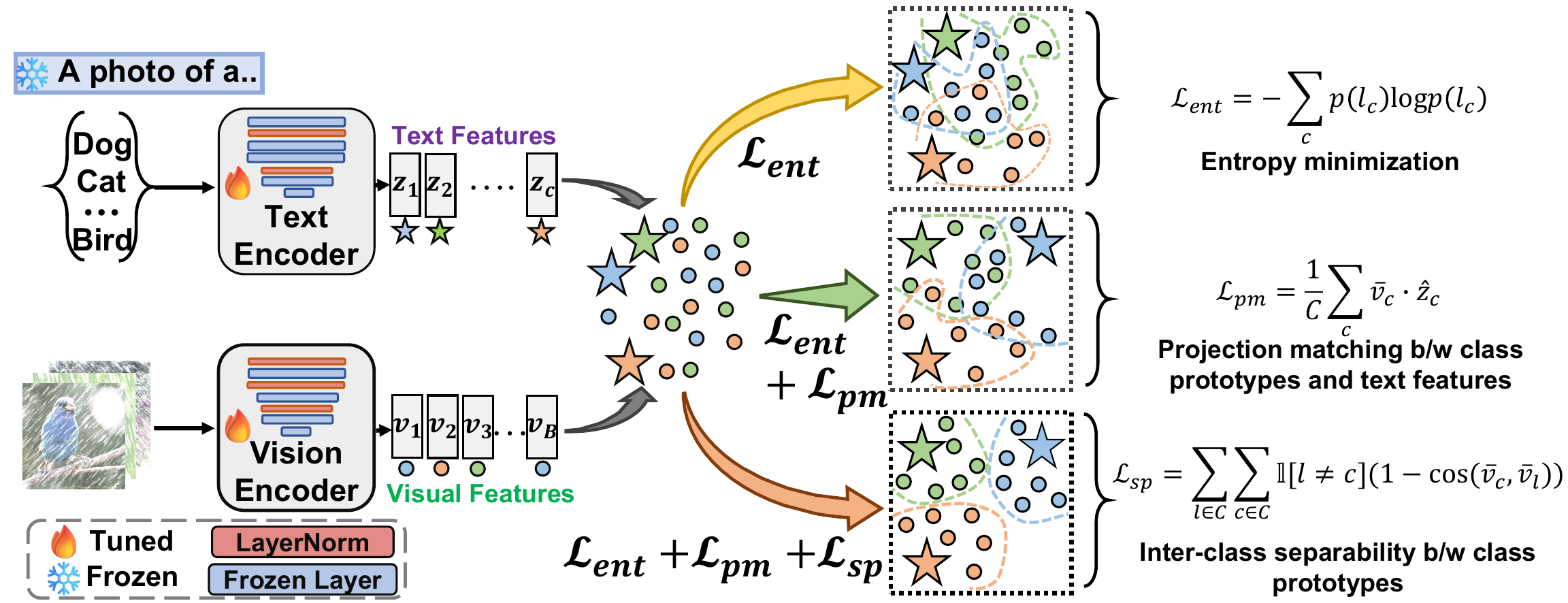}
\end{tabular}
\caption{
Illustration of our proposed approach: \framework\ not only adapts the visual encoder for highly discriminative image features but also promotes a strong alignment between image and text features by adapting the text encoder too, leading to improved performance following test-time adaptation. We adapt only the LayerNorm parameters of CLIP encoders.}
\label{fig:framework}
\vspace{-10pt}
\end{figure*}

\vspace{-5pt}
The comprehensive analysis in Section \ref{analysis} reveals that the zero-shot vision encoder $f_{vis}$ of CLIP is very sensitive to image corruption with increasing severity. Similarly, the performance of the text encoder $f_{txt}$ is invariant to the different text prompt templates and is also impractical to tune at test-time. Therefore, for effective adaptation, both the image and text encoders of CLIP need to be \textit{adapted} to the incoming domain of test batches to focus on \textit{achieving strong image-text alignment and feature separability}. We illustrate $\framework$ in Figure \ref{fig:framework}.

\noindent \textbf{From unimodal adaptation to bimodal adaptation.} The goal of TTA is to enhance a model's performance in the current domain to ensure accurate predictions. To handle the complexities of CLIP adaptation to a specific domain of image corruption, we dissect our analysis of each encoder's adaptation. Consider the \textit{unimodal} adaptation of the vision encoder $f_{vis}$ via entropy minimization \cite{wang2020tent}. While the image features are adjusted for a specific test batch and domain, the text features remain fixed, still being optimized for the data from CLIP pre-training, as seen in Figure \ref{fig:framework} (top row). Likewise, if only the text encoder $f_{txt}$ is updated in this manner, the generated text features may not align properly with the distribution of the incoming data, potentially causing misalignment between the image and text features, missing in the ideas of TPT \cite{shu2022test} and VTE \cite{dobler2024lost}.

To benefit from the feature space and learn richer representations across both modalities, we propose adapting both $f_{vis}$ and $f_{txt}$ to a domain, enabling an input-aware knowledge transfer between the encoders and enhancing domain-specific adaptation. While the encoders can be updated based on entropy minimization, they come with inherent limitations. Though entropy models the prediction uncertainty, it does not guarantee increasing the likelihood of alignment between the image and text features. Additionally, due to the image corruption, for images within a test batch, there could be a possible overlap of visual features belonging to different classes \cite{kurakin2016adversarial}. The robustness of CLIP would be challenged in such a case. To address these limitations, we propose two loss components that facilitate a more effective \textit{bimodal} adaptation of CLIP to new domains at test-time. The losses focus on maximizing alignment between the visual and text features while increasing separation between the visual features to learn good decision boundaries.

\noindent \textbf{Projection matching between the visual and text features.} We propose learning visual and text features that are domain-specific and mutually aware by \textit{jointly updating the encoders}, as in Figure \ref{fig:framework} (middle row), opposed to a \textit{unimodal}. From Eq. \ref{eq:likelihood}, the visual feature $v_{k,i}^t$ of the $k^{th}$ image in batch $x_i^t$ needs to have a high similarity with the text feature $z_c$ of class \textit{c} for good alignment. To quantify this similarity, an approach is to project the visual feature onto the text feature \textit{i.e.,} compute the scalar projection --- $v_{k,i}^t \cdot \hat{z}_c$, where $\hat{z}_c$ is the normalized text feature of $\hat{z}_c$ \textit{i.e.,} $\hat{z}_c = \frac{z_c}{||z_c||_2}$. Geometrically speaking, maximizing this projection leads to more similarity between the corresponding features and hence, a better alignment. However, computing such a projection could have limitations. Within the test batch, due to the image corruption, the pseudo-labels could be wrong/noisy. Instead, we propose modeling the projection of the class prototype with its corresponding text feature. A prototype is useful, in such a scenario, because it encompasses the entire class distribution without relying on individual visual features \cite{snell2017prototypical}. In particular, we compute a class prototype as,
\begin{equation}\label{eq:pro}
\begin{aligned}
    \Bar{v}_c &= \frac{1}{\sum_{k=1}^{\mathcal{B}} \mathds{1}[\hat y = c]}\sum_{k=1}^{\mathcal{B}} \mathds{1}[\hat y = c] v_{k,i}^t \\
    \hat y &= \underset{c}{\arg\!\max} \hspace{0.1cm} p(y|x_i^t)
\end{aligned}
\end{equation}

i.e., \textit{for the current test batch}, we compute the mean feature of all the support visual features constituting a class \textit{c}, based on the computed pseudo-labels (Eqs. \ref{eq:likelihood} and \ref{eq:pro}), where $\hat y$ refers to the pseudo-labels computed via Eq. \ref{eq:likelihood} and $\Bar{v}_c$ is the class prototype of class \textit{c}. Based on the class prototype, the projection matching loss is, 
\begin{equation}\label{eq:l_pm}
    \mathcal{L}_{pm} = \frac{1}{C}\sum_{c}\Bar{v}_c \cdot \hat z_c
\end{equation}
Eq. \ref{eq:l_pm} encourages a class prototype to have a larger projection on its corresponding text feature. In this way, during the adaptation of CLIP to a certain domain, both encoders learn to generate richer visual and text features with maximum alignment by maximizing Eq. \ref{eq:l_pm}. Figure \ref{fig:framework} (middle row) shows such an illustration.

\noindent \textbf{Inter-class separability between class prototypes.} The projection matching loss introduced in Eq. \ref{eq:l_pm} encourages the $f_{vis}$ and $f_{txt}$ encoders to produce domain-specific, well-aligned, and input-aware features via jointly updating the encoders. However, when TTA occurs at a batch level with image corruption, visual features within the batch could overlap, leading CLIP to poorly differentiate between classes. This ultimately hinders effective class separation. Hence, with a desire to obtain separation between visual features from different classes, as illustrated in Figure \ref{fig:framework} (last row), we propose maximizing the distance between the class prototypes. Since the visual and text features align across modalities via Eq. \ref{eq:l_pm}, class separation, in addition, is needed for good generalization, robustness, and adaptation. Therefore, we additionally incorporate a loss to increase the cosine distance between the class prototypes,
\begin{equation}
    \mathcal{L}_{sp} = \sum_{l \in C}\sum_{c \in C}\mathds{1}[l \neq c] (1 - \text{cos}(\Bar{v}_c, \Bar{v}_l))
\end{equation}

\noindent \textbf{Optimization.} TENT \cite{wang2020tent} optimizes the output logits by minimizing the entropy \cite{shannon1948mathematical}. When applied to CLIP, the entropy, with output logits \textit{l}, is defined as,  
\begin{equation}
    \mathcal{L}_{ent} = -\sum_{c}^{}p(l_c)\log p(l_c)
\end{equation}
where $p(l_c)$ is the likelihood for class \textit{c} that is computed via Eq. \ref{eq:likelihood}. The overall optimization objective for our approach is as, 
\begin{equation}\label{eq:final_loss}
    \underset{\phi_v, \phi_t}{\arg\!\min} (\mathcal{L}_{ent} - \mathcal{L}_{pm} - \mathcal{L}_{sp})
\end{equation}
where $\phi_v$ and $\phi_t$ refer to the LayerNorm parameters of $f_{vis}$ and $f_{txt}$, respectively. 

\noindent \textbf{Motivation of choice of parameters.} Through thorough experimentation, Zhao et al. \cite{zhao2023tuning} demonstrate that fine-tuning \textit{LayerNorm} (LN) parameters in attention-based models leads to yielding strong results on multimodal tasks. This also brings a reduction in trainable parameters. Updating normalization layers has also been deeply studied in tasks involving domain shifts \cite{li2016revisiting}. Motivated by these works, we update the \textit{LayerNorm} parameters of both CLIP encoders. This constitutes updating $\sim0.044$\% of all parameters. For every new task/corruption, we reset the model parameters following TENT \cite{wang2020tent} i.e., we perform a single domain TTA. In summary, our \textit{bimodal} \textbf{online} test-time adaptation approach jointly updates the LayerNorm parameters of the CLIP encoders that are optimized synergistically through loss components aware of the input domain, leading to a more robust multimodal learning process.


\vspace{-5pt}
\section{Experiments and Results}
\label{exps}

\vspace{-3pt}

\noindent \textbf{Baselines.} We compare our approach with zero-shot CLIP (\textbf{ZS}), state-of-the-art online TTA methods adopted for CLIP following \cite{dobler2024lost} i.e., \textbf{TENT} \cite{wang2020tent}, \textbf{RoTTA} \cite{yuan2023robust}, \textbf{RPL} \cite{rusak2021if}, and \textbf{SAR} \cite{niu2023towards}. We update only the vision encoder $f_{vis}$ as in the baselines. We benchmark against other \underline{online} TTA methods using CLIP - \textbf{TPT} \cite{shu2022test} and \textbf{VTE} \cite{dobler2024lost}. Based on the discussion in Section \ref{sec:related}, we adapt the two variants of WATT \cite{osowiechi2024watt}, for our setup, to \textbf{WATT-P*} and \textbf{WATT-S*}, while maintaining method consistency. For a ``fairer" comparison, on each test batch, the model is adapted once using each of the suggested 8 prompt templates. Note that this setup is \textbf{not online} yet. In addition, CLIP parameters are reset after every task. We report the results on CIFAR-10C and CIFAR-100C only, based on their suggested settings. On ImageNet-C, we also report results against online \textbf{Stat$\textit{A}$} \cite{zanella2025realistic}, based on their chosen hyperparameters. The model predictions can then be computed as in Eq. \ref{eq:likelihood}. We mention the details of each TTA method in the Supplementary.


\noindent \textbf{Implementation Details.} We query the text encoder $f_{txt}$ with a general prompt template ``{\fontfamily{cmss}\selectfont a photo of a $<$CLS$>$.}" for all the datasets, as motivated earlier. We optimize both the vision encoder $f_{vis}$ and text encoder $f_{txt}$. For CIFAR-10C, we use an AdamW optimizer at a learning rate of $10^{-3}$. Similarly, for CIFAR-100C and ImageNet-C, Adam and AdamW optimizers are respectively used, at a fixed learning rate of 5$\times$$10^{-4}$, with the model being reset after each task. For fairness to all the baselines, the batch sizes are set to 200, 200, and 64 for the datasets, following various TTA benchmarks, at a corruption severity level of 5 for each task. For WATT-P* and WATT-S*, we use the same templates and learning rates as mentioned in \cite{osowiechi2024watt}. For Stat$\textit{A}$, the class correlation in a batch is controlled by a Dirichlet distribution ($\gamma$). We report results for $\gamma$ = 0.1 (low correlation) and -1 (sequential). All experiments are run on a single NVIDIA RTX A5000 GPU using ViT-B/16 as the visual backbone. Results on ViT-B/32 are reported in the Supplementary.

\begin{table*}[t]
    \vspace{-10pt}
    \small
    \centering
    \resizebox{\textwidth}{!}{
    \begin{tabular}{cc|c|ccccccccccccccc|c} \\ \toprule
    & Method & Venue & \rotatebox[origin=c]{70}{Gaussian} & \rotatebox[origin=c]{70}{Shot} & \rotatebox[origin=c]{70}{Impulse} & \rotatebox[origin=c]{70}{Defocus} & \rotatebox[origin=c]{70}{Glass} & \rotatebox[origin=c]{70}{Motion} & \rotatebox[origin=c]{70}{Zoom} & \rotatebox[origin=c]{70}{Snow} & \rotatebox[origin=c]{70}{Frost} & \rotatebox[origin=c]{70}{Fog} & \rotatebox[origin=c]{70}{Brightness} & \rotatebox[origin=c]{70}{Contrast} & \rotatebox[origin=c]{70}{Elastic} & \rotatebox[origin=c]{70}{Pixelate} & \rotatebox[origin=c]{70}{JPEG} & Mean  \\ \midrule
    \multirow{9}{*}{\rotatebox[origin=c]{90}{\LARGE{CIFAR-10C}}}
    & \begin{tabular}[c]{@{}c@{}}ZS\end{tabular} &
  ICLR'21 & 37.92 & 41.7 & 54.42 & 71.75 & 40.89 & 67.93 & 73.62 & 73.89 & 77.35 & 70.22 & 84.45 & 62.36 & 53.81 & 47.65 & 59.43 & 61.16 \\ 
& \begin{tabular}[c]{@{}c@{}}TENT\end{tabular} & ICLR'21 &
  15.49 & 18.28 & 38.12 & 81.59 & 21.73 & 76.32 & 82.35 & \textbf{84.62} & 82.19 & 80.60 & \textbf{91.83} & 80.55 & 63.52 & 58.57 & 54.71 & 62.03 \\
& \begin{tabular}[c]{@{}c@{}}RoTTA\end{tabular} & CVPR'23 &
  39.17 & 42.90 & 55.41 & 72.18 & 41.32 & 68.02 & 74.01 & 74.38 & 78.01 & 70.80 & 84.80 & 63.19 & 54.62 & 49.32 & 60.15 & 61.89 \\ 
& \begin{tabular}[c]{@{}c@{}}RPL\end{tabular} & arXiv & 
15.47 & 17.43 & 40.73 & \textbf{81.76} & 20.08 & 69.89 & \textbf{82.93} & 84.43 & 83.19 & 81.84 & 91.80 & 79.42 & 64.89 & 54.07 & 54.90 & 61.52 \\ 
& \begin{tabular}[c]{@{}c@{}}SAR\end{tabular} & ICML'22 &
  47.98 & 53.60 & 60.56 & 74.30 & 47.56 & 73.15 & 76.43 & 77.91 & 79.88 & 75.66 & 86.79 & 71.62 & 58.34 & 62.03 & 64.71 & 67.37 \\ \cdashline{2-19}
& \begin{tabular}[c]{@{}c@{}}TPT\end{tabular} & NeurIPS'22 &
  37.74 & 42.24 & 60.57 & 72.88 & 44.80 & 69.69 & 75.37 & 75.96 & 78.84 & 72.12 & 85.68 & 62.04 & 58.90 & 55.14 & 62.64 & 63.64 \\ 
& \begin{tabular}[c]{@{}c@{}}VTE\end{tabular} & ECCV-W'24 &
  42.42 & 46.26 & 64.23 & 71.10 & 45.58 & 68.50 & 73.66 & 76.75 & 78.27 & 71.02 & 85.28 & 57.24 & 59.54 & 60.59 & 61.85 & 64.15\\ 
& \begin{tabular}[c]{@{}c@{}}WATT-P*\end{tabular} & NeurIPS'24 &
44.70 & 49.66 & 60.48 & 74.35 & 46.33 & 72.21 & 76.28 & 78.18 & 80.97 & 74.90 & 87.66 & 68.63 & 58.35 & 56.13 & 63.47 & 66.15 \\ 
& \begin{tabular}[c]{@{}c@{}}WATT-S*\end{tabular} & NeurIPS'24 &
57.22 & 61.86 & \textbf{67.63} & 78.61 & 53.07 & 77.85 & 80.14 & 81.84 & 83.46 & 80.38 & 89.67 & 78.45 & 65.76 & \textbf{68.63} & \textbf{67.67} & 72.81 \\
& \cellcolor{light_gray}\begin{tabular}[c]{@{}c@{}}Ours\end{tabular} &  &
  \cellcolor{light_gray}\textbf{61.13} & \cellcolor{light_gray}\textbf{64.09} & \cellcolor{light_gray}65.76 & \cellcolor{light_gray}80.51 & \cellcolor{light_gray}\textbf{54.96} & \cellcolor{light_gray}\textbf{80.65} & \cellcolor{light_gray}81.94 & \cellcolor{light_gray}83.04 & \cellcolor{light_gray}\textbf{84.19} & \cellcolor{light_gray}\textbf{80.84} & \cellcolor{light_gray}88.95 & \cellcolor{light_gray}\textbf{82.15} &\cellcolor{light_gray} \cellcolor{light_gray}\textbf{69.16} & \cellcolor{light_gray}62.68 & \cellcolor{light_gray}66.64 & \cellcolor{light_gray}\textbf{73.85} \\ 

\midrule
\multirow{9}{*}{\rotatebox[origin=c]{90}{\LARGE{CIFAR-100C}}}
&\begin{tabular}[c]{@{}c@{}}ZS\end{tabular} & ICLR'21 &
  19.64 & 21.40 & 25.26 & 42.54 & 20.03 & 43.17& 47.95& 48.35& 49.74& 41.57&  57.02&  34.58& 29.15&  23.96& 32.43& 35.79\\ 
& \begin{tabular}[c]{@{}c@{}}TENT\end{tabular} & ICLR'21 &
  7.60 & 8.21 & 8.33 & 51.81 & 7.95 & \textbf{52.45} & 55.34 & 54.16 & 36.17 & 50.92 & 65.63 & \textbf{54.51} & 36.52 & \textbf{43.99} & 35.81 & 37.96 \\ 
& \begin{tabular}[c]{@{}c@{}}RoTTA\end{tabular} & CVPR'23 &
  20.65 & 22.22 & 26.17 & 42.48 & 20.26 & 42.90 & 47.88 & 48.75 & 49.92 & 41.86 & 57.00 & 34.52 & 29.27 & 25.08 & 32.88 & 36.12 \\ 
& \begin{tabular}[c]{@{}c@{}}RPL\end{tabular} & arXiv & 
  6.44 & 7.09 & 7.09 & \textbf{52.16} & 11.81 & 52.33 & \textbf{55.50} & \textbf{54.20 }& 38.83 & \textbf{51.99} & \textbf{66.07} & 54.45 & \textbf{36.86} & 42.83 & \textbf{39.45} & 38.47\\ 
& \begin{tabular}[c]{@{}c@{}}SAR\end{tabular} & ICML'22 & 
  \textbf{25.30} & 27.19 & 32.78 & 47.12 & 23.42 & 47.16 & 51.70 & 51.94 & \textbf{52.48} & 48.77 & 61.54 & 44.50 & 32.26 & 33.67 & 38.06 & 41.19 \\ 
  \cdashline{2-19}
&\begin{tabular}[c]{@{}c@{}}TPT\end{tabular} & NeurIPS'22 & 
  17.95 & 19.51 & 27.13 & 43.53 & 20.08 & 42.65& 48.63& 49.11&  49.48& 42.14& 57.35& 33.26&  31.13& 27.59& 32.75& 36.15\\
&\begin{tabular}[c]{@{}c@{}}VTE\end{tabular} & ECCV-W'24 &
  17.96 & 18.72 & 28.17 & 40.38 & 19.60 &39.50& 45.33& 48.24&46.87&40.73&55.31&30.04& 32.47&30.35& 31.45&35.01\\ 
& \begin{tabular}[c]{@{}c@{}}WATT-P*\end{tabular} & NeurIPS'24 &
20.53 & 22.22 & 27.3 & 43.14 & 17.51 & 42.37 & 48.17 & 47.31 & 49.34 & 41.49 & 57.07 & 35.29 & 27.75 & 25.83 & 31.89 & 35.81 \\  
& \begin{tabular}[c]{@{}c@{}}WATT-S*\end{tabular} & NeurIPS'24 &
21.20 & 23.11 & 28.23 & 44.16 & 18.45 & 43.44 & 49.14 & 48.16 & 50.05 & 42.35 & 57.82 & 36.43 & 28.36 & 26.85 & 32.93 & 36.71\\  
&\cellcolor{light_gray}\begin{tabular}[c]{@{}c@{}}Ours\end{tabular} &  &
 \cellcolor{light_gray}24.91 &\cellcolor{light_gray}\textbf{27.73}& \cellcolor{light_gray}\textbf{33.66} & \cellcolor{light_gray}50.11 & \cellcolor{light_gray}\textbf{26.27} & \cellcolor{light_gray}48.49 & \cellcolor{light_gray}54.85 & \cellcolor{light_gray}52.35 & \cellcolor{light_gray}51.62 & \cellcolor{light_gray}48.38 & \cellcolor{light_gray}63.27 & \cellcolor{light_gray}45.21 & \cellcolor{light_gray}34.74 & \cellcolor{light_gray}32.38 & \cellcolor{light_gray}37.31 & \cellcolor{light_gray}\textbf{42.09} \\

\midrule
\multirow{9}{*}{\rotatebox[origin=c]{90}{\LARGE{ImageNet-C}}}
&\begin{tabular}[c]{@{}c@{}}ZS\end{tabular} & ICLR'21 &
  11.18 & 12.54 & 12.04 & 23.36 & 15.18 & 24.50& 22.58& 32.32& 29.88& 35.88& 54.18& 17.20&  12.72&  30.96&33.26&24.51\\ 
& \begin{tabular}[c]{@{}c@{}}TENT\end{tabular} & ICLR'21 & 
  5.14 & 5.70 & 7.44 & 25.22 & 19.34 & 26.80 & 24.16 & 33.56 & 30.42 & 37.74 & 54.24 & 22.50 & 13.90 & 35.02 & 36.08 & 25.15 \\ 
& \begin{tabular}[c]{@{}c@{}}RoTTA\end{tabular} & CVPR'23 & 
  11.34 & 12.96 & 12.32 & 23.38 & 15.50 & 24.66 & 22.90 & 32.56 & 30.02 & 35.98 &54.32 & 17.20 & 12.80 & 31.06 & 33.46 & 24.78\\
& \begin{tabular}[c]{@{}c@{}}RPL\end{tabular} & arXiv & 
  9.04 & 10.04 & 10.96 & 24.40 & 17.40 & 26.28 & 23.76 & 32.70 & 30.62 & 36.64 & 54.04 & 19.38 & 13.24 & 33.14 & 34.60 & 25.08 \\ 
& \begin{tabular}[c]{@{}c@{}}SAR\end{tabular} & ICML'22 & 
  17.96 & 20.46 & 20.68 & 25.72 & 23.04 & 29.52 & 26.04 & 34.92 & 32.74 & 39.00 & 55.00 & 27.14 & 19.64 & 36.66 & 37.50 & 29.73 \\ \cdashline{2-19}
&\begin{tabular}[c]{@{}c@{}}TPT\end{tabular} & NeurIPS'22 & 
  8.48 & 9.46 & 10.20 & 23.98 & 15.16 & 25.10& 24.00&33.94&32.12&37.08&  55.64&  16.54& 13.68& 34.06&33.58& 24.87\\ 
&\begin{tabular}[c]{@{}c@{}}VTE\end{tabular} & ECCV-W'24 & 
  9.18 & 10.76 & 10.78 & 24.72 & 14.30 & 24.36& 25.24& 35.38 & 32.46 & 38.16& 55.56& 16.14& 14.26& 38.72 & 33.98& 25.60\\
& \begin{tabular}[c]{@{}c@{}}Stat$\textit{A}$ ($\gamma$=0.1)\end{tabular} & CVPR'25 &
10.56 & 11.22 & 10.86 & 22.11 & 14.12 & 22.39 & 20.85 & 31.39 & 29.90 & 34.63 & 53.32 & 16.00 & 12.46 & 29.55 & 32.79 & 23.47 \\
& \begin{tabular}[c]{@{}c@{}}Stat$\textit{A}$  ($\gamma$=-1)\end{tabular} & CVPR'25 &
11.83 & 13.02 & 12.41 & 23.71 & 15.02 & 23.64 & 21.79 & 31.80 & 30.22 & 36.52 & 54.15 & 17.56 & 13.00 & 32.01 & 33.37 & 24.67 \\ 
&\cellcolor{light_gray}\begin{tabular}[c]{@{}c@{}}Ours\end{tabular} &  &
  \cellcolor{light_gray}\textbf{19.32} & \cellcolor{light_gray}\textbf{21.38} & \cellcolor{light_gray}\textbf{19.60} & \cellcolor{light_gray}\textbf{26.58} & \cellcolor{light_gray}\textbf{21.94} & \cellcolor{light_gray}\textbf{ 30.88} & \cellcolor{light_gray}\textbf{29.02} & \cellcolor{light_gray}\textbf{36.48} & \cellcolor{light_gray}32.00 & \cellcolor{light_gray}\textbf{40.98} & \cellcolor{light_gray}\textbf{56.72} & \cellcolor{light_gray}\textbf{ 26.14} & \cellcolor{light_gray}\textbf{23.74} &  \cellcolor{light_gray}37.67& \cellcolor{light_gray}\textbf{38.34} & \cellcolor{light_gray}\textbf{30.72} \\  
\bottomrule
\end{tabular}}
\caption{Mean accuracy (\%) on CIFAR-10C, CIFAR-100C, and ImageNet-C - TTA mean accuracy of the 15 corruptions (tasks) at a severity level of 5, using ViT-B/16.}
\label{tab:mainresults}
\vspace{-15pt}
\end{table*}

\vspace{-0.1cm}

\subsection{Results: Online Test-Time Adaptation}
\begin{figure}[t!]
\centering
\setlength{\tabcolsep}{1pt}
\begin{tabular}{c cccc}

  & { \large \textit{Gaussian}} & {\large \textit{Defocus}} & {\large \textit{Fog}} & { \large \textit{Pixelate}}  \\
  
 \raisebox{3ex}{\rotatebox[]{90}{\small ViT-B/16}} 
  & \includegraphics[width=0.22\columnwidth]{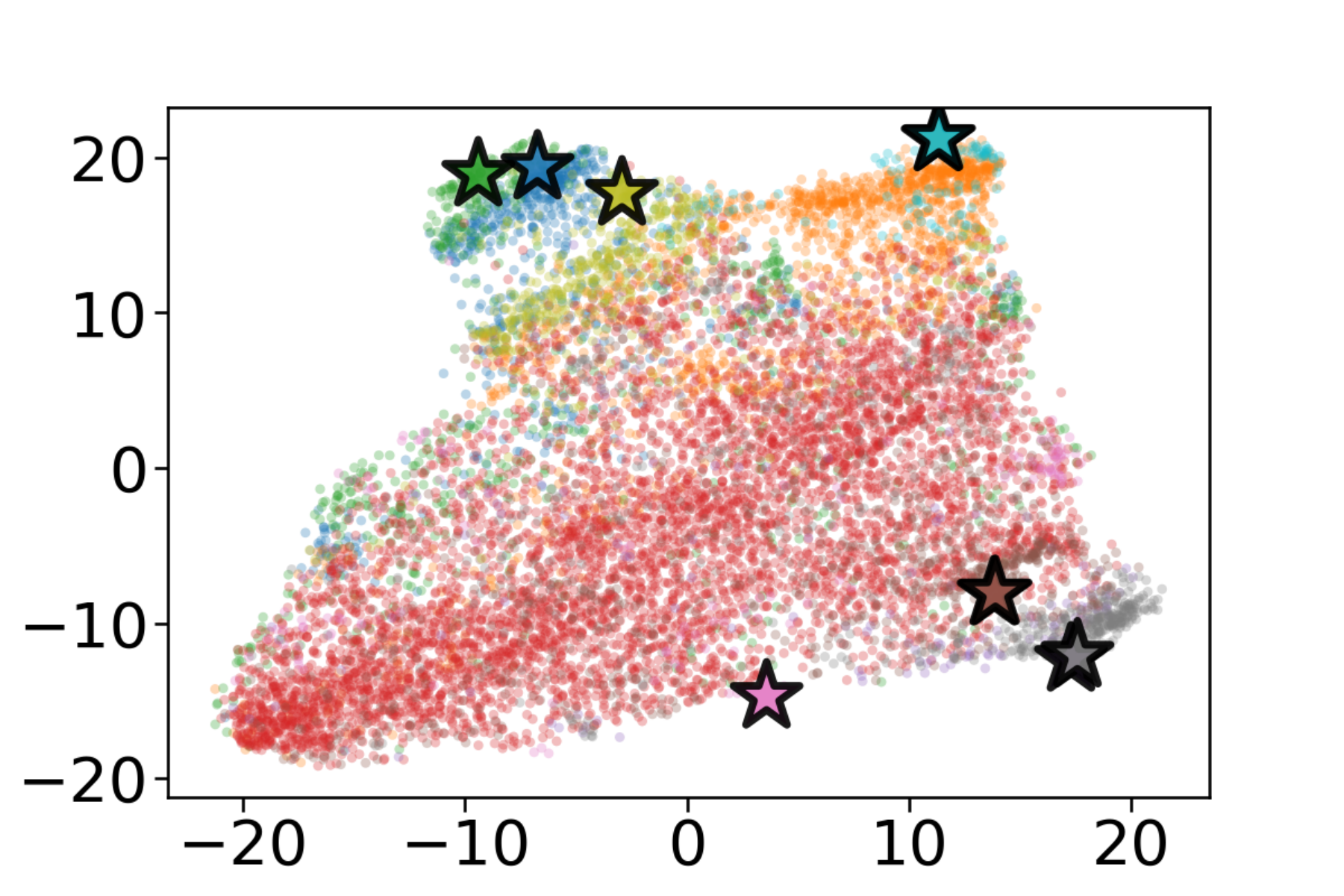}   
  & \includegraphics[width=0.22\columnwidth]{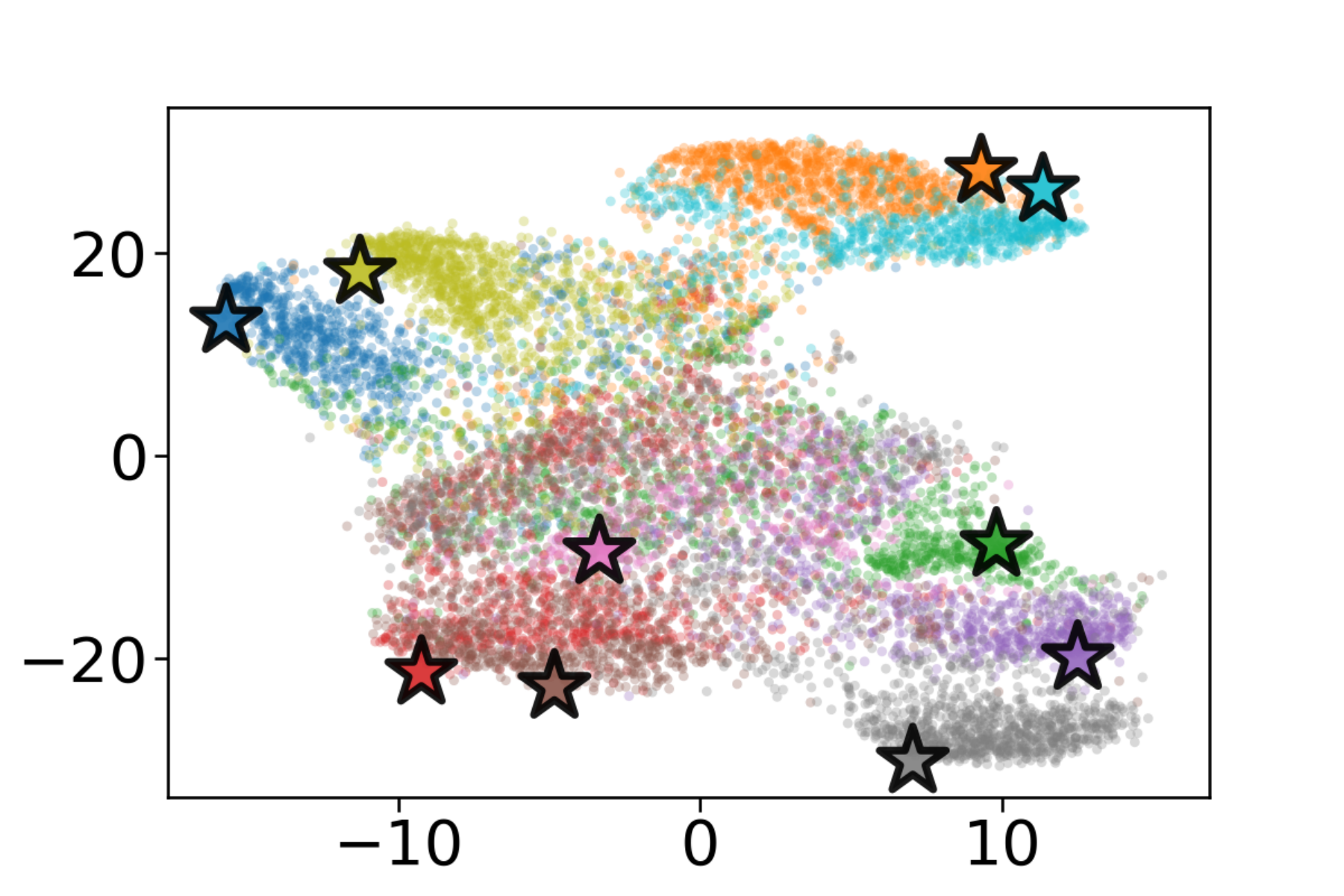}  
  & \includegraphics[width=0.22\columnwidth]{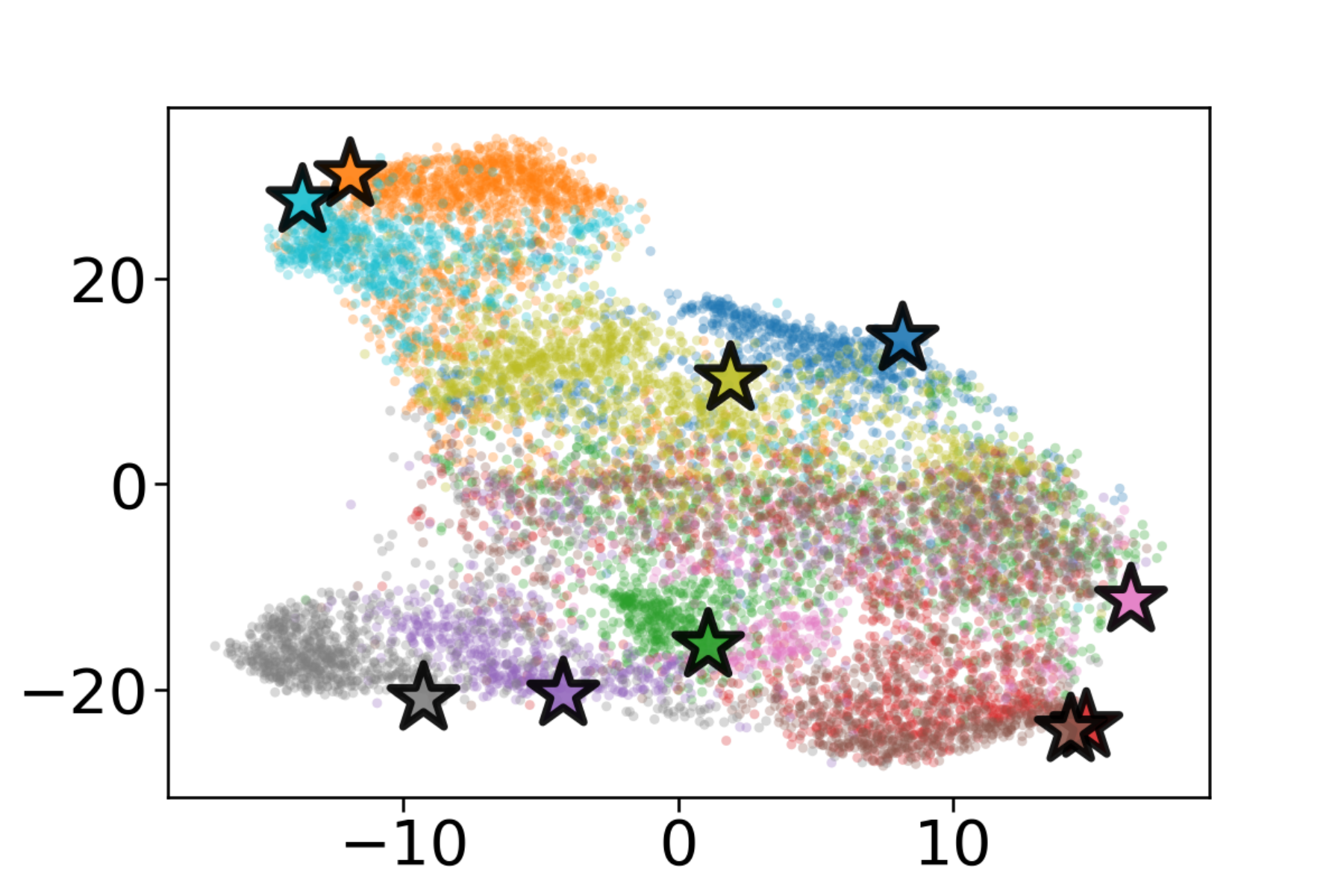}  
  & \includegraphics[width=0.22\columnwidth]{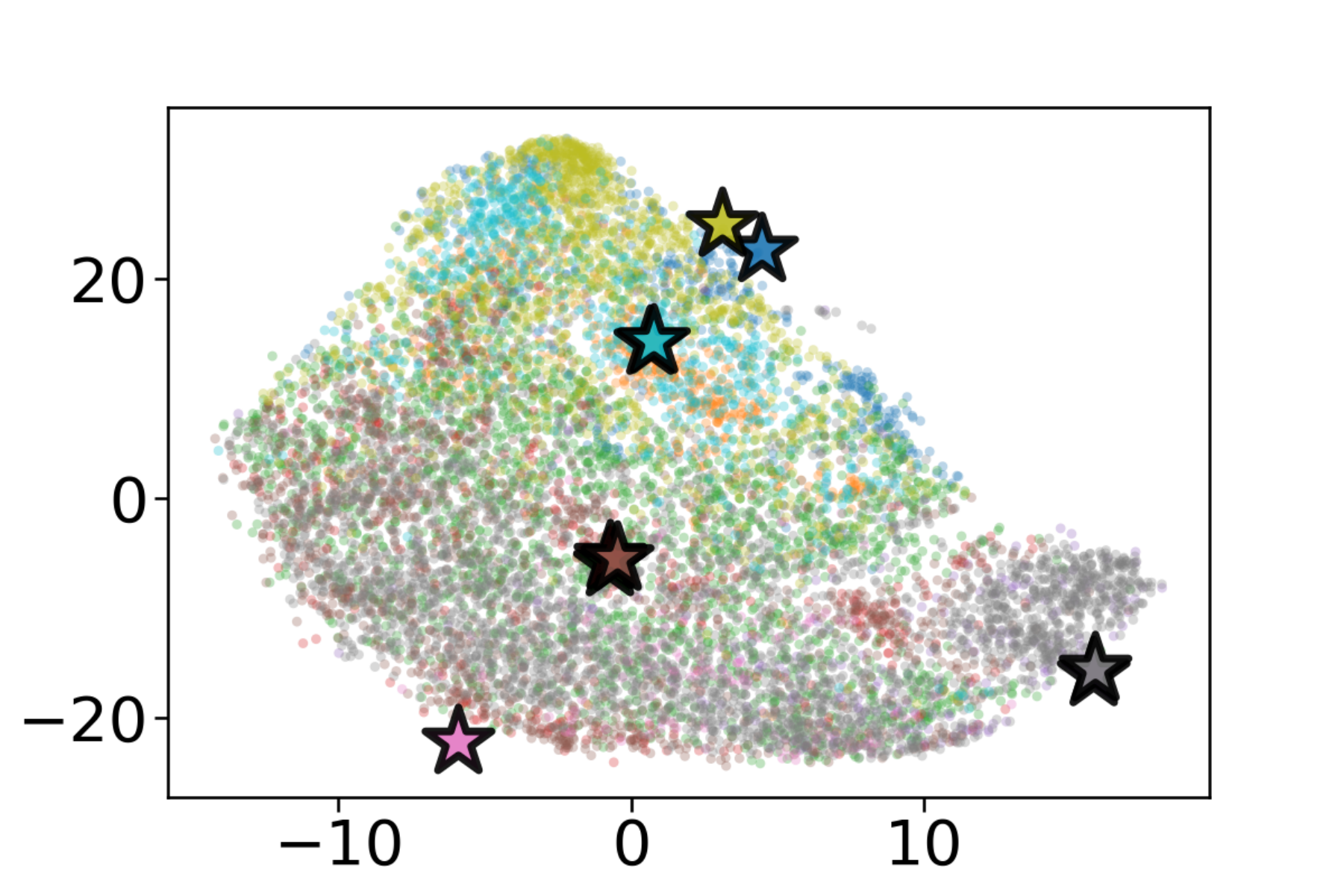} \\
  \midrule
  \raisebox{2.5ex}{\rotatebox[]{90}{\small Ours}} & \includegraphics[width=0.22\columnwidth]{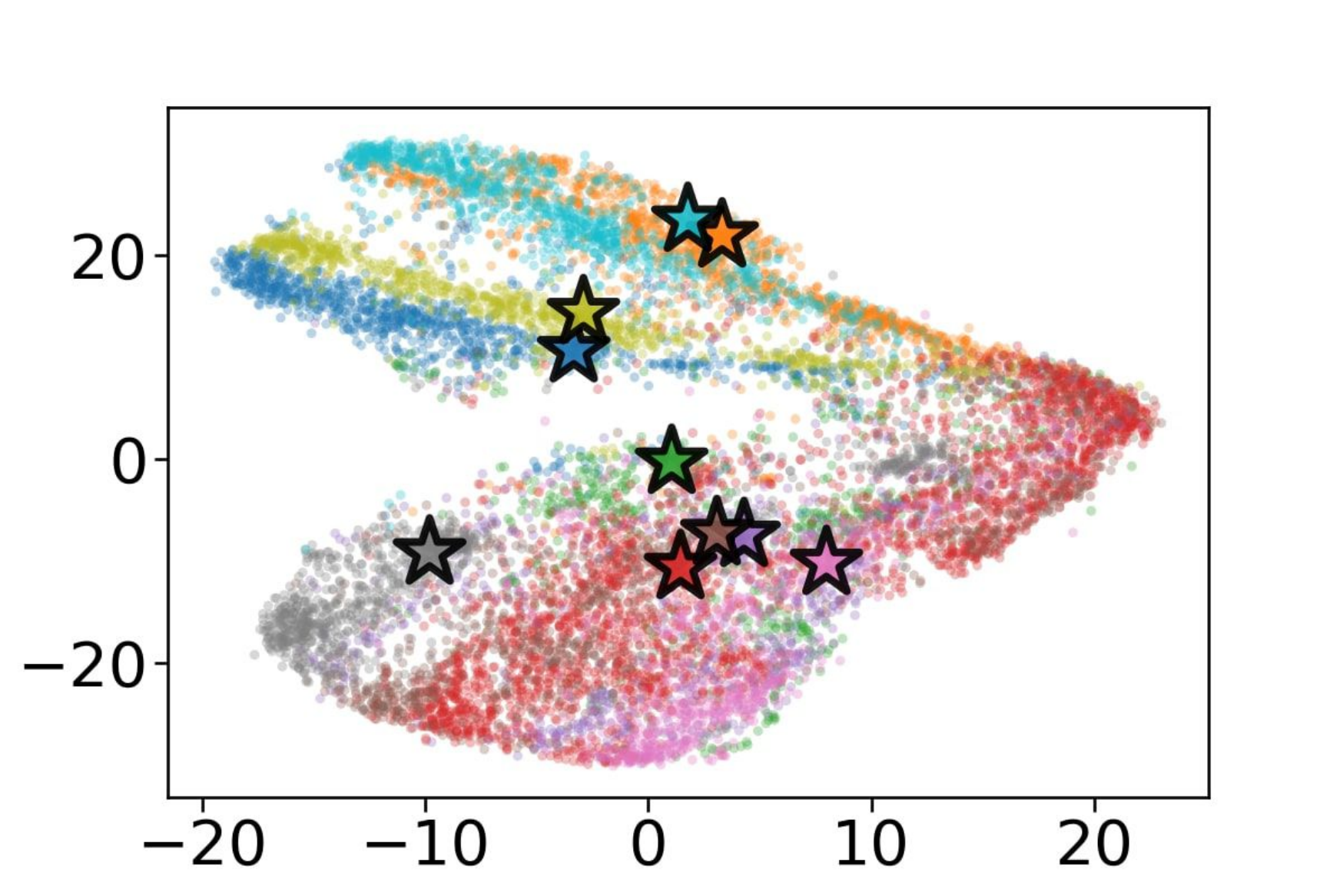}   
  & \includegraphics[width=0.22\columnwidth]{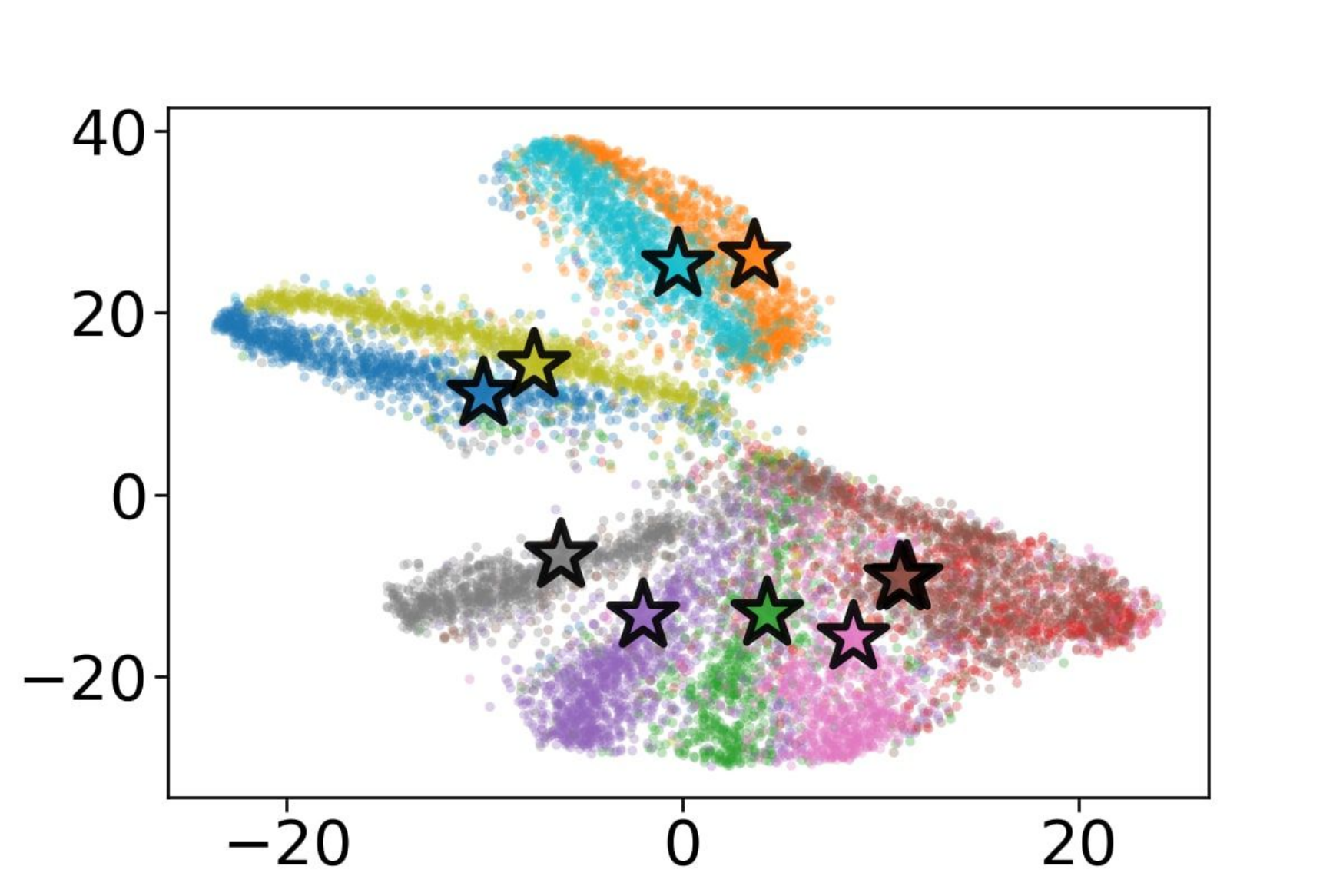}  
  & \includegraphics[width=0.22\columnwidth]{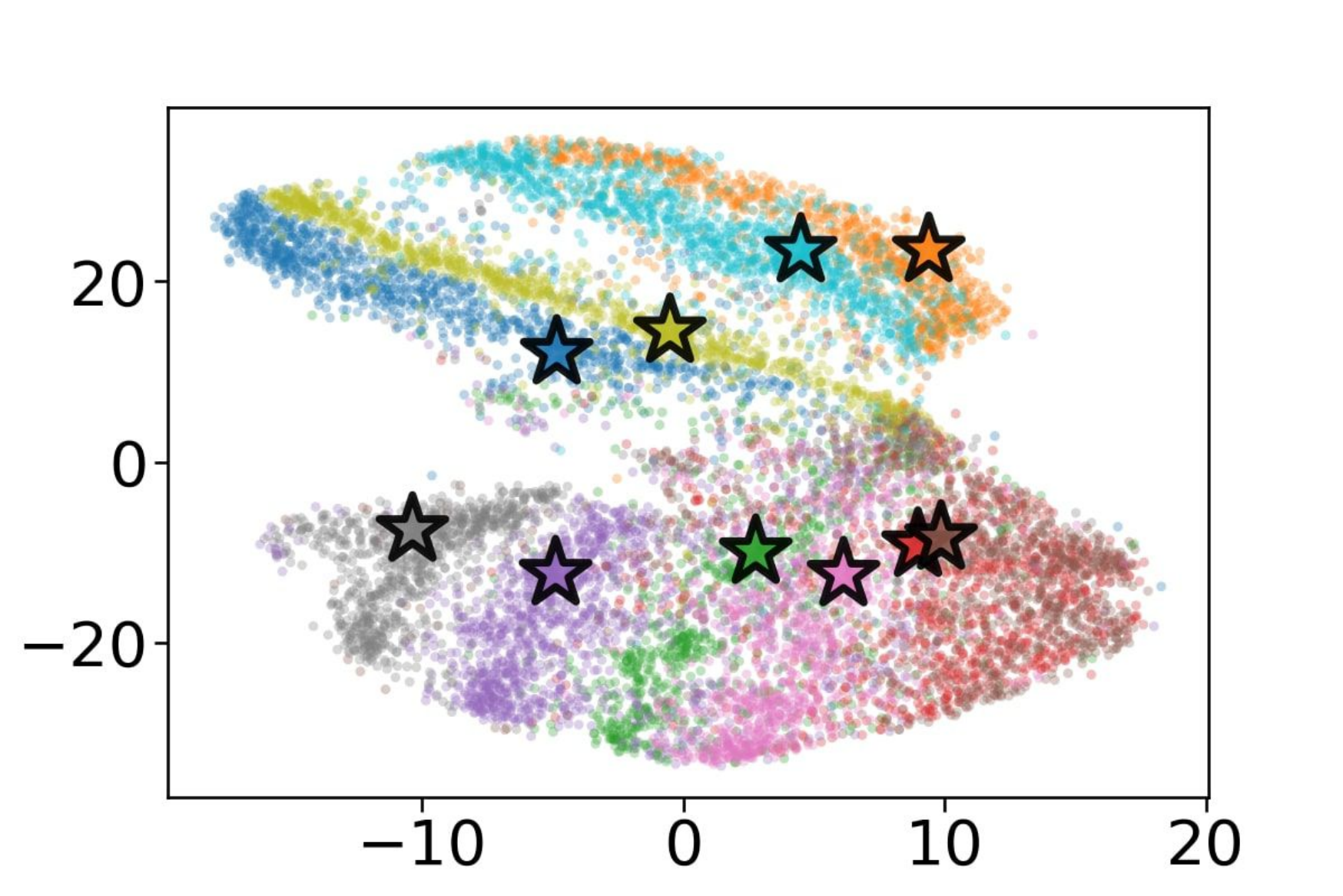}  
  & \includegraphics[width=0.22\columnwidth]{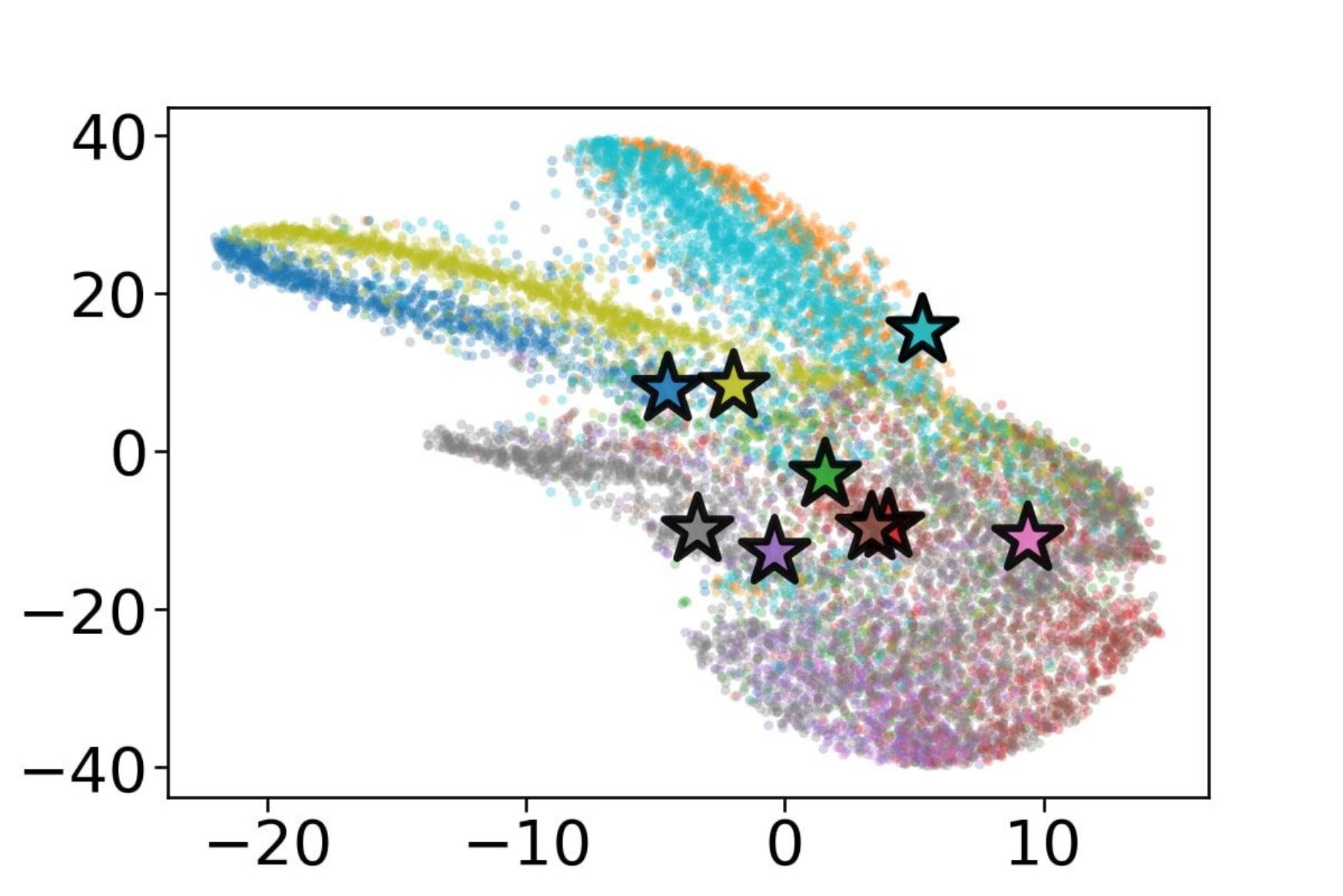} \\
  
\end{tabular}
\caption{\textit{$\framework$ yields more discriminative visual features that exhibit stronger alignment with their corresponding text features.} The t-SNE \cite{van2008visualizing} plots show visual (\(\circ\)) and text ($\bigstar$) features for CIFAR-10C, comparing zero-shot ViT-B/16 with our approach. }
\label{fig: tsne}
\vspace{-5pt}
\end{figure}

\vspace{-3pt}
\noindent \textbf{Results on CIFAR-10C, CIFAR-100C, and ImageNet-C.}
We present the \textbf{online} TTA results in Table \ref{tab:mainresults}. While most TTA methods outperform zero-shot CLIP on average, $\framework$ consistently exceeds the performance of these methods across all datasets. SAR shows significant performance improvements over others. Although it performs similarly to our method on CIFAR-100C and ImageNet-C, $\framework$ outperforms SAR on CIFAR-10C, achieving improvements of 6.48\%. While all the prior approaches operate solely in the vision space, our approach leverages CLIP's joint vision-language feature space, resulting in superior TTA performance. For TTA approaches using CLIP, $\framework$ surpasses TPT and VTE across all tasks and datasets. For our WATT-P* and WATT-S* implementation, adaptation happens once on each prompt template for a batch. Interestingly, despite CLIP being \textit{trained} on the batch by averaging adapted weights from multiple templates before inference, the performance remains subpar. This indicates that WATT \cite{osowiechi2024watt} is not well suited for online TTA. In Stat$\textit{A}$ \cite{zanella2025realistic}, $\gamma$ controls the number of effective classes in a batch, for zero-shot VLM adaptation. On ImageNet-C, the large drop in performance for each task indicates the need for CLIP adaptation to severe image degradation and the importance of relevant text features. In Figure \ref{fig: tsne}, we show t-SNE \cite{van2008visualizing} plots of visual and text features, of a few tasks, and compare them against zero-shot ViT-B/16 for CIFAR-10C. It highlights how our approach enhances alignment with text features, fosters better class separation, and forms more distinct clusters, leading to significant improvements. The Supplementary shows detailed t-SNE plots for CIFAR-10C and CIFAR-100C.

\begin{table}[htb!]
\centering
\small
\resizebox{\linewidth}{!}{
\begin{tabular}{@{}c c c c c c c@{}}
\toprule
Dataset & Domain & ZS & TENT & WATT-P* & WATT-S* & Ours\\
\midrule
\multirow{4}{*}{OfficeHome} 
  & Art        & 78.29 & 79.4 & 80.35 & \textbf{80.43} & \cellcolor{light_gray}79.86 \\
  & Clipart    & 64.03 & 63.18 & 66.85 & \textbf{66.9} & \cellcolor{light_gray}66.44 \\
  & Product    & 87.11 & 88.42 & 87.5 & 87.54 & \cellcolor{light_gray}\textbf{88.51} \\
  & Real World & 88.96 & 89.6 & \textbf{90.04} & 89.99 & \cellcolor{light_gray}89.67 \\
\midrule
\multirow{4}{*}{PACS}
  & Art     & 97.22 & \textbf{98.05} & 97.75 & 97.75 & \cellcolor{light_gray}97.56\\
  & Cartoon & 97.4 & \textbf{97.65} & 97.53 & 97.53 & \cellcolor{light_gray}97.48 \\
  & Photo   & 99.58 & 99.58 & 99.59 & 99.52 & \cellcolor{light_gray}\textbf{99.72} \\
  & Sketch  & 86.23 & \textbf{88.75} & 88.52 & 88.65 & \cellcolor{light_gray}87.76 \\
\midrule
\multirow{4}{*}{VLCS}
  & Caltech101 & 99.43 & 99.43 & 99.36 & 99.36 & \cellcolor{light_gray}\textbf{99.51}\\
  & LabelMe    & 68.15 & 68.14 & 66.92 & 68.49 & \cellcolor{light_gray}\textbf{68.94} \\
  & SUN09      & 73.4 & 73.4 & 74.53 & 74.68 & \cellcolor{light_gray}\textbf{75.23} \\
  & VOC2007    & 84.75 & 84.75 & 84.0 & 84.03 & \cellcolor{light_gray}\textbf{85.6}  \\
\midrule
\multirow{4}{*}{TerraInc}
& L38 & 20.3 & 20.3 & 27.79 & 29.08 & \cellcolor{light_gray}\textbf{37.33} \\
& L43 & 31.52 & 31.52 & 33.98 & \textbf{34.13} & \cellcolor{light_gray}32.71 \\
& L46 & 28.98 & 28.98 & 27.07 & 28.13 & \cellcolor{light_gray}\textbf{31.05} \\
& L100 & 52.35 & 52.35 & 43.59 & 42.32 & \cellcolor{light_gray}\textbf{55.22} \\
\bottomrule
\end{tabular}}
\vspace{-5pt}
\caption{Accuracy on domain generalization datasets \cite{gulrajani2020search} with other domain shifts using a ViT-B/16 visual backbone.}
\label{tab:style}
\end{table}

\subsection{Results: Domain Generalization Datasets}
\vspace{-3pt}
In this section, we evaluate our $\framework$ on commonly used domain generalization datasets \cite{gulrajani2020search} and benchmark against zero-shot ViT-B/16 (ZS), TENT, WATT-P* and WATT-S*. From Table \ref{tab:style}, we observe that $\framework$ does better or comparable, across all datasets and domain shifts, to all the baselines. TENT matches zero-shot CLIP since LN normalization works per image across the features, independent of the batch statistics. We demonstrate that $\framework$ can be generalized and be robust to other various domain shifts, with a single step adaptation and global prompt template as opposed to WATT-P* and WATT-S* that use multiple prompt templates. The largest performance improvements come on the Terra Incognita dataset. Large style variations in OfficeHome and PACS make it difficult for CLIP to generate effective text features, with a single step, from our default prompt. The implementation details are described in the Supplementary.

\vspace{-3pt}
\subsection{Ablation Study and Analysis}
\vspace{-2pt}

\begin{figure}[htb!]
\vspace{-8pt}
\small
\centering
\setlength{\tabcolsep}{2pt}
    \begin{tabular}{c c c}
     \includegraphics[width=0.33\columnwidth]{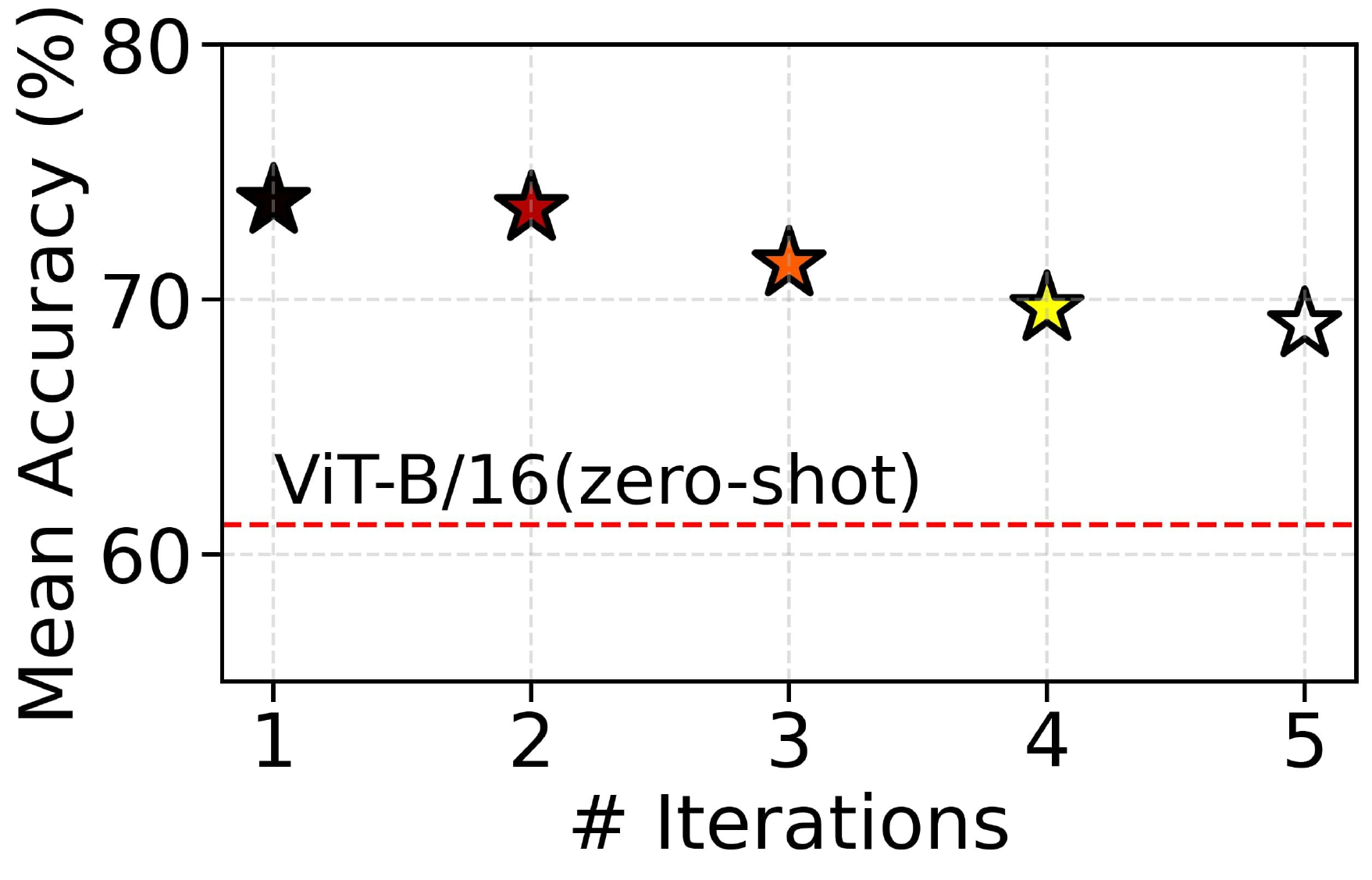} & 
     \includegraphics[width=0.33\columnwidth]{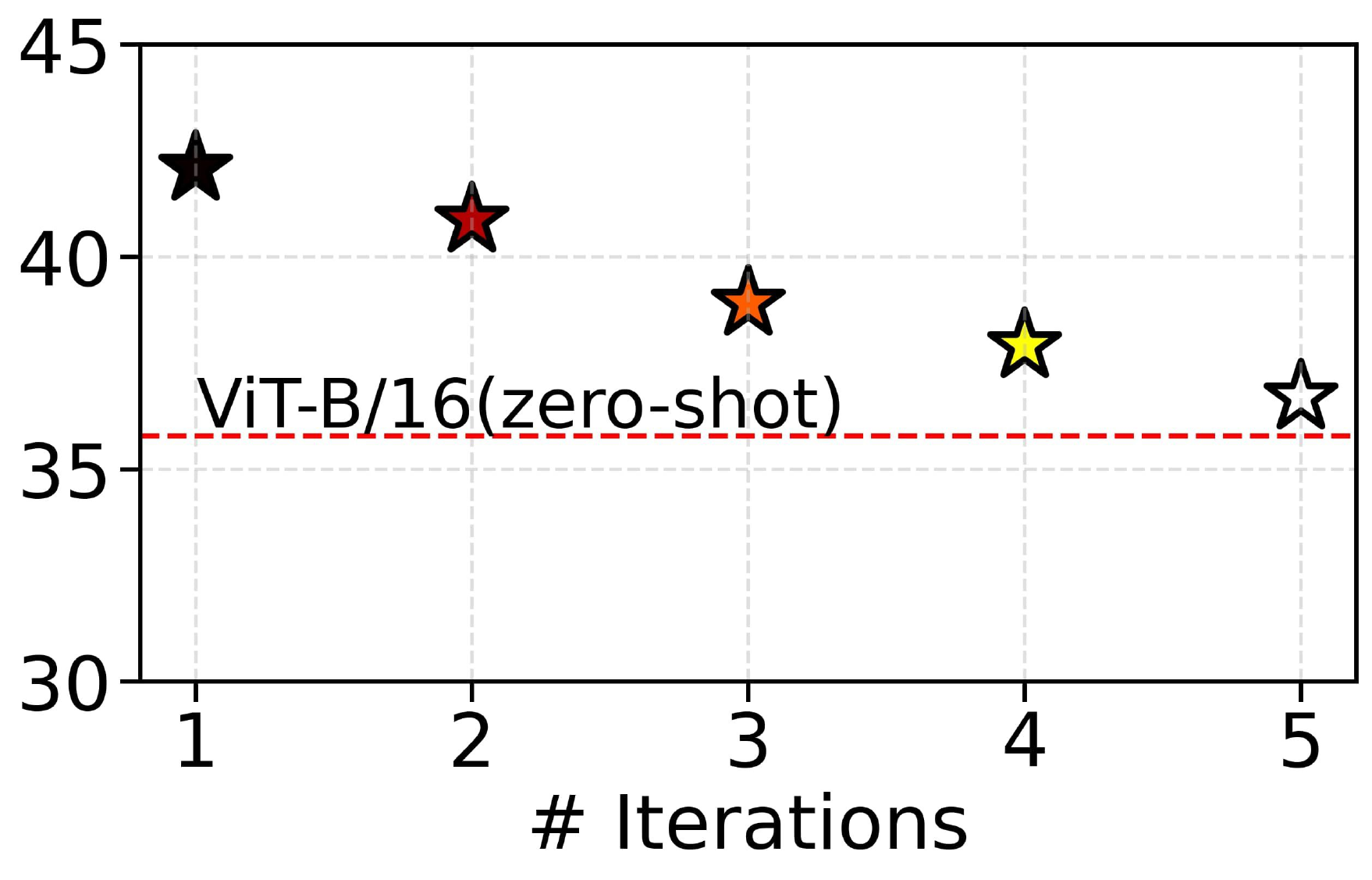} &
     \includegraphics[width=0.33\columnwidth, trim = 0mm 3mm 0mm 0mm, clip]{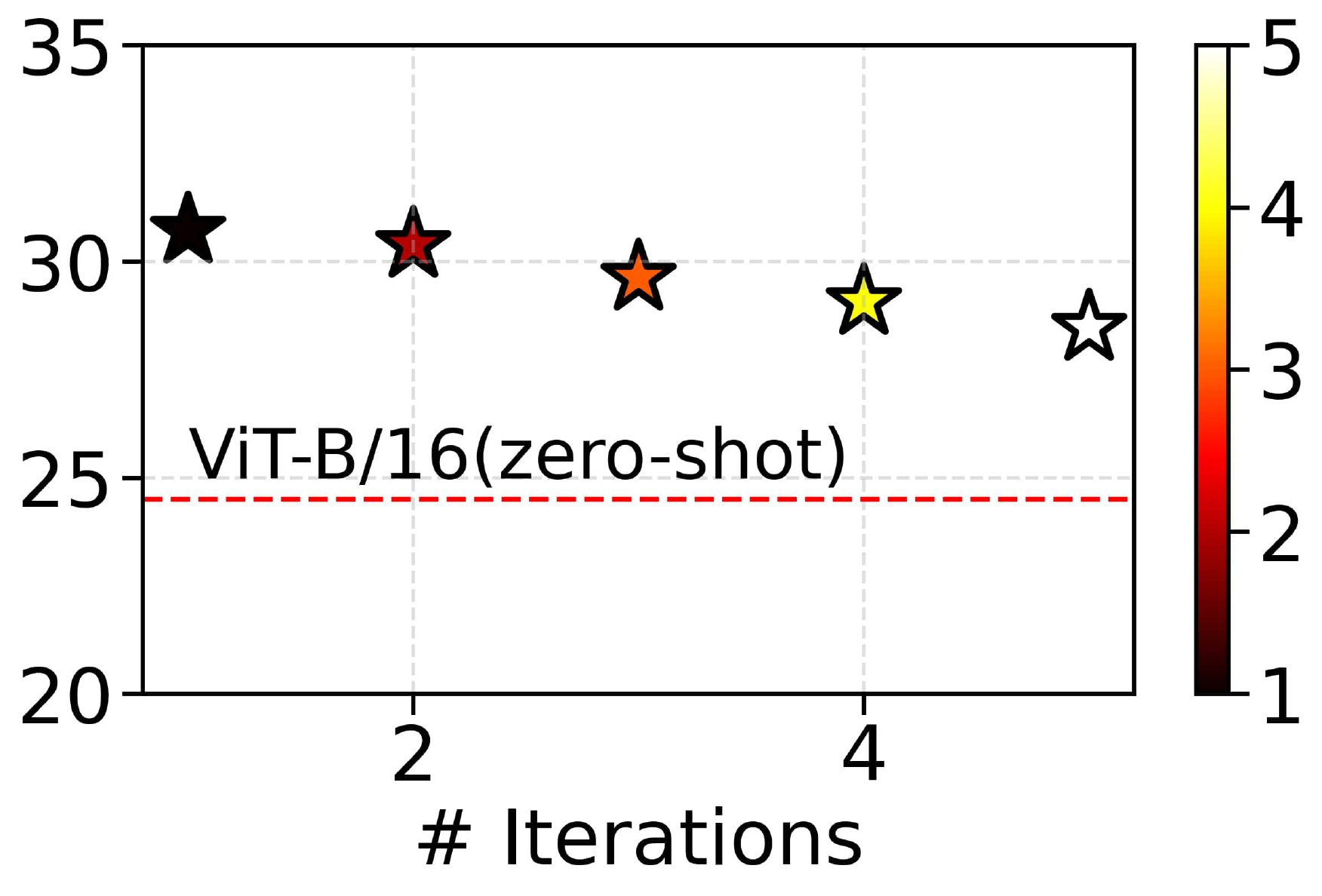} \\
     a) CIFAR-10C  & b) CIFAR-100C & c) ImageNet-C \\
    \end{tabular}
    \caption{Analysis on increasing \# iterations: Mean accuracy for \# iterations, on each test batch, using a ViT-B/16 backbone.}
    \label{fig:multiplesteps}
    \vspace{-10pt}
\end{figure}

\vspace{-5pt}
\begin{table}[ht]
\large
\centering
\vspace{-2pt}
\resizebox{\columnwidth}{!}{
\begin{tabular}{cccc}
\toprule
Backbone (ViT-B/16) & CIFAR-10C & CIFAR-100C & ImageNet-C \\ \midrule
$\mathcal{L}_{ent}$ & 60.65 & 38.17 & 24.03 \\
$\mathcal{L}_{pm}$ & 61.23 &  36.63 & 22.83 \\
$\mathcal{L}_{sp}$ & 73.16 & 41.36 & 30.05 \\
$\mathcal{L}_{ent}$ + $\mathcal{L}_{pm}$ & 62.60 & 39.32 & 25.21 \\
$\mathcal{L}_{ent}$ + $\mathcal{L}_{sp}$ & 72.69 & 41.84 & 30.08 \\
$\mathcal{L}_{ent}$ + $0.9\mathcal{L}_{pm}$ + 0.1$\mathcal{L}_{sp}$ & 72.21 & 42.20 & 30.82 \\
$\mathcal{L}_{ent}$ + $0.1\mathcal{L}_{pm}$ + 0.9$\mathcal{L}_{sp}$& 72.96 & 41.85 & 30.20 \\
$\mathcal{L}_{ent}$ + $0.3(\mathcal{L}_{pm}$+$\mathcal{L}_{sp}$)& 73.31 & 41.93 & 30.40 \\
$\mathcal{L}_{ent}$ + $0.7(\mathcal{L}_{pm}$ +$\mathcal{L}_{sp}$)& 73.85 & 42.04 & 30.61 \\
$\mathcal{L}_{ent}$ + $\mathcal{L}_{pm}$ + $\mathcal{L}_{sp}$ & \textbf{73.85} & \textbf{42.09} & \textbf{30.72} \\ 
\bottomrule
\end{tabular}}
\caption{Ablation on loss components - Mean accuracy (in \%).}
\label{tab:lossablation}
\end{table}

\noindent \textbf{Adaptation for multiple iterations.} Here, we analyze adaptation for multiple iterations on a single batch, shown in Figure \ref{fig:multiplesteps}, using the ViT-B/16 backbone. Continuous adaptation of LN parameters, which normalize along the features of a single sample, can lead to overfitting as it fits to the noise patterns of the batch, ultimately degrading generalization to other domains. This could also lead to a decline in the loss of CLIP pre-trained knowledge. In the Supplementary, we report the post-adaptation results back on the source test sets - CIFAR10 and CIFAR100 to evaluate \textit{catastrophic forgetting}. Our approach outperforms zero-shot CLIP, with a smaller decline in mean accuracy, especially on CIFAR-10C and ImageNet-C, showing that our losses improve robustness. 
under challenging conditions.

\noindent \textbf{Ablation on loss components.} In Table \ref{tab:lossablation}, we ablate the loss components using ViT-B/16 (see more in the Supplementary). $\mathcal{L}_{pm}$ (row 4) and $\mathcal{L}_{sp}$ (row 5) improve model performance than entropy minimization via $\mathcal{L}_{ent}$. In fact, the addition of $\mathcal{L}_{sp}$ brings larger improvements, proving that $f_{vis}$ produces discriminative features. Interestingly, using only $\mathcal{L}_{ent}$ achieves comparable or better accuracy than zero-shot CLIP, with fully hyperparameter-free objectives.





\begin{figure}[t!]
\vspace{-5pt}
\small
\centering
\setlength{\tabcolsep}{3pt}
    \begin{tabular}{c c}
     \includegraphics[width=0.5\columnwidth, trim = 1mm 1mm 1mm 1mm , clip]{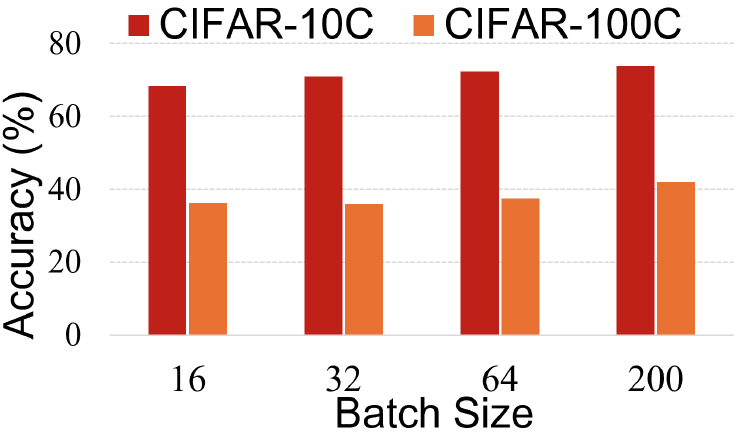} & 
    \includegraphics[width=0.5\columnwidth, trim = 1mm 1mm 1mm 1mm , clip]{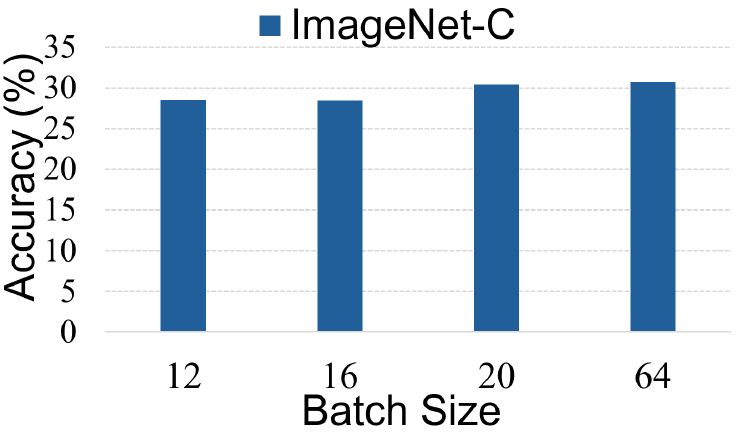} \\
    \end{tabular}
    \caption{Ablation on low batch size - CIFAR-10C and CIFAR-100C (\textit{left}) and ImageNet-C (\textit{right}) using a ViT-B/16 backbone.}
    \label{fig:batch_ablation}
    \vspace{-10pt}
\end{figure}

\noindent \textbf{Low batch size setting.} In Eq \ref{eq:l_pm}, the prototype of each class can be affected by the batch size, with a lower size indicating the absence of a few classes. In Figure \ref{fig:batch_ablation}, we illustrate the mean accuracies on the benchmark datasets. As seen, \framework\ still outperforms or performs comparably to TPT and VTE in low batch size settings. Since TTA happens at the batch-level of a specific domain, the prototypes are computed with the available pseudo-labels, aiding in the continual learning of the feature space as new batches come.

\begin{table}[htb!]
\centering
\footnotesize
\setlength{\tabcolsep}{2pt} 
\vspace{-7pt}
\begin{subtable}[t]{0.45\columnwidth} 
\centering
    \begin{tabular}{ccccc}
    \toprule
    Dataset & ZS & TPT & VTE & Ours \\ 
    \midrule
    CIFAR-10C & 75.84 & 77.75 & 75.32 & 84.74\\ 
    \midrule
    CIFAR-100C & 47.82 & 49.62 & 48.53 & 49.25\\ 
    \midrule
    ImageNet-C & 39.55 & 40.96 & 40.55 & 42.84\\ 
    \bottomrule
    \end{tabular}
\vspace{-1pt}
\caption{Mean acc. w/. a ViT-L/14.}
\label{tab:results3}
\end{subtable}
\hfill
\begin{subtable}[t]{0.45\columnwidth} 
\centering
\setlength{\tabcolsep}{2pt} 
    \begin{tabular}{cc}
    \toprule
    TTA Method & ImageNet-C \\ 
    \midrule
    TENT & 40.62 \\
    RPL & 40.00 \\
    RoTTA & 39.76 \\
    SAR & 42.10 \\
    Ours & \textbf{42.84} \\
    \bottomrule
    \end{tabular}
\vspace{-1pt}
\caption{Results on ImageNet-C.}
\label{tab:results5}
\end{subtable}
\vspace{-9pt}
\caption{Ablation on ViT-L/14 - Mean accuracy (in \%).}
\vspace{-16pt}
\label{tab:backbones}
\end{table}

\noindent \textbf{Extension to larger backbones.} We use ViT-L/14 and report the results in Table \ref{tab:backbones}. \textit{The key advantage lies in the projection matching loss ($\mathcal{L}_{pm}$) which updates the text encoder to produce more image-aware text features—an aspect missing in other state-of-the-art TTA baselines.}

\noindent \textbf{Computational efficiency.} Due to an additional text encoder update, $\framework$ updates $\sim0.044$\% of all CLIP parameters, with a ViT-B/16 backbone, as opposed to $\sim0.026$\% in other TTA baselines. Though the increase is marginal, this leads to superior performances as adaptation of the text encoder produces input-aware CLIP text features, aligning better with the image features. On an NVIDIA RTX A5000 GPU, it takes 0.2s/batch on ImageNet-C with a batch size of 64, compared to 2.34s for WATT \cite{osowiechi2024watt}. Also, WATT adapts per template (for several iterations), making it expensive every time. $\framework$ achieves faster on-the-fly adaptation rates and is favorable for real-time deployment. 

\vspace{-3pt}

\section{Conclusion}
\label{conclusion}

\vspace{-5pt}
In this work, we propose $\framework$, a \textit{bimodal} \textbf{online} test-time adaptation framework for CLIP aimed at handling diverse image corruptions simulating real-time environments. Our in-depth analysis of CLIP's zero-shot performance under increasing corruption severity reveals significant shortcomings in generalization, highlighting the need for effective adaptation. Although prior work on TTA for CLIP has predominantly been \textit{unimodal} focusing on prompt-tuning or prompt ensemble with no model updates,  $\framework$, through synergistic CLIP encoder updates,  encourages learning rich class-separated visual features via vision encoder updates and strengthens alignment between the image class prototype and corresponding text feature via text encoder updates. Empirical studies, including ablations, show robust performance gains over all baselines on benchmark corruption and domain generalization datasets.

\section{Acknowledgement}
This project was supported by a grant from UT Dallas and an NVIDIA Academic Grant Program Award.


{
    \small
    \bibliographystyle{ieeenat_fullname}
    \bibliography{main}
}

\clearpage
\setcounter{page}{1}
\maketitlesupplementary

In this work, we study the problem of \textbf{online} test-time adaptation (TTA) of CLIP \cite{radford2021learning} towards common image corruptions and propose improved schemes for increasing the robustness of CLIP. In addition, to demonstrate the broad impact of our proposed approach, we evaluate on common domain generalization datasets \cite{gulrajani2020search}-OfficeHome \cite{venkateswara2017deep}, PACS \cite{li2017deeper}, VLCS \cite{fang2013unbiased}, and Terra Incognita \cite{beery2018recognition}. We put forward a \textit{bimodal} domain adaptation scheme, at test-time, wherein we exploit the shared feature space of CLIP. In essence, leaning towards a more effective multimodal learning and adaptation method, we propose loss components that improve alignment between the class-specific visual prototype and corresponding text features via maximizing the projection. We also increase the cosine distance between the class prototypes to enhance discrimination between visual features. In this Supplementary, we provide additional insights and experimental results, that has been organized as follows, 

\begin{enumerate} 
    \item Section \ref{dataset} offers a detailed discussion of the standard common corruption datasets \cite{hendrycks2019benchmarking} used, supplemented with visual illustrations. 
    \item To ensure full transparency, we outline the implementation details of all the methods in Section \ref{implementation}, including those for prior TTA approaches \cite{wang2020tent, yuan2023robust, rusak2021if, niu2023towards} adapted for CLIP. We also discuss details of the adapted version of WATT \cite{osowiechi2024watt} - WATT-P* and WATT-S* for online TTA, and other details of experiments run on the domain generalization datasets. \item Section \ref{add_results} presents further results and analysis: \begin{itemize} 
        \item In subsection \ref{severity_backbones}, we explore the limitations of zero-shot CLIP when using ViT-B/32 and ViT-L/14 backbones under increasing image corruption severity. 
        \item In Subsection \ref{tta_suppl}, we report the main online TTA results on CIFAR-10C, CIFAR-100C, and ImageNet-C with a ViT-B/32 backbone. 
        \item We study the effect of different prompt templates on $\framework$ in Subsection \ref{effect_prompts}. Subsection \ref{bimodal} discusses the ablation of $\mathcal{L}_{pm}$ which is responsible for updating the text encoder.
        \item In subsections \ref{loss_add_study} and \ref{post_adap},  we present a detailed loss ablation study and the post-adaptation zero-shot generalization on source test sets, respectively --- essentially evaluating \textit{catastrophic forgetting} of \framework.
        \item Lastly, we include task-wise t-SNE visualizations in subsection \ref{add_tsne} for CIFAR-10C and CIFAR-100C, comparing our method against zero-shot CLIP (ViT-B/16), to illustrate the effectiveness of \framework.
        \end{itemize}         
\end{enumerate}

\begin{figure}[t!]
\centering
\setlength{\tabcolsep}{1pt}
\begin{tabular}{ccccc}

  { \large \textit{Gaussian}} & {\large \textit{Shot}} & {\large \textit{Impulse}} & {\large \textit{Defocus}} & {\large \textit{Glass}}\\
  
  \includegraphics[width=0.15\columnwidth]{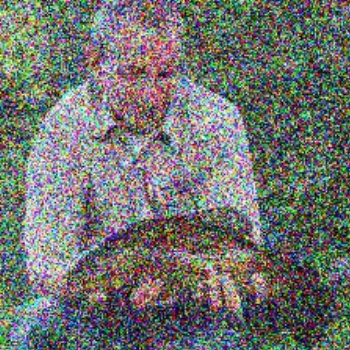}   
  & \includegraphics[width=0.15\columnwidth]{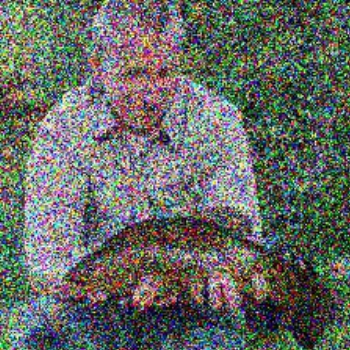}  
  & \includegraphics[width=0.15\columnwidth]{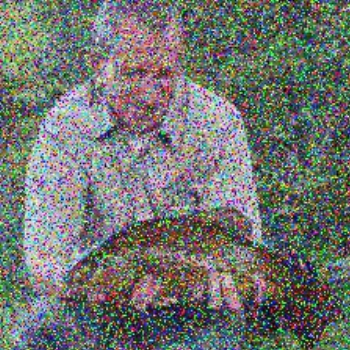}
  & \includegraphics[width=0.15\columnwidth]{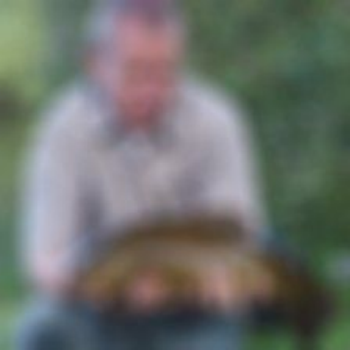}
  & \includegraphics[width=0.15\columnwidth]{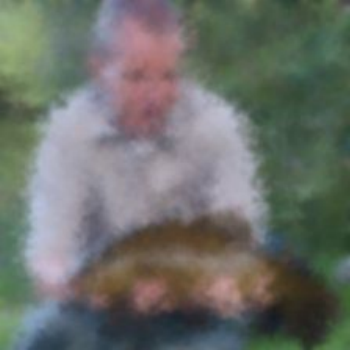}\\

  { \large \textit{Motion}} & {\large \textit{Zoom}} & {\large \textit{Snow}} & {\large \textit{Frost}} & {\large \textit{Fog}}\\
  \includegraphics[width=0.15\columnwidth]{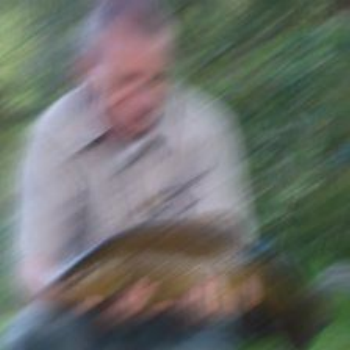}   
  & \includegraphics[width=0.15\columnwidth]{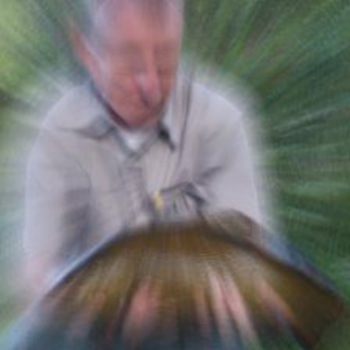}  
  & \includegraphics[width=0.15\columnwidth]{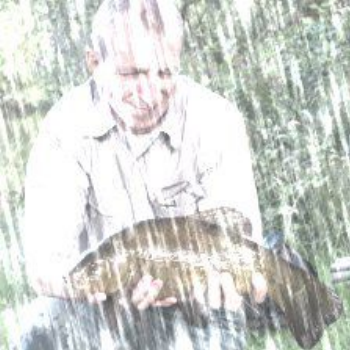}
  & \includegraphics[width=0.15\columnwidth]{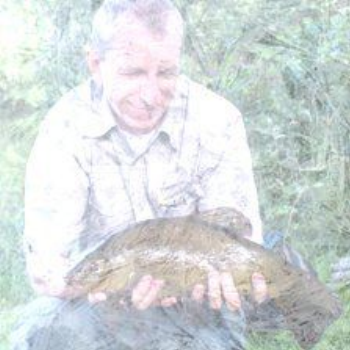}
  & \includegraphics[width=0.15\columnwidth]{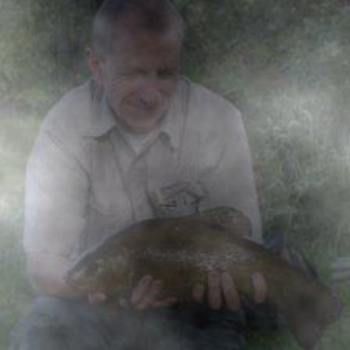}\\

  { \large \textit{Brightness}} & {\large \textit{Contrast}} & {\large \textit{Elastic}} & {\large \textit{Pixelate}} & {\large \textit{JPEG}}\\
  \includegraphics[width=0.15\columnwidth]{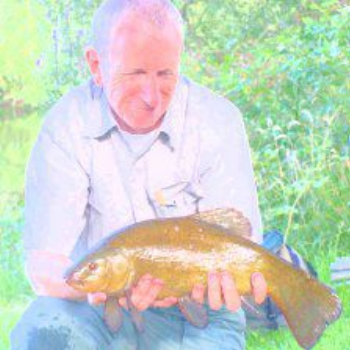}   
  & \includegraphics[width=0.15\columnwidth]{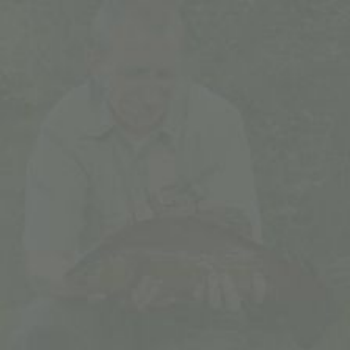}  
  & \includegraphics[width=0.15\columnwidth]{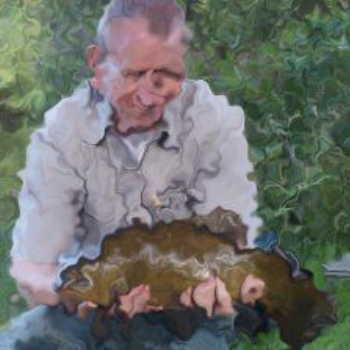}
  & \includegraphics[width=0.15\columnwidth]{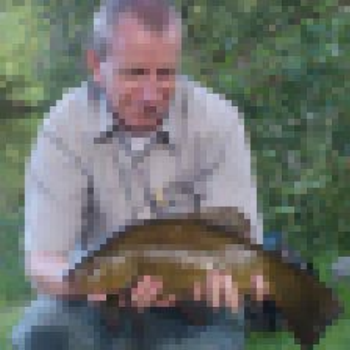}
  & \includegraphics[width=0.15\columnwidth]{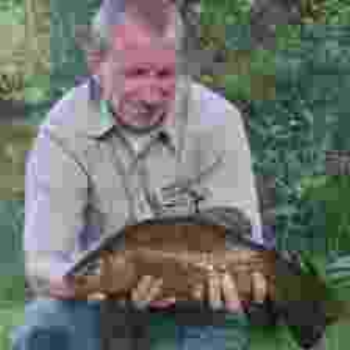}\\

\end{tabular}
\caption{We provide visualizations of an image from ImageNet-C \cite{hendrycks2019benchmarking} for different corruption types, at an image severity level of 5.}
\label{fig: viz}
\end{figure}

\subsection{Datasets}\label{dataset}

We employ the \textbf{CIFAR-10C}, \textbf{CIFAR-100C}, and \textbf{ImageNet-C} datasets, for our experiments, as introduced by \cite{hendrycks2019benchmarking}. Each dataset includes 15 distinct types of image corruptions, referred to as tasks in a test-time adaptation setting, applied to the test sets of CIFAR10, CIFAR100 \cite{krizhevsky2009learning}, and ImageNet \cite{deng2009imagenet}. These corruptions are applied at 5 different severity levels, ranging from mild to severe. For each task, CIFAR-10C and CIFAR-100C contain 10,000 test samples, whereas ImageNet-C has 5000 samples.

The image corruptions are categorized into four primary groups: noise, blur, weather, and digital distortions. Noise-based corruptions include \textit{Gaussian}, \textit{Shot}, and \textit{Impulse} noise, which introduce random pixel-level variations. The blur category encompasses \textit{Defocus}, \textit{Glass}, \textit{Motion}, and \textit{Zoom} blur effects, all of which simulate different types of distorted imagery. Weather-related corruptions, such as \textit{Snow}, \textit{Frost}, and \textit{Fog}, replicate environmental conditions that obscure image details. Lastly, digital distortions include effects like \textit{Brightness}, \textit{Contrast}, \textit{Elastic Transform}, \textit{Pixelate}, and \textit{JPEG} compression, which reflect various forms of post-processing or compression artifacts that degrade image quality.

These corruption types, as proposed by \cite{hendrycks2019benchmarking}, provide a comprehensive framework for assessing model robustness, which has been and is still being studied \cite{hendrycks2016early, metzen2017detecting, papernot2016distillation, subbaswamy2021evaluating, liu2024comprehensive}. Their ability to emulate real-world image degradation scenarios is advantageous, allowing for a more realistic evaluation of a model's robustness. We provide corruption visualizations, via an image example, in Figure \ref{fig: viz}. We urge the readers to check out \cite{hendrycks2019benchmarking} for further inspection.

\begin{figure*}[htb!]
\small
\centering
\setlength{\tabcolsep}{5pt}
\begin{tabular}{c c}
\includegraphics[width=\textwidth, trim = 0mm 0mm 0mm 0mm, clip]{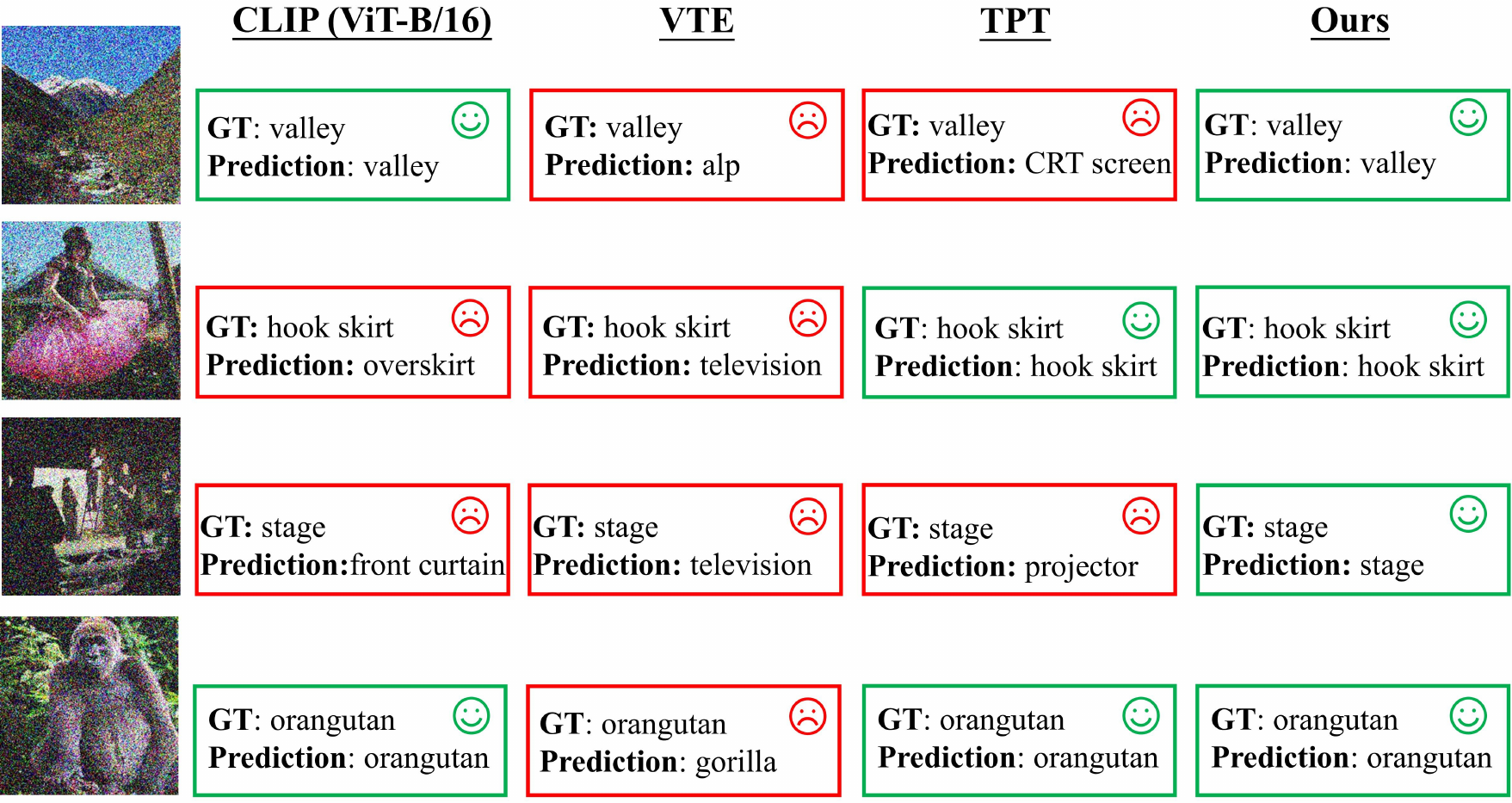}
\end{tabular}
\caption{Comparison of classification predictions across various methods (Zero-shot CLIP (ViT-B/16), VTE \cite{dobler2024lost}, TPT \cite{shu2022test}, and Ours) on ImageNet-C samples with \textit{Gaussian} noise. Each row illustrates an example, displaying the ground truth (GT) label alongside the predictions from each method. Correct predictions are highlighted in green, while incorrect ones are marked in red. Our approach demonstrates enhanced robustness and higher accuracy, especially in challenging image corruption conditions.
}
\label{fig:add_comparison}
\end{figure*}

\subsection{Implementation Details}\label{implementation}

In this section, we summarize the implementation details of all the baseline methods that have been mentioned in the main paper, including ours. We build our approach on the standard benchmark code base \footnote{\href{https://github.com/mariodoebler/test-time-adaptation/tree/maink}{https://github.com/mariodoebler/test-time-adaptation/tree/main}} that also houses the hyperparameters and training details of all the prior TTA methods. CLIP-like models are used as provided by \href{https://github.com/mlfoundations/open_clip}{OpenCLIP}. Only the vision encoder is updated for existing online TTA methods \cite{wang2020tent, yuan2023robust, rusak2021if, niu2023towards} adopted for CLIP.

\subsubsection{Experiments on CIFAR-10C, CIFAR-100C, and ImageNet-C}

\noindent \textbf{$\framework$ (Ours)}: For domain-specific test adaptation, we conducted experiments using ViT-B/16 and ViT-B/32 \citep{dosovitskiy2020image} as the vision backbones. For CIFAR-10C, both the vision encoder ($f_{vis}$) and text encoder ($f_{txt}$) were updated using the AdamW optimizer with a learning rate of 10$^{-3}$. Similarly, for CIFAR-100C and ImageNet-C, we employed the Adam optimizer and AdamW optimizer, respectively, with a learning rate of 5$\times$10$^{-4}$. The batch size $\mathcal{B}$ used was set to 200 for CIFAR-10C and CIFAR-100C, and 64 for ImageNet-C. Throughout, the prompt template is fixed to ``{\fontfamily{cmss}\selectfont a photo of a $<$CLS$>$.}".

\noindent \textbf{TENT \citep{wang2020tent}}: We follow all the hyperparameters that TENT provides in their official implementation \footnote{\href{https://github.com/DequanWang/tent}{https://github.com/DequanWang/tent}}. To update the vision encoder, we use Adam as the optimizer with a learning rate of 10$^{-3}$ for CIFAR-10C and CIFAR-100C. For ImageNet-C, we update using SGD with a learning rate of 25$\times$10$^{-5}$.

\noindent \textbf{RoTTA \citep{yuan2023robust}}: For fairness, the batch sizes are set to 200 for the CIFAR datasets and 64 for ImageNet-C. The vision encoder is updated based on the Adam rule with a learning rate of 10$^{-3}$. The capacity of the memory bank is set to 64, for all the datasets. Following the notations in the paper, $\alpha$ = 0.05, $\delta$ = 0.1, $\nu$ = 0.001, $\lambda_t$ and $\lambda_u$ = 1.0. We implement the details exactly as described in their main paper.

\noindent \textbf{RPL \citep{rusak2021if}}: We use an Adam optimizer with a learning rate of 10$^{-3}$ for CIFAR-10C and CIFAR-100C. For ImageNet-C, the update rule is SGD with a learning rate of 5$\times$10$^{-4}$. To compute the generalized cross-entropy loss, \textit{q} is set to 0.8 for all the datasets. 

\noindent \textbf{SAR \citep{niu2023towards}}: The training details/hyperparameters for SAR are the same as RPL \citep{rusak2021if} for CIFAR-10 and CIFAR-100. For ImageNet-C, the learning rate is set to 25$\times$10$^{-5}$ with an SGD update rule. The entropy threshold \textit{$E_{0}$} is 0.4xln(C), where C is the number of classes. $\rho$ is set to a default of 0.05. The moving average factor is 0.9 for $e_m$ and $e_0$ is set to 0.2. All parameters are the same as in \citep{niu2023towards}.

\noindent \textbf{TPT \citep{shu2022test}}: For each test image, 63 augmentations are generated based on random resized crops, yielding a batch of 64 images, in addition to the original test image. The prompt/context vectors are initialized based on ``{\fontfamily{cmss}\selectfont a photo of a $<$CLS$>$.}" and tokenized using pre-trained CLIP weights. The confidence threshold is set to 10\% \textit{i.e.,} the marginal entropy over the 10\% confident samples is minimized. For all the datasets, we follow their core implementation and optimize the prompt vectors using an AdamW optimizer with a learning rate of 5$\times$10$^{-3}$.

\noindent \textbf{VTE \citep{dobler2024lost}}: In VTE, an ensemble of different prompt templates is considered based on the idea of \cite{radford2021learning}. An example of templates includes ``{\fontfamily{cmss}\selectfont a photo of a $<$CLS$>$.}", ``{\fontfamily{cmss}\selectfont a sketch of a $<$CLS$>$.}", ``{\fontfamily{cmss}\selectfont a painting of a $<$CLS$>$.}", etc. The prompt templates are then averaged. On the vision side, similar to TPT \citep{shu2022test}, a batch of random augmentation is created for a test image with no model updates.

\noindent \textbf{WATT-P* and WATT-S* \cite{osowiechi2024watt}}: The two original variants of WATT \cite{osowiechi2024watt} were proposed with weight-averaging of adapted weights from multiple prompt templates. Additionally, for each test batch, adaptation was performed over multiple iterations. To fit our online TTA scheme, we reduced the number of iterations to a single step for each prompt template and reset CLIP parameters only after a domain. However, this is still not fully online, as \textit{training} is performed on the test batch using 8 selected prompt templates: ``{\fontfamily{cmss}\selectfont a photo of a $<$CLS$>$}", ``{\fontfamily{cmss}\selectfont itap of a $<$CLS$>$}", ``{\fontfamily{cmss}\selectfont a bad photo of the $<$CLS$>$}", ``{\fontfamily{cmss}\selectfont a origami $<$CLS$>$}", ``{\fontfamily{cmss}\selectfont a photo of the large $<$CLS$>$}", ``{\fontfamily{cmss}\selectfont a $<$CLS$>$ in a video game}", ``{\fontfamily{cmss}\selectfont art of the $<$CLS$>$}", and ``{\fontfamily{cmss}\selectfont a photo of the small $<$CLS$>$}". As they report performance on CIFAR-10C and CIFAR-100C only, we follow their original implementation details. For experiments using WATT-S*, we set a batch size of 200 (for a fair comparison to other baselines) and a learning rate of 10$^{-3}$ using an Adam optimizer. For WATT-P*, a learning rate of 10$^{-4}$ is used with the same batch size.

\noindent \textbf{Stat$\textit{A}$ \cite{zanella2025realistic}}: We adopt the original hyperparameter settings of Stat$\textit{A}$ for online TTA in our reported ImageNet experiments and apply them to ImageNet-C. To control the effective number of correlated classes in a test batch, we set $\gamma$ to 0.1 (low correlation) and -1 (separate, sequential). The default prompt template is ``{\fontfamily{cmss}\selectfont a photo of a $<$CLS$>$}."

\subsubsection{Experiments on Domain Generalization Datasets}

We evaluate $\framework$ on standard domain generalization datasets \cite{gulrajani2020search}. For a fair comparison, as reported in WATT \cite{osowiechi2024watt}, we use a batch size of 128 for all the online TTA experiments on VLCS, PACS, and Office Home. We use an AdamW optimizer for model updates using $\framework$ with learning rates of 5$\times$10$^{-4}$, 10$^{-3}$, and 5$\times$10$^{-3}$ for OfficeHome, PACS, and VLCS, respectively. We use the same learning settings for TENT, WATT-S*, and WATT-P*, as in WATT \cite{osowiechi2024watt}. In addition to the mentioned datasets, we also run experiments on the Terra Incognita dataset and optimize using an AdamW optimizer with a learning rate of 5$\times$10$^{-4}$ and a batch size of 256.

\begin{figure}[t!]
\small
\centering
\setlength{\tabcolsep}{5pt}
\begin{tabular}{c c}
\includegraphics[width=\columnwidth, trim = 0mm 2mm 0mm 0mm, clip]{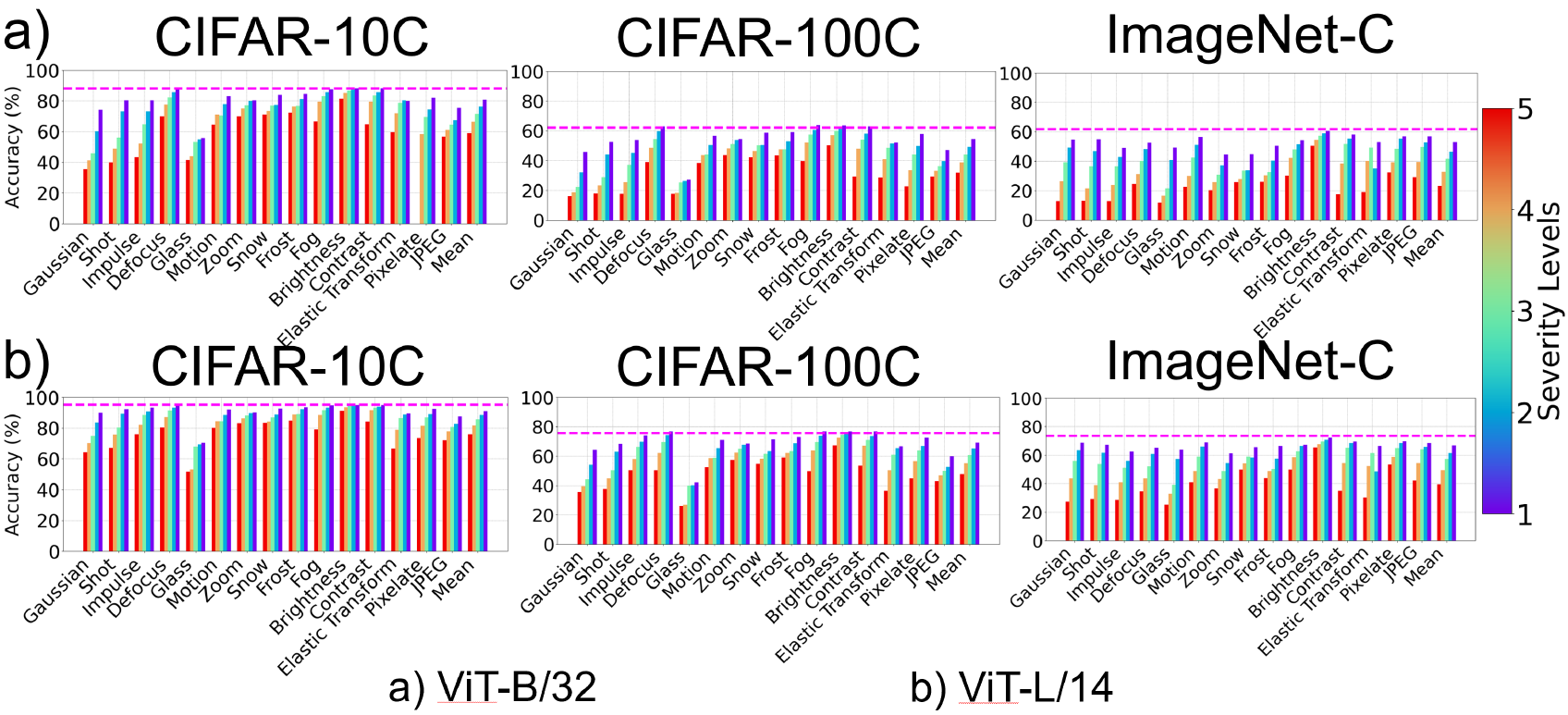}
\end{tabular}
\caption{Task-wise mean accuracy (\%) of zero-shot CLIP across different corruption severity levels. [Top]: ViT-B/32 backbone. [Bottom]: ViT-L/14 backbone. The \textcolor{magenta}{\textbf{dashed lines}} indicate the performance of zero-shot CLIP (w/ respective visual backbones) on the corresponding source datasets.}
\label{fig:add_severity}
\end{figure}

\subsection{Additional Results}\label{add_results}
In the following subsections, we provide additional results and discussions.

\subsubsection{Zero-shot performance analysis of ViT-B/32 and ViT-L/14} \label{severity_backbones}

In the main paper, we analyze and evaluate the zero-shot performance of ResNet-101 (RN101) \citep{he2016deep} and ViT-B/16 \citep{dosovitskiy2020image} and conclude that such CLIP backbones are extremely sensitive, in terms of classification accuracy, to increasing severity levels of image corruption. This could be a major concern in situations involving real-time deployment of CLIP.
Here, we present a similar analysis in Figure \ref{fig:add_severity}, using ViT-B/32 and ViT-L/14 as backbones. Our analysis, from the main paper, carries forward. To summarise, irrespective of the CLIP visual backbone, the robustness towards image corruption is limited. The classification performance degrades with an increase in the severity of corruption in an image.

\begin{table*}[htb!]
    \centering
    \resizebox{\textwidth}{!}{
    \begin{tabular}{cc|c|ccccccccccccccc|c} \\ \toprule
    & Method & Venue & \rotatebox[origin=c]{70}{Gaussian} & \rotatebox[origin=c]{70}{Shot} & \rotatebox[origin=c]{70}{Impulse} & \rotatebox[origin=c]{70}{Defocus} & \rotatebox[origin=c]{70}{Glass} & \rotatebox[origin=c]{70}{Motion} & \rotatebox[origin=c]{70}{Zoom} & \rotatebox[origin=c]{70}{Snow} & \rotatebox[origin=c]{70}{Frost} & \rotatebox[origin=c]{70}{Fog} & \rotatebox[origin=c]{70}{Brightness} & \rotatebox[origin=c]{70}{Contrast} & \rotatebox[origin=c]{70}{Elastic} & \rotatebox[origin=c]{70}{Pixelate} & \rotatebox[origin=c]{70}{JPEG} & Mean  \\ \midrule
    \multirow{9}{*}{\rotatebox[origin=c]{90}{\Large{CIFAR-10C}}}
&\begin{tabular}[c]{@{}c@{}}{ZS}\end{tabular} & ICLR'21 & 
  35.47& 39.94 & 43.23 & 69.95 & 41.43 & 64.50& 70.13& 70.85& 72.33& 66.66& 81.37& 64.57&  59.69& 48.28& 56.62& 59.00\\ 
& \begin{tabular}[c]{@{}c@{}}TENT\end{tabular} &  ICLR'21 &
  20.09 & 23.45 & 34.47 & 69.85 & 23.01 & 39.79 & 60.35 & 76.83 & 77.49 & 76.07 & \textbf{88.88} & \textbf{81.38} & 65.35 & 57.01 & 51.19 & 56.35 \\ 
& \begin{tabular}[c]{@{}c@{}}RoTTA\end{tabular} & CVPR'23 & 
  36.55 & 40.91 & 43.99 & 70.03 & 42.45 & 64.52 & 70.08 & 71.23& 72.68 & 67.31 & 81.92 & 64.99 &60.33 & 49.40 & 57.11 & 59.57 \\
& \begin{tabular}[c]{@{}c@{}}RPL\end{tabular} & arXiv & 
  15.89 & 19.08 & 34.04 & \textbf{77.84} & 18.72 & 41.22 & 62.39 & 78.17 & 78.86 & \textbf{76.31} & 88.83 & 81.15 & 68.98 & 54.19 & 51.91 & 56.51 \\ 
& \begin{tabular}[c]{@{}c@{}}SAR\end{tabular} & ICML'22 &
  50.28 & 54.12 & 49.65 & 73.08 & 51.98 & 71.17 & 74.65 & 73.73 & 75.22 & 70.99 & 84.25 & 72.08 & 63.93 & 51.57 & 60.32 & 65.13 \\ \cdashline{2-19}
&\begin{tabular}[c]{@{}c@{}}TPT \end{tabular} &  NeurIPS'22 &
   43.11  & 46.53 & 48.29& 71.31 & 47.80 & 66.89& 71.96&  74.00&  76.00& 68.81& 84.12& 66.35& 63.86&  51.86& 58.01& 62.59\\ 
&\begin{tabular}[c]{@{}c@{}}VTE\end{tabular} & ECCV-W'24 &
  47.59 & 50.18 & \textbf{53.15} & 71.39 & 53.86 & 67.92&  72.90&  76.37& 76.30& 70.78& 83.27&  61.07& 69.00 & 58.57 & 61.14& 64.90\\ 
&\begin{tabular}[c]{@{}c@{}}WATT-P*\end{tabular} & NeurIPS'24 &
43.64 & 47.1 & 45.97 & 74.98 & 48.04 & 70.42 & 74.74 & 76.1 & 76.86 & 71.85 & 85.15 & 70.36 & 64.6 & 56.0 & 60.37 & 64.41 \\
&\begin{tabular}[c]{@{}c@{}}WATT-S*\end{tabular} & NeurIPS'24 &
\textbf{56.38} & \textbf{58.08} & 52.48 & 78.07 & \textbf{58.29} & \textbf{76.42} & \textbf{79.16} & \textbf{78.9 }& \textbf{79.71} & 76.76 & 87.19 & 76.49 & \textbf{70.35} & \textbf{64.11} & \textbf{65.05} & \textbf{70.49} \\

&\cellcolor{light_gray}\begin{tabular}[c]{@{}c@{}}Ours\end{tabular} & & 
 \cellcolor{light_gray}52.39 & \cellcolor{light_gray}55.99 & \cellcolor{light_gray}52.54&  \cellcolor{light_gray}76.79&  \cellcolor{light_gray}54.04&  \cellcolor{light_gray}74.90&  \cellcolor{light_gray}75.79& \cellcolor{light_gray} 77.67 & \cellcolor{light_gray}79.10 & \cellcolor{light_gray}75.31 & \cellcolor{light_gray}86.33&  \cellcolor{light_gray}77.34&  \cellcolor{light_gray}67.41&  \cellcolor{light_gray}57.06& \cellcolor{light_gray}61.29 & \cellcolor{light_gray}68.26\\

\midrule
 \multirow{9}{*}{\rotatebox[origin=c]{90}{\Large{CIFAR-100C}}}
&\begin{tabular}[c]{@{}c@{}}{ZS}\end{tabular} & ICLR'21 &
  16.23 & 17.83 & 17.57 & 39.07 & 17.63 & 38.55& 43.81&  42.32&  43.46&39.71&  50.32&29.34& 28.74& 22.85& 29.42&  31.79\\ 
& \begin{tabular}[c]{@{}c@{}}TENT\end{tabular} & ICLR'21 &
  5.53 & 7.64 & 6.85 & \textbf{49.60} & 4.47 & 48.45 & 52.35 & 49.77 & 26.77 & 37.50 & 63.05 & \textbf{50.53} & 13.89 & 27.00 & 30.80 & 31.61 \\ 
& \begin{tabular}[c]{@{}c@{}}RoTTA\end{tabular} & CVPR'23 &
  16.63 & 18.25 & 17.78 & 38.62 & 17.76 & 38.38 & 43.52 & 42.39 & 43.41 & 39.37 & 50.60 & 28.85 & 28.89 & 23.50 & 29.65 & 31.84 \\
& \begin{tabular}[c]{@{}c@{}}RPL\end{tabular} & arXiv &
  4.50 & 5.80 & 9.61 & 50.26 & 4.43 & \textbf{48.88} & \textbf{52.61} & \textbf{50.27} & 22.36 & 25.34 & \textbf{63.36} & 50.31 & 9.10 & 18.65 & \textbf{34.53} & 30.00 \\ 
& \begin{tabular}[c]{@{}c@{}}SAR\end{tabular} & ICML'22 &
  \textbf{24.63} & \textbf{27.14} & 21.25 & 44.57 & 22.98 & 43.95 & 48.40 & 48.01 & \textbf{47.76} & 44.85 & 57.76 & 42.11 & 32.69 & \textbf{28.02} & 33.08 & \textbf{37.81} \\ \cdashline{2-19}
&\begin{tabular}[c]{@{}c@{}}TPT\end{tabular} & NeurIPS'22 &
  16.08 & 17.65 & 17.54 & 39.21 & 19.47 &  38.91& 44.01&  43.45&  44.46&40.15& 50.93& 27.77& 30.91& 23.36& 29.55&  32.23\\ 
&\begin{tabular}[c]{@{}c@{}}VTE\end{tabular} & ECCV-W'24 &
  16.84 & 18.33 & 18.94 & 39.63 & 22.88 & 39.13&  43.80& 44.56&  44.88& 39.21&  49.37&28.37&  34.13&  26.87&  30.12&  33.14\\ 
&\begin{tabular}[c]{@{}c@{}}WATT-P*\end{tabular} & NeurIPS'24 &
15.55 & 17.02 & 16.16 & 40.25 & 16.16 & 38.74 & 43.6 & 42.49 & 43.51 & 39.39 & 51.17 & 31.85 & 28.35 & 23.94 & 29.74 & 31.86 \\
&\begin{tabular}[c]{@{}c@{}}WATT-S*\end{tabular} & NeurIPS'24 &
16.22 & 17.72 & 16.85 & 41.54 & 17.04 & 39.66 & 44.55 & 43.33 & 44.26 & 40.26 & 52.13 & 33.13 & 29.34 & 24.65 & 30.39 & 32.73 \\
&\cellcolor{light_gray}\begin{tabular}[c]{@{}c@{}}Ours\end{tabular} & &
  \cellcolor{light_gray}21.35&  \cellcolor{light_gray}24.71&  \cellcolor{light_gray}\textbf{22.32}&  \cellcolor{light_gray}46.26&  \cellcolor{light_gray}\textbf{23.07}&  \cellcolor{light_gray}44.64&  \cellcolor{light_gray}50.12& \cellcolor{light_gray}47.23&  \cellcolor{light_gray}46.88&  \cellcolor{light_gray}\textbf{44.92}&  \cellcolor{light_gray}58.55&  \cellcolor{light_gray}38.52&  \cellcolor{light_gray}\textbf{34.56}&  \cellcolor{light_gray}27.73& \cellcolor{light_gray}33.19 & \cellcolor{light_gray}37.60\\   

\midrule
\multirow{9}{*}{\rotatebox[origin=c]{90}{\Large{ImageNet-C}}}
&\begin{tabular}[c]{@{}c@{}}{ZS}\end{tabular} & ICLR'21 & 
  12.88 & 13.04 & 12.90 & 24.42 & 11.86 &  22.72& 20.20&  25.70& 25.84& 30.28& 50.54&17.32& 18.96& 32.20& 29.12& 23.20\\ 
& \begin{tabular}[c]{@{}c@{}}TENT\end{tabular} & ICLR'21 & 
  9.18 & 8.50 & 10.42 & \textbf{26.02} & 15.72 & 26.06 & 21.64 & 27.12 & 26.18 & 31.60 & 50.58 & 22.28 & 20.12 & 34.06 & 31.30 & 24.05 \\ 
& \begin{tabular}[c]{@{}c@{}}RoTTA\end{tabular} &CVPR'23&
  13.10 & 13.30 & 13.02 & 24.48 & 11.96 & 22.86 & 20.28 & 26.06& 26.06 & 30.24 & 50.46 & 17.34 &19.16 & 32.44 & 29.18 & 23.33 \\ 
& \begin{tabular}[c]{@{}c@{}}RPL\end{tabular} &arXiv&
  11.68 & 10.98 & 12.10 & 25.68 & 13.24 & 23.98 & 20.84 & 26.32 & 26.12 & 30.86 & 50.62 & 19.30 & 19.48 & 33.14 & 29.92 & 23.62 \\ 
& \begin{tabular}[c]{@{}c@{}}SAR\end{tabular} &ICML'22&
  \textbf{19.82} & \textbf{20.36} & \textbf{20.92} & 25.78 & 20.40 & 28.34 & 23.10 & 28.12 & \textbf{28.38} & 34.74 & 51.10 & \textbf{24.60} & 24.38 & \textbf{36.54 }& 34.40 & \textbf{28.07} \\ \cdashline{2-19}
&\begin{tabular}[c]{@{}c@{}}TPT\end{tabular} &NeurIPS'22&
  12.04 & 12.64 & 12.52 & 25.38 & 12.28 & 22.68& 20.78& 26.36&26.64&  30.78& 51.02& 16.50& 19.90& 33.62&30.62& 23.58\\ 
&\begin{tabular}[c]{@{}c@{}}VTE\end{tabular} &ECCV-W'24&
  11.96 & 12.32 & 13.44 & 25.06 & 11.70 & 22.58& 22.40& 27.38& 27.02& 32.28& \textbf{51.52} & 16.84&19.94&34.80& 32.82& 24.14\\ 
&\begin{tabular}[c]{@{}c@{}}Stat$\textit{A}$ ($\gamma$=0.1) \end{tabular} &CVPR'25&
  11.57 & 12.28 & 11.74 & 21.91 & 10.91 & 20.43 & 18.87 & 24.04& 25.0& 29.12& 49.21 & 16.02 &18.87&29.15& 26.72& 21.73\\ 
&\begin{tabular}[c]{@{}c@{}}Stat$\textit{A}$ ($\gamma$=-1)\end{tabular} &CVPR'25&
  12.52 & 13.10 & 12.74 & 22.43 & 11.39 & 21.17 & 19.61 & 24.37& 25.60& 29.44& 49.89 & 17.71 &19.21&29.87& 27.52& 22.44\\ 
& \cellcolor{light_gray}\begin{tabular}[c]{@{}c@{}}Ours\end{tabular} & &
  \cellcolor{light_gray}16.84& \cellcolor{light_gray}18.20& \cellcolor{light_gray}16.10&  \cellcolor{light_gray}25.04&  \cellcolor{light_gray}\textbf{20.90}&  \cellcolor{light_gray}\textbf{ 28.90}& \cellcolor{light_gray}\textbf{25.24}&  \cellcolor{light_gray}\textbf{29.42}&  \cellcolor{light_gray}27.18&  \cellcolor{light_gray}\textbf{36.02}& \cellcolor{light_gray}50.18&  \cellcolor{light_gray} 17.66&  \cellcolor{light_gray}\textbf{27.68}&  \cellcolor{light_gray}36.20&  \cellcolor{light_gray}\textbf{35.42}& \cellcolor{light_gray}27.39\\  

\bottomrule
\end{tabular}}
\caption{Mean accuracy (\%) on CIFAR-10C, CIFAR-100C, and ImageNet-C - TTA mean accuracy of the 15 corruptions (tasks) at a severity level of 5, using ViT-B/32.}
\label{tab:mainresultsB32}
\vspace{-15pt}
\end{table*}

\subsubsection{Online TTA results using a ViT-B/32 backbone}\label{tta_suppl}
In the main paper, we had presented the online TTA results on CIFAR-10C, CIFAR-100C, and ImageNet-C, using a ViT-B/16 backbone. In Table \ref{tab:mainresultsB32}, we provide results using a ViT-B/32 backbone. Across all the datasets, we see that $\framework$ achieves the best or comparable performance against all the baseline approaches.


\subsubsection{Detailed results from the loss ablation study}\label{loss_add_study}
In the main paper, we provide ablation of loss components and their combinations \textit{i.e.,} the mean accuracy across all the tasks for ViT-B/16 on the benchmark corruption datasets. Here, we provide additional task-wise accuracy in Table \ref{tab:add_loss}. The addition of loss components $\mathcal{L}_{pm}$ and $\mathcal{L}_{sp}$ help CLIP in adapting its feature space to a specific domain/corruption.


\begin{table*}[t!]
    \centering
    \resizebox{\textwidth}{!}{
    \begin{tabular}{cc|ccccccccccccccc|c} \\ \toprule
    & Method & \rotatebox[origin=c]{70}{Gaussian} & \rotatebox[origin=c]{70}{Shot} & \rotatebox[origin=c]{70}{Impulse} & \rotatebox[origin=c]{70}{Defocus} & \rotatebox[origin=c]{70}{Glass} & \rotatebox[origin=c]{70}{Motion} & \rotatebox[origin=c]{70}{Zoom} & \rotatebox[origin=c]{70}{Snow} & \rotatebox[origin=c]{70}{Frost} & \rotatebox[origin=c]{70}{Fog} & \rotatebox[origin=c]{70}{Brightness} & \rotatebox[origin=c]{70}{Contrast} & \rotatebox[origin=c]{70}{Elastic} & \rotatebox[origin=c]{70}{Pixelate} & \rotatebox[origin=c]{70}{JPEG} & Mean  \\ \midrule
    \multirow{9}{*}{\rotatebox[origin=c]{90}{\Large{CIFAR-10C}}}
    &\begin{tabular}[c]{@{}c@{}}ViT-B/16\end{tabular} &
\\ \cmidrule{2-2}
& \begin{tabular}[c]{@{}c@{}}$\mathcal{L}_{ent}$\end{tabular} &
  14.62 & 17.29 & 49.25 & \textbf{81.06} & 20.23 & 74.27 & 81.10 & \textbf{84.25} & 81.93 & \textbf{80.93} & \textbf{91.86} & 78.92 & 53.09 & 51.18 & 49.79 & 60.65 \\ 
& \begin{tabular}[c]{@{}c@{}}$\mathcal{L}_{pm}$\end{tabular} &
41.82 & 45.66 & 52.88 & 70.13 & 39.29 & 66.61 & 71.27 & 71.49 & 74.71 & 69.20 & 80.98 & 72.89 & 55.58 & 55.55 & 50.44 & 61.23 \\
& \begin{tabular}[c]{@{}c@{}}$\mathcal{L}_{sp}$\end{tabular} & 
62.47 & 65.43 & 63.41 & 79.96 & 52.73 & 80.02 & 81.38 & 82.35 & 83.44 & 80.46 & 88.85 & 81.22 & 67.77 & 60.52 & 67.74 & 73.16 \\
  
& \begin{tabular}[c]{@{}c@{}}$\mathcal{L}_{ent}$+$\mathcal{L}_{pm}$\end{tabular} &
  16.58 & 19.89 & 42.69 & 79.45 & 23.41 & 77.03 & 80.95 & 81.74 & 78.45 & 80.66 & 90.52 & \textbf{82.55} & 62.56 & \textbf{64.35} & 58.16 & 62.60 \\ 
  
& \begin{tabular}[c]
{@{}c@{}}$\mathcal{L}_{ent}$ + 0.1 *($\mathcal{L}_{pm}$ + $\mathcal{L}_{sp}$)\end{tabular}&
44.92 & 51.41 & 63.30 & 80.65 & 50.79 & 79.83 & 83.13 & 83.59 & 83.89 & 81.91 & 89.56 & 83.16 & 67.78 & 66.69 & 67.87 & 71.90 \\
& \begin{tabular}[c]
{@{}c@{}}$\mathcal{L}_{ent}$ + 0.5 *($\mathcal{L}_{pm}$ + $\mathcal{L}_{sp}$)\end{tabular}&
58.81 & 63.85 & 65.99 & 80.26 & 53.90 & 80.30 & 82.30 & 83.08 & 84.04 & 81.66 & 89.19 & 82.67 & 68.29 & 62.70 & 68.52 & 73.70 \\

& \begin{tabular}[c]
{@{}c@{}}$\mathcal{L}_{ent}$+$\mathcal{L}_{pm}$+$\mathcal{L}_{sp}$\end{tabular} &
  \textbf{61.13} & \textbf{64.09} & \textbf{65.76} & 80.51 & \textbf{54.96} & \textbf{80.65} & \textbf{81.94} & 83.04 & \textbf{84.19} & 80.84 & 88.95 & 82.15 & \textbf{69.16} & 62.68& \textbf{67.64} & \textbf{73.85} \\

\midrule
\multirow{9}{*}{\rotatebox[origin=c]{90}{\Large{CIFAR-100C}}}
    &\begin{tabular}[c]{@{}c@{}}ViT-B/16\end{tabular} &
\\ \cmidrule{2-2}
& \begin{tabular}[c]{@{}c@{}}$\mathcal{L}_{ent}$\end{tabular} &
  7.71 & 10.05 & 11.52 & 49.42 & 12.49 & \textbf{49.36} & 53.79 & \textbf{54.11} & 50.76 & \textbf{49.92} & \textbf{64.32} & \textbf{47.07} & 33.40 & \textbf{38.63} & \textbf{39.95} & 38.17 \\ 
& \begin{tabular}[c]{@{}c@{}}$\mathcal{L}_{pm}$\end{tabular} &
  19.88 & 24.06 & 21.26 & 45.57 & 22.66 & 43.66 & 49.37 & 46.05 & 45.96 & 43.69 & 57.76 & 37.33 & 33.66 & 25.85 & 32.65 & 36.63 \\ 
& \begin{tabular}[c]{@{}c@{}}$\mathcal{L}_{sp}$\end{tabular} &
  24.69 & 27.28 & 33.62 & 49.08 & 25.29 & 47.84 & 53.86 & 51.70 & 50.93 & 46.93 & 62.57 & 44.76 & 33.88 & 31.89 & 36.05 & 41.36 \\
& \begin{tabular}[c]{@{}c@{}}$\mathcal{L}_{ent}$+$\mathcal{L}_{pm}$\end{tabular} &
  12.26 & 12.62 & 13.14 & 48.90 & 26.22 & 48.99 & 53.10 & 53.10 & \textbf{52.43} & 49.44 & 63.36 & 46.78 & 33.27 & 37.77 & 38.36 & 39.32 \\ 
& \begin{tabular}[c]
{@{}c@{}}$\mathcal{L}_{ent}$ + 0.1 *($\mathcal{L}_{pm}$ + $\mathcal{L}_{sp}$)\end{tabular}&
 24.99 & 27.10 & 32.95 & 49.92 & 25.91 & 48.42 & 54.43 & 52.87 & 51.57 & 47.70 & 63.68 & 45.25 & 39.74 & 31.58 & 37.16 & 41.88 \\
& \begin{tabular}[c]
{@{}c@{}}$\mathcal{L}_{ent}$ + 0.5 *($\mathcal{L}_{pm}$ + $\mathcal{L}_{sp}$)\end{tabular}&
 25.24 & 27.59 & 33.41 & 50.05 & 25.73 & 48.55 & 54.44 & 52.85 & 51.80 & 47.81 & 63.75 & 45.08 & 34.63 & 31.67 & 37.17 & 41.98 \\
& \begin{tabular}[c]{@{}c@{}}$\mathcal{L}_{ent}$+$\mathcal{L}_{pm}$+$\mathcal{L}_{sp}$\end{tabular} &
  \textbf{24.91} &\textbf{ 27.73}& \textbf{33.66} & \textbf{50.11} & \textbf{26.27} & 48.49 & \textbf{54.85} & 52.35 & 51.62 & 48.38 & 63.27 & 45.21 & \textbf{34.74} & 32.38 & 37.31 & \textbf{42.09} \\

\midrule
\multirow{9}{*}{\rotatebox[origin=c]{90}{\Large{ImageNet-C}}}
    &\begin{tabular}[c]{@{}c@{}}ViT-B/16\end{tabular} &
\\ \cmidrule{2-2}
& \begin{tabular}[c]{@{}c@{}}$\mathcal{L}_{ent}$\end{tabular} &
  0.90 & 1.06 & 1.16 & \textbf{29.12} & 13.02 & \textbf{32.14} & 27.34 & 35.32 & 11.14 & 40.92 & \textbf{56.90} & 23.78 & 7.78 & 39.62 & 40.22 & 24.03 \\ 
& \begin{tabular}[c]{@{}c@{}}$\mathcal{L}_{pm}$\end{tabular} &
  10.20 & 11.40 & 10.74 & 19.56 & 15.18 & 20.06 & 19.28 & 27.66 & 29.72 & 34.10 & 53.50 & 22.66 & 13.80 & 24.38 & 30.26 & 22.83 \\ 
& \begin{tabular}[c]{@{}c@{}}$\mathcal{L}_{sp}$\end{tabular} &
19.32 & 20.98 & 19.26 & 25.90 & 21.22 & 30.06  & 28.56 & 35.22 & 31.34  &  40.36 & 55.20 & 25.64 & 23.68  & 36.90 & 37.18  & 30.05 \\
& \begin{tabular}[c]{@{}c@{}}$\mathcal{L}_{ent}$+$\mathcal{L}_{pm}$\end{tabular} &
  0.90 & 1.16 & 1.30 & 28.90 & 17.04 & 31.56 & 26.24 & 36.26 & 12.22 & \textbf{42.12} & 57.92 & \textbf{30.34} & 10.36 & \textbf{40.66} & \textbf{41.20} & 25.21 \\ 
& \begin{tabular}[c]
{@{}c@{}}$\mathcal{L}_{ent}$ + 0.1 *($\mathcal{L}_{pm}$ + $\mathcal{L}_{sp}$)\end{tabular}&
 19.48 & 21.14 & 19.16 & 26.80 & 20.70 & 29.80 & 29.32 & 35.92 & 30.68 & 41.04 & 55.80 & 25.66 & 22.50 & 37.68 & 37.92 & 30.25 \\
& \begin{tabular}[c]
{@{}c@{}}$\mathcal{L}_{ent}$ + 0.5 *($\mathcal{L}_{pm}$ + $\mathcal{L}_{sp}$)\end{tabular}&
 19.56 & 21.38 & 19.16 & 26.96 & 21.26 & 30.06 & 29.24 & 35.92 & 31.26 & 41.34 & 56.32 & 25.74 & 22.58 & 37.94 & 37.92 & 30.44 \\
& \begin{tabular}[c]{@{}c@{}}$\mathcal{L}_{ent}$+$\mathcal{L}_{pm}$+$\mathcal{L}_{sp}$\end{tabular} &
  \textbf{19.32} & \textbf{21.38} & \textbf{19.60} & 26.58 & \textbf{21.94} & 30.88 & \textbf{29.02} & \textbf{36.48} & 32.00 & 40.98 & 56.72 & 26.14 & \textbf{23.74} & 37.68 & 38.34 & \textbf{30.72} \\  
\bottomrule
\end{tabular}}
\caption{Task-wise loss ablation results (accuracy) on CIFAR-10C, CIFAR-100C, and ImageNet-C.}
\vspace{-4pt}
\label{tab:add_loss}
\end{table*}

\begin{table*}[htb!]
\centering
    \centering
    \resizebox{\textwidth}{!}{
    \begin{tabular}{cc|ccccccccccccccc|c} \\ \toprule
    & Method & \rotatebox[origin=c]{70}{Gaussian} & \rotatebox[origin=c]{70}{Shot} & \rotatebox[origin=c]{70}{Impulse} & \rotatebox[origin=c]{70}{Defocus} & \rotatebox[origin=c]{70}{Glass} & \rotatebox[origin=c]{70}{Motion} & \rotatebox[origin=c]{70}{Zoom} & \rotatebox[origin=c]{70}{Snow} & \rotatebox[origin=c]{70}{Frost} & \rotatebox[origin=c]{70}{Fog} & \rotatebox[origin=c]{70}{Brightness} & \rotatebox[origin=c]{70}{Contrast} & \rotatebox[origin=c]{70}{Elastic} & \rotatebox[origin=c]{70}{Pixelate} & \rotatebox[origin=c]{70}{JPEG} & Zero-Shot  \\ \midrule
    & \begin{tabular}[c]{@{}c@{}}\underline{ViT-B/16}\end{tabular} &&&&&&&&&&&&&&&& 90.1\\ 
& \begin{tabular}[c]{@{}c@{}}Ours\end{tabular} &
  84.51& 84.29 & 88.69 & 88.48 & 86.44 & 86.46 & 87.04 & 91.38 & 91.01 & 90.22 & 90.87 & 88.12 & 88.13 & 74.74 & 87.48 & 87.19 (mean)\\ \cmidrule(lr){2-18}

& \begin{tabular}[c]{@{}c@{}}\underline{ViT-B/32}\end{tabular} &
&&&&&&&&&&&&&&& 88.3\\ 
& \begin{tabular}[c]{@{}c@{}}Ours\end{tabular} &
  67.41& 68.28 & 84.23 & 80.50 & 75.37 & 79.75 & 78.55 & 87.67 & 86.36 & 85.83 & 90.04 & 80.89 & 81.56 & 82.74 & 82.36 & 80.77 (mean)  \\ 
\bottomrule
\end{tabular}}
\caption{Zero-shot performance on CIFAR10 (source) after adaptation of \framework\ on a task.}
\vspace{-0.2cm}
\label{tab:cifar10zs}
\end{table*}

\begin{table*}[htb!]
\centering
    \centering
    \resizebox{\textwidth}{!}{
    \begin{tabular}{cc|ccccccccccccccc|c} \\ \toprule
    & Method & \rotatebox[origin=c]{70}{Gaussian} & \rotatebox[origin=c]{70}{Shot} & \rotatebox[origin=c]{70}{Impulse} & \rotatebox[origin=c]{70}{Defocus} & \rotatebox[origin=c]{70}{Glass} & \rotatebox[origin=c]{70}{Motion} & \rotatebox[origin=c]{70}{Zoom} & \rotatebox[origin=c]{70}{Snow} & \rotatebox[origin=c]{70}{Frost} & \rotatebox[origin=c]{70}{Fog} & \rotatebox[origin=c]{70}{Brightness} & \rotatebox[origin=c]{70}{Contrast} & \rotatebox[origin=c]{70}{Elastic} & \rotatebox[origin=c]{70}{Pixelate} & \rotatebox[origin=c]{70}{JPEG} & Zero-Shot   \\ \midrule
    & \begin{tabular}[c]{@{}c@{}}\underline{ViT-B/16} \end{tabular}  & &&&&&&&&&&&&&&& 66.6\\ 
& \begin{tabular}[c]{@{}c@{}}Ours\end{tabular} &
  67.09& 67.12 & 67.08 & 70.31 & 63.70 & 66.83 & 70.12 & 70.10 & 68.19 & 69.55 & 71.05 & 66.49 & 65.13 & 59.98 & 67.60 & 67.36 (mean)\\ \cmidrule(lr){2-18}

& \begin{tabular}[c]{@{}c@{}}\underline{ViT-B/32}\end{tabular} & &&&&&&&&&&&&&&& 62.3
\\ 
& \begin{tabular}[c]{@{}c@{}}Ours\end{tabular} &
  45.99& 46.90 & 60.86 & 63.62 & 57.73 & 59.59 & 61.66 & 66.21 & 61.50 & 63.61 & 66.39 & 57.05 & 62.13 & 62.82 & 65.37 & 60.09 (mean) \\ 
\bottomrule
\end{tabular}}
\caption{Zero-shot performance on CIFAR100 (source) after adaptation of \framework\ on a task.}
\vspace{-0.2cm}
\label{tab:cifar100zs}
\end{table*}

\subsubsection{Effect of different prompt templates} In Table \label{effect_prompts}, we show results with ``relevant" prompt templates to show the independence of such prompt selection, at test-time. As seen, the performance gain over zero-shot ViT-B/32 is fairly large for all the prompt templates. Though TPT \cite{shu2022test} fine-tunes a pre-trained prompt on each image, and VTE \cite{dobler2024lost} uses an ensemble of prompts, our method is agnostic to the prompt template being used, making it favorable for real-time usage.  

In all of our prior experiments, we use a generic prompt template ``{\fontfamily{cmss}\selectfont a photo of a $<$CLS$>$.}" for all of the datasets and methods. Here, we replace this with ``relevant" prompt templates to show the independence of such a prompt selection, at test-time, and report the results in Table \ref{tab:prompt_ablation}. As seen, the performance gain over zero-shot ViT-B/32 is fairly large for all the prompt templates. Though TPT \cite{shu2022test} fine-tunes a pre-trained prompt on each test image, and VTE \cite{dobler2024lost} uses an ensemble of prompts, our method is agnostic to the prompt template being used, making it favorable for real-time deployment.  
\vspace{-5pt}
\begin{table}[htb!]
\large
\centering
    \centering
    \resizebox{\columnwidth}{!}{
    \begin{tabular}{cccc} \\ 
    \toprule
    Prompt Template & CIFAR-10C & CIFAR-100C & ImageNet-C  \\ \midrule
\begin{tabular}[c]{@{}c@{}}``{\fontfamily{cmss}\selectfont a low contrast photo of a $<$CLS$>$.}" \end{tabular} &
  68.53 (\textbf{\textcolor{blue}{+7.81}})& 37.09 (\textbf{\textcolor{blue}{+4.97}})& 27.31 (\textbf{\textcolor{blue}{+3.71}})\\ 
\begin{tabular}[c]{@{}c@{}}``{\fontfamily{cmss}\selectfont a blurry photo of a $<$CLS$>$.}" \end{tabular} &
  68.84 (\textbf{\textcolor{blue}{+10.96}})& 36.80 (\textbf{\textcolor{blue}{+5.33}})&  26.92 (\textbf{\textcolor{blue}{+3.52}})\\ 
\begin{tabular}[c]{@{}c@{}}``{\fontfamily{cmss}\selectfont a photo of a big $<$CLS$>$.}" \end{tabular} &
  67.49 (\textbf{\textcolor{blue}{+10.10}})& 35.79 (\textbf{\textcolor{blue}{+4.87}})&  25.64 (\textbf{\textcolor{blue}{+3.29}})\\ 
\bottomrule
\end{tabular}}
\caption{Prompt template selection. \textbf{\textcolor{blue}{+}} denotes the accuracy gain over zero-shot ViT-B/32.}
\label{tab:prompt_ablation}
\end{table}

\subsubsection{Impact of \textit{bimodal} adaptation}\label{bimodal} To show the effectiveness of \textit{bimodal} adaptation, we ablate $\mathcal{L}_{pm}$, which is responsible for updating the text encoder $f_{txt}$. We report the results in Table \ref{tab:bimodal_eff}. We see a drop in accuracy when $f_{txt}$ is ``frozen" i.e., when $\mathcal{L}_{pm}$ isn't used, necessitating the need for \textit{bimodal} adaptation of CLIP encoders.

\begin{table}[htb!]
\large
\centering
\vspace{-10pt}
\resizebox{0.65\columnwidth}{!}{
\begin{tabular}{cccc}
\toprule
Dataset & $f_{vis}$ update & $f_{txt}$ update & Ours \\ \midrule
CIFAR-10C
& \cmark & \xmark & 72.58 \\
& \cmark & \cmark & \textbf{73.85} \\
\midrule
CIFAR-100C
& \cmark & \xmark & 41.00 \\
& \cmark & \cmark & \textbf{42.09} \\
\midrule
ImageNet-C
& \cmark & \xmark & 29.88 \\
& \cmark & \cmark & \textbf{30.72} \\
\bottomrule
\end{tabular}}
\caption{Ablation on $\mathcal{L}_{pm}$ to demonstrate the need of \textit{bimodal} adaptation of CLIP encoders - using a ViT-B/16 backbone.}
\label{tab:bimodal_eff}
\vspace{-5pt}
\end{table}


\begin{table}[ht]
\centering
\resizebox{0.9\columnwidth}{!}{%
\begin{tabular}{c|c|cccc}
\toprule
Camera Sensor         & Condition   & ZS    & TENT  & SAR   & Ours     \\
\midrule
\multirow{2}{*}{Auto-Exposure}   & Light on    & 50.90 & 51.62 & 52.60 & \cellcolor{light_gray}{\textbf{53.82}} \\
                             & Light off   & 46.96 & 47.44 & 47.88 & \cellcolor{light_gray}{\textbf{49.32}} \\
\midrule
\multirow{2}{*}{Manual} & Light on    & 60.44 & 60.64 & 60.30 & \cellcolor{light_gray}{\textbf{61.26}} \\
                             & Light off   & 60.76 & 60.99 & 59.62 & \cellcolor{light_gray}{\textbf{61.61}} \\
\bottomrule
\end{tabular}%
}
\caption{Online TTA experiments on ImageNet-ES \cite{baek2024unexplored}. Mean accuracy (in \%).}
\label{tab:es}
\end{table}

\begin{table}[ht]
\centering
\resizebox{\columnwidth}{!}{
\begin{tabular}{lc|cccc}
\toprule
SigLIP & Clean (ImageNet) & Source & TENT & SAR & $\framework$ \\ 
\midrule
ImageNet-C & 82.00 & 35.44  & 37.58 & 39.62 & \cellcolor{light_gray}\textbf{40.10}\\ 
\bottomrule
\end{tabular}}
\caption{ Online TTA experiments on ImageNet-C with SigLIP \cite{zhai2023sigmoid}.}
\label{tab:siglip}
\end{table}

\subsubsection{$\framework$ for other vision-language models}

We employ SigLIP \cite{zhai2023sigmoid}, a recent pre-trained vision-language model with 877 million parameters, for our online TTA experiments on ImageNet-C. We use a batch size of 8 and a learning rate of $5 \times 10^{-5}$ with the AdamW optimizer. Despite the small batch size, which could impact prototypes, $\framework$ achieves better performance, as shown in Table \ref{tab:siglip}.

\subsubsection{Results on distribution shifts caused by lighting conditions, camera types, or object scales}

To evaluate $\framework$ across a wide spectrum of shifts, we extend our setup to datasets exhibiting variations due to lighting conditions, camera types, and object scales. We conduct experiments on ImageNet-ES \cite{baek2024unexplored}, which introduces significant variations in lighting and camera sensor settings (e.g., ISO, shutter speed). While more details can be found in the original work, but ImageNet-ES introduces wide variations in lighting conditions and camera sensor factors (ISO, shutter speed, etc.). We report the results in Table \ref{tab:es} using a ViT-B/16 backbone. To be noted that, ``Auto-Exposure" has 5 tasks for each condition, while ``Manual" has 27. On average, $\framework$ outperforms all the reported baselines.

\subsubsection{Post-adaptation results on source test sets}\label{post_adap}
Thanks to the natural language supervision and also due to the pre-training on large amounts of (image, text) pairs, CLIP has shown strong generalization capabilities. However, for an efficient adaptation to a downstream task, fine-tuning the full model is infeasible due to large model updates. The primary reason is the loss of useful pre-trained knowledge of CLIP, which could eventually lead to overfitting to a downstream task. However, for attention-based models, tuned for multimodal tasks, \cite{zhao2023tuning} show that tuning the \textit{LayerNorm} parameters leads to strong results. Inspired by \cite{zhao2023tuning}, in our \textit{bimodal} test-adaptation scheme, we update the \textit{LayerNorm} parameters of both CLIP encoders, to a specific corruption task, which makes it parametric-efficient. We perform a single-domain TTA or adapt CLIP, at test-time, to a single domain only and then reset the parameters. Now, with continual adaptation to a certain corruption task, it gets difficult to preserve CLIP's pre-trained knowledge since the normalization parameters begin to overfit to this domain. Then, a natural question arises -
\begin{center}
    \emph{Given that CLIP has been adapted to a specific corruption task, will the zero-shot generalization still hold back on its source test set?}
\end{center}
In this crucial experiment, we challenge our $\framework$, and evaluate its zero-shot generalization performance back on the source test set, to check the preservation of pre-trained CLIP knowledge. After the adaptation of CLIP on each corruption task, we report the adapted model's zero-shot performance on its corresponding source test set. We report results for CIFAR-10C and CIFAR-100C in Tables \ref{tab:cifar10zs} and \ref{tab:cifar100zs}, using ViT-B/16 and ViT-B/32 backbones. For all of the results, we use the prompt template ``{\fontfamily{cmss}\selectfont a photo of a $<$CLS$>$.}". As an example, for CIFAR-10C, upon adaptation of CLIP to \textit{Gaussian noise} following our approach, we report the adapted model's zero-shot accuracy on its source test set - CIFAR10 test set.

From Table \ref{tab:cifar10zs}, we observe that, on average, there is a 2.91\% drop in accuracy compared to a zero-shot evaluation using pre-trained CLIP ViT-B/16. Similarly, for ViT-B/32, we see a drop of about 7.53\% in mean accuracy. In Table \ref{tab:cifar100zs}, for CIFAR-100C using a ViT-B/16 backbone, we see an improvement of 0.76\% in mean accuracy. 

On the whole, we conclude that since the adaptation for a task happens over multiple test batches, the zero-shot performance back on the source data largely depends on the distribution of the image corruption. Overall, ViT-B/16 visual backbones preserve larger amounts of CLIP pre-trained knowledge. This proves the effectiveness of our method \framework, on average.

\subsubsection{t-SNE visualizations of CIFAR-10C and CIFAR-100C}\label{add_tsne}

\begin{figure}[ht!]
\centering
\setlength{\tabcolsep}{1pt}
\begin{tabular}{ccccc}
  { \large \textit{Gaussian}} & {\large \textit{Shot}} & {\large \textit{Impulse}} & {\large \textit{Defocus}} & 
  {\large \textit{Glass}}  \\
  
  \includegraphics[width=0.2\columnwidth]{figures/ours_vitb16_cifar10/t-SNE_gaussian_noise.pdf}   
  & \includegraphics[width=0.2\columnwidth]{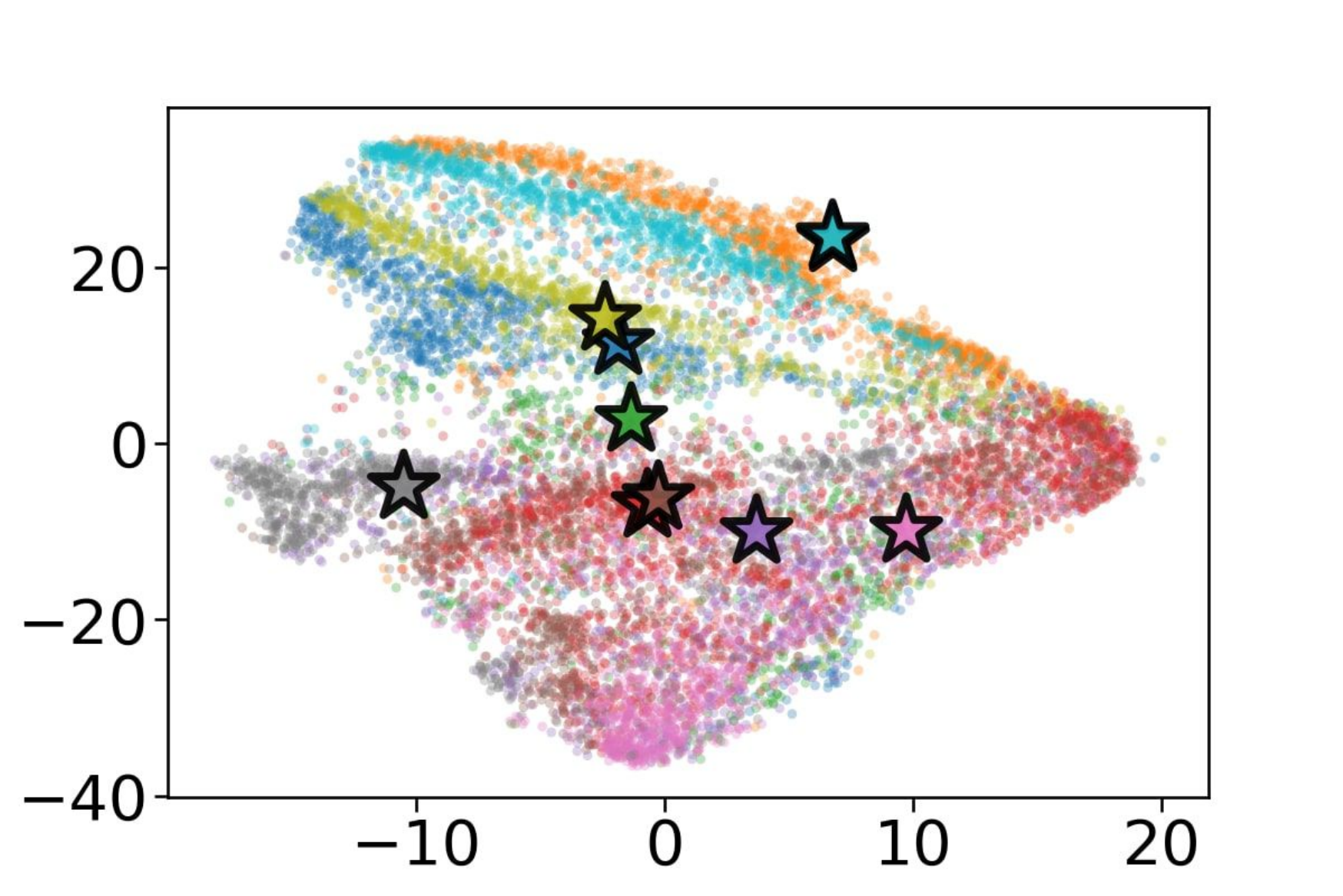}  
  & \includegraphics[width=0.2\columnwidth]{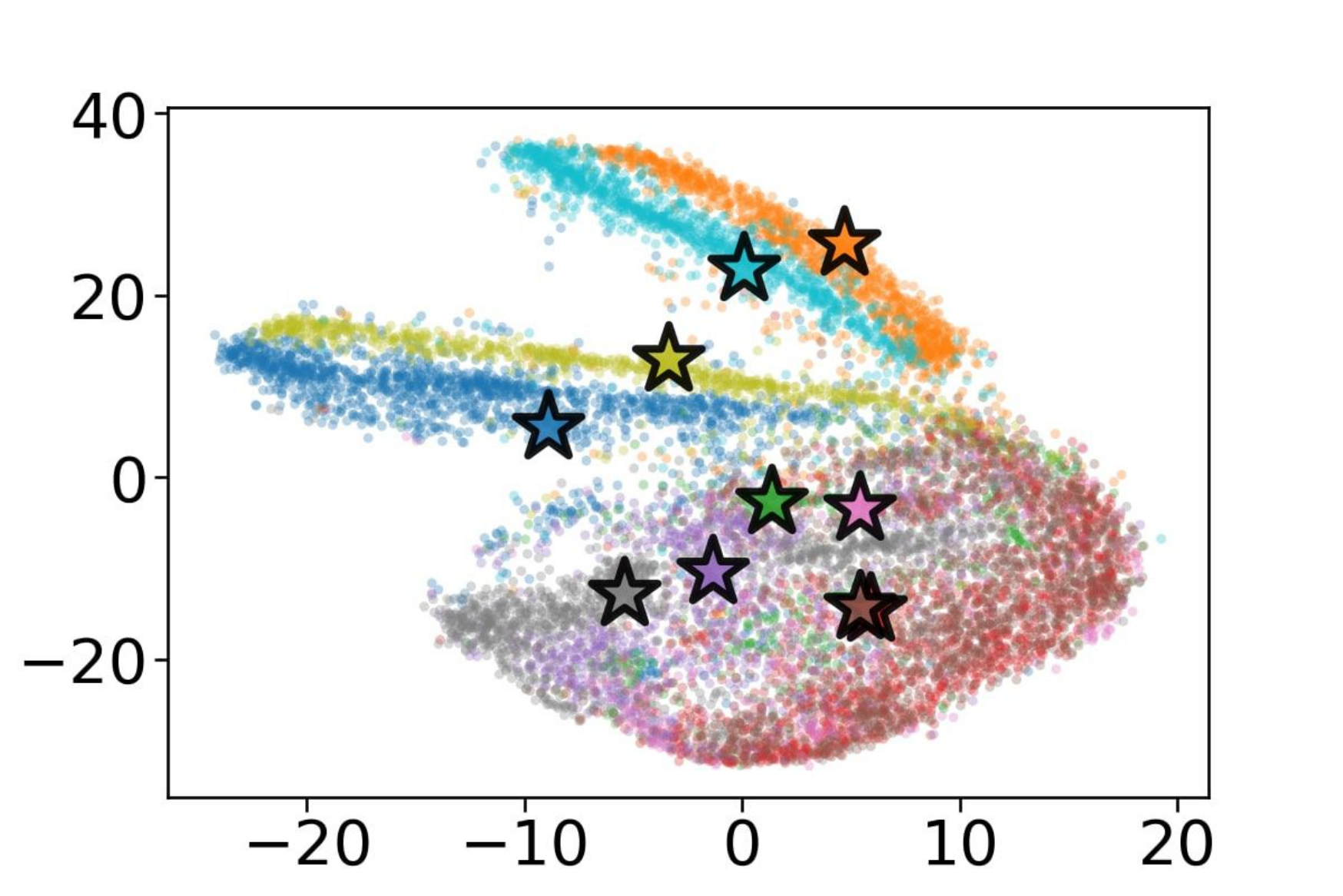}  
  & \includegraphics[width=0.2\columnwidth]{figures/ours_vitb16_cifar10/t-SNE_defocus_blur.pdf}
  & \includegraphics[width=0.2\columnwidth]{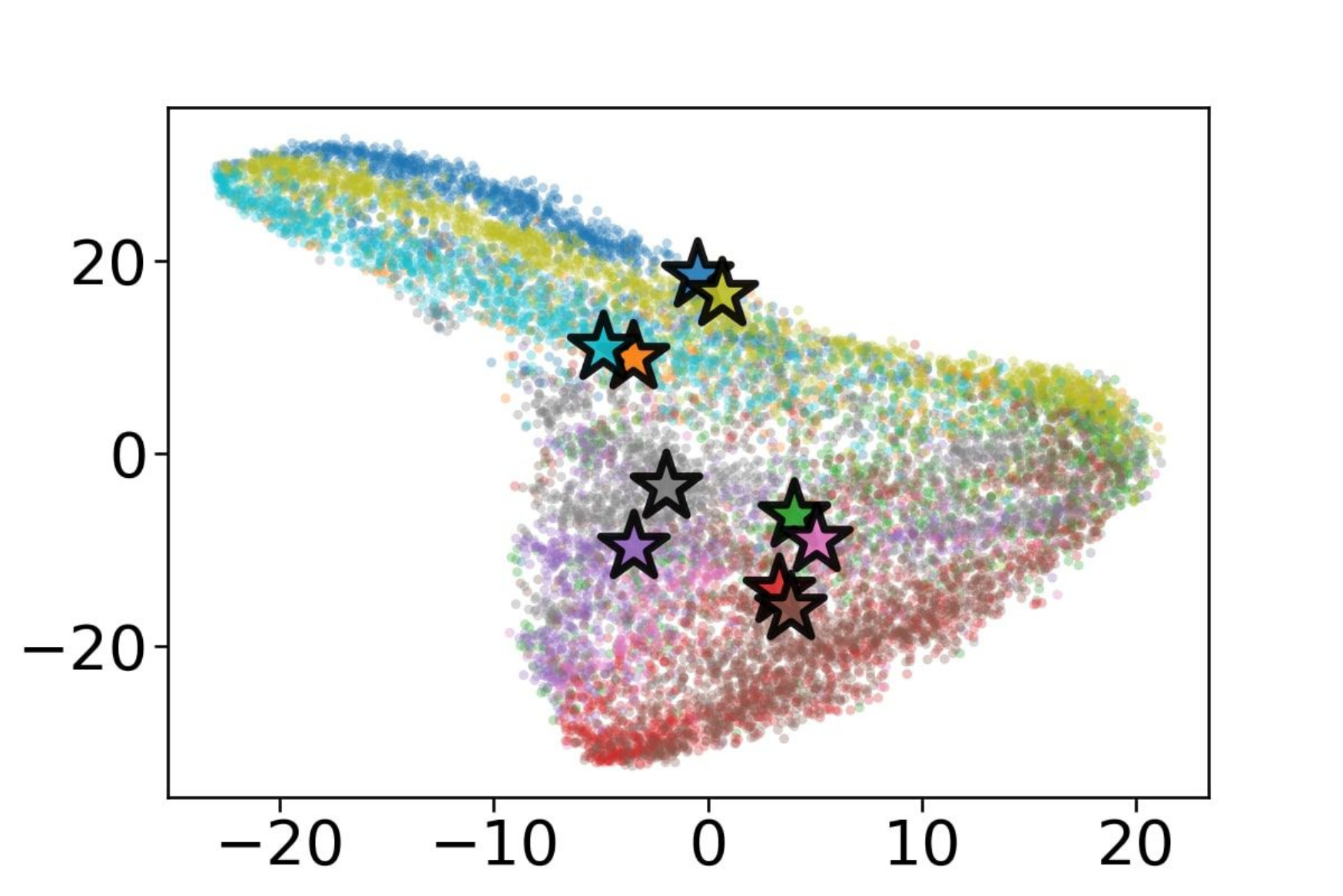}\\
  
  { \large \textit{Motion}} & {\large \textit{Zoom}} & {\large \textit{Snow}} & {\large \textit{Frost}} & 
  {\large \textit{Fog}} \\
  \includegraphics[width=0.2\columnwidth]{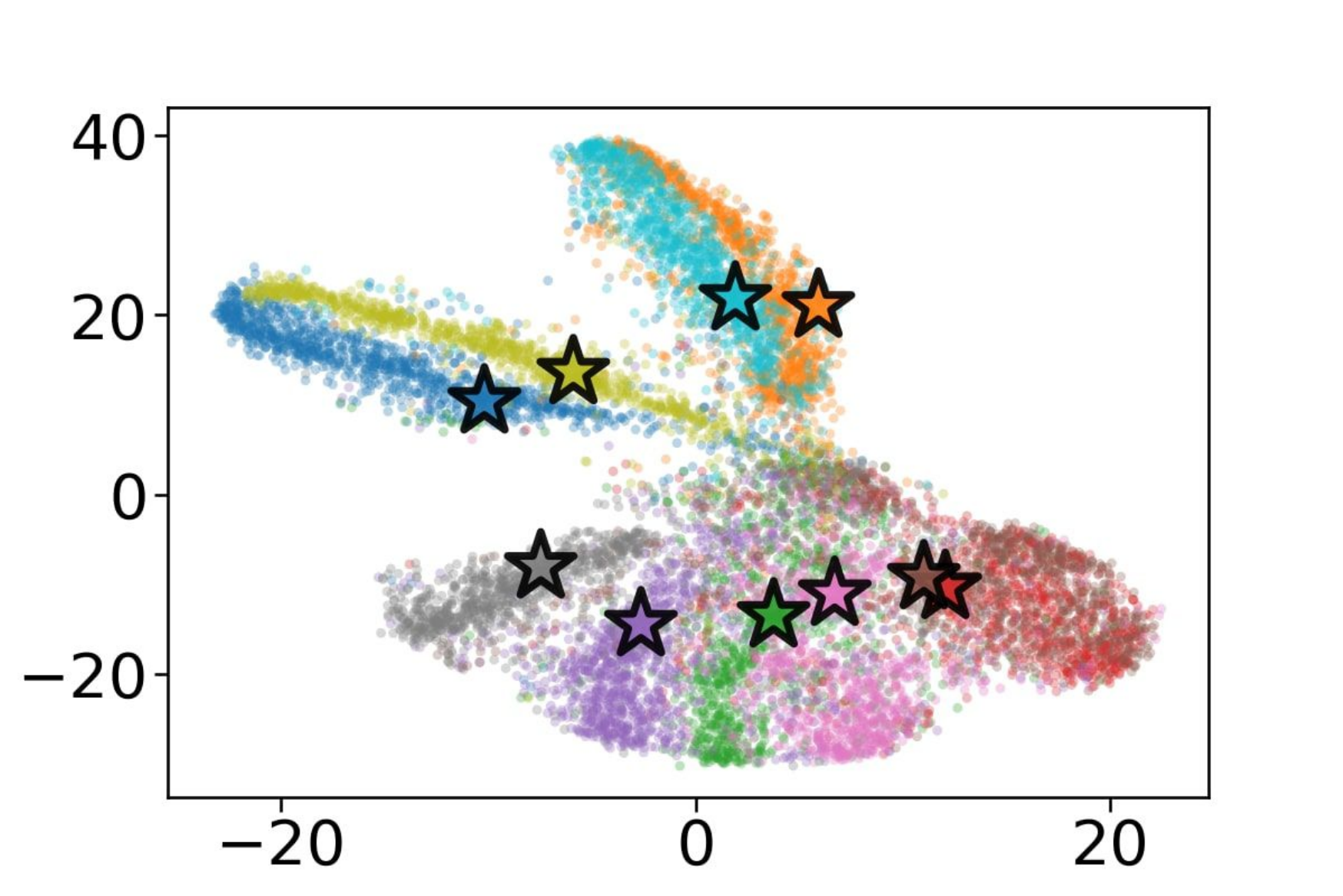} 
  & \includegraphics[width=0.2\columnwidth]{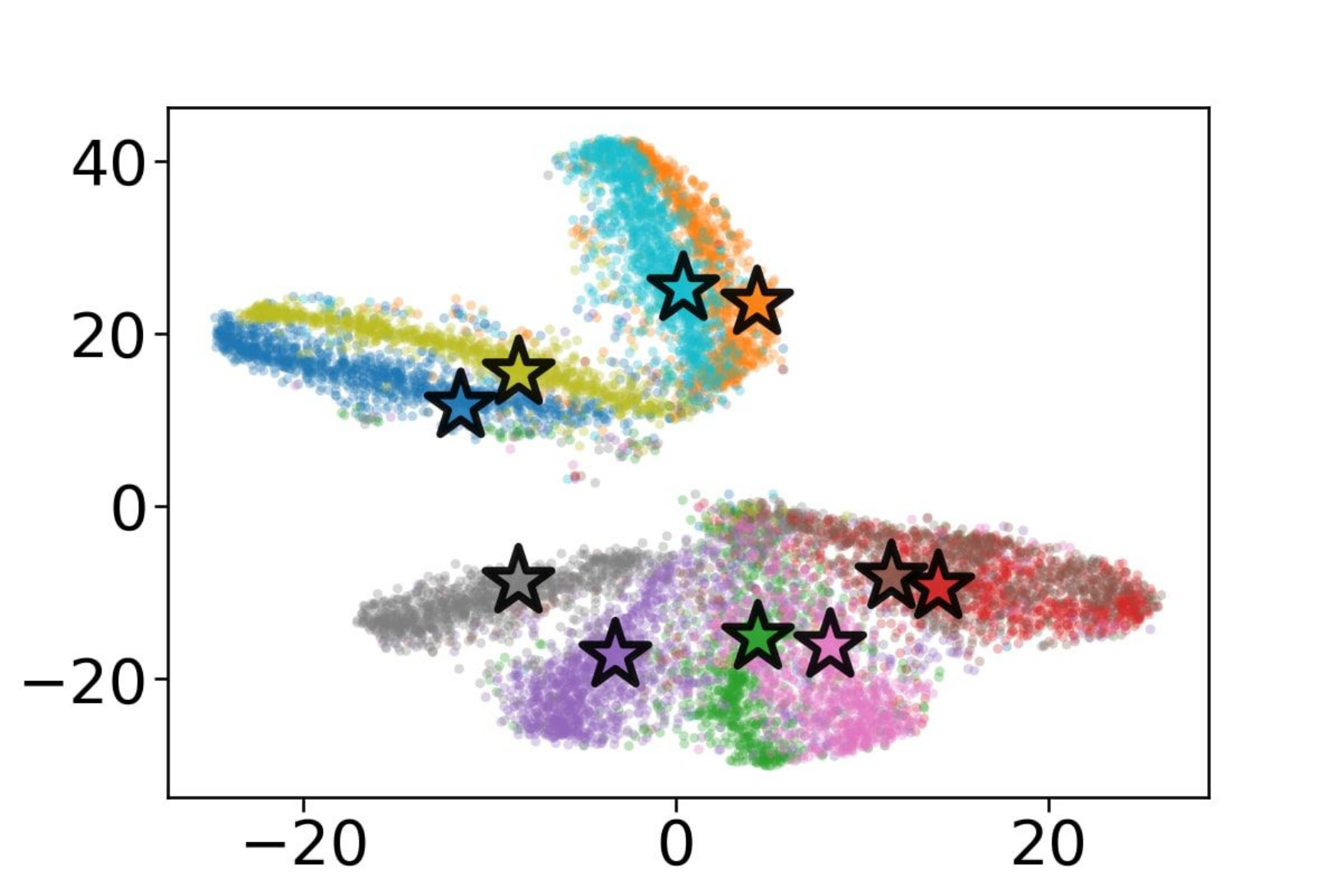}
  & \includegraphics[width=0.2\columnwidth]{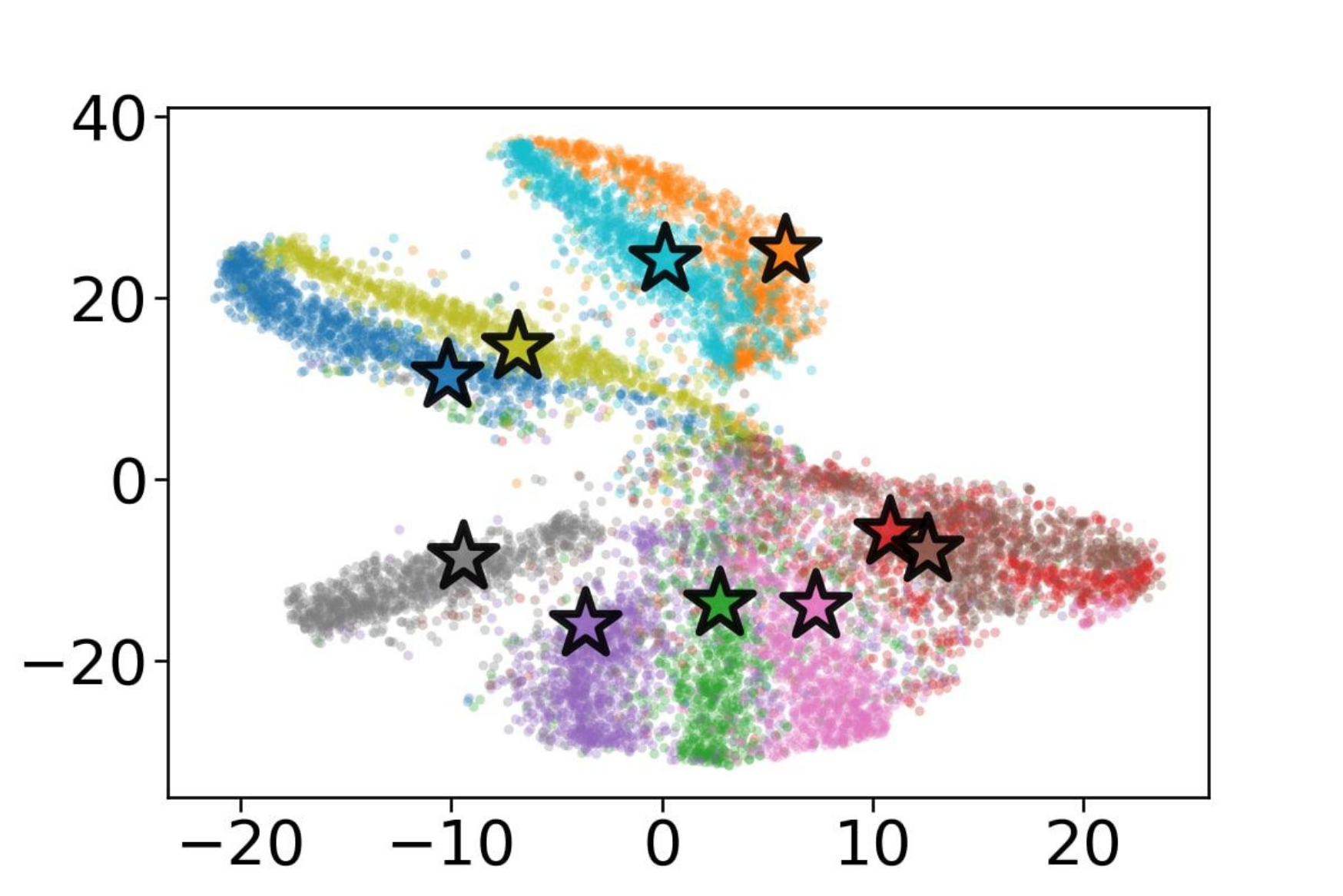}
  & \includegraphics[width=0.2\columnwidth]{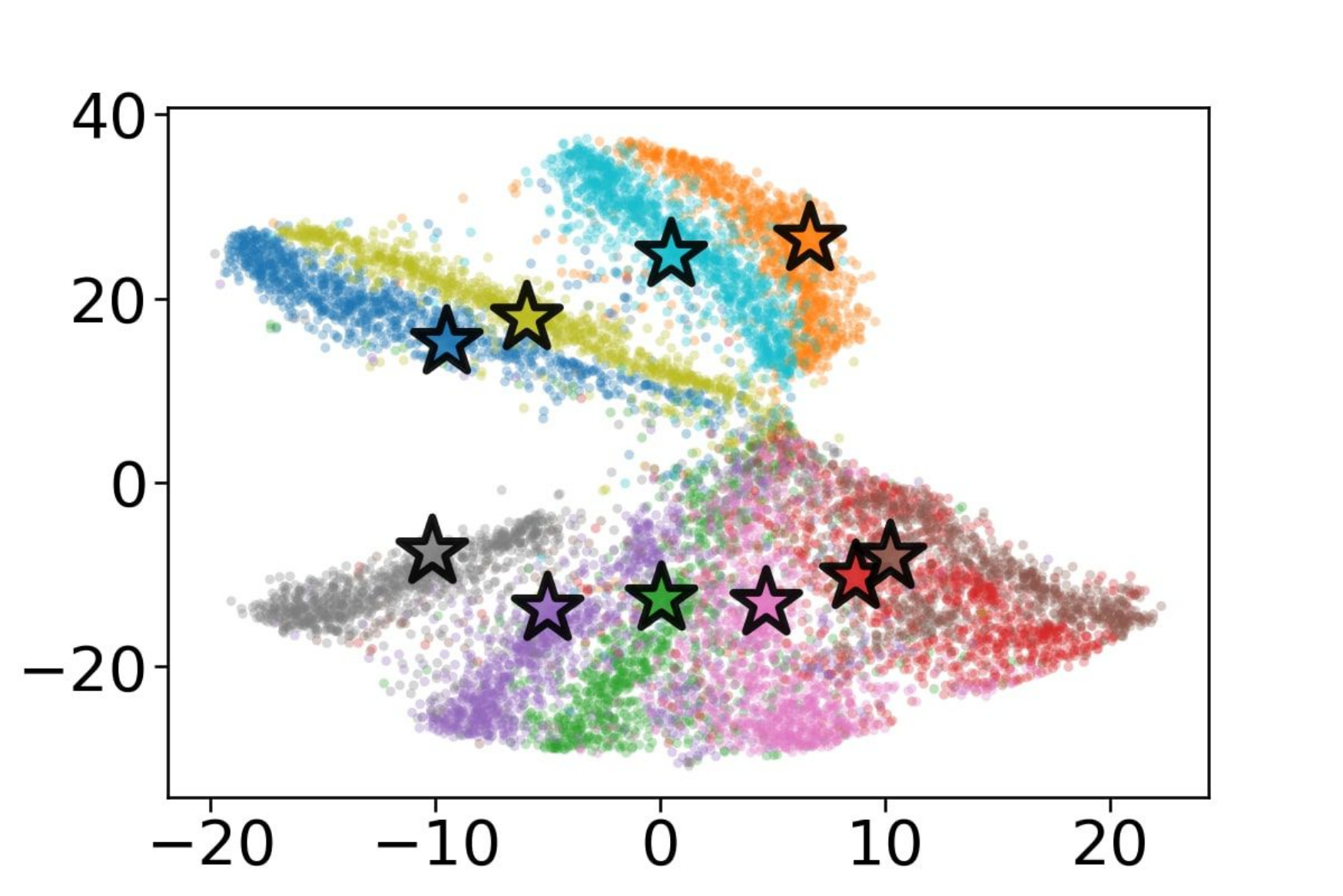}
  & \includegraphics[width=0.2\columnwidth]{figures/ours_vitb16_cifar10/t-SNE_fog.pdf}\\

  { \large \textit{Brightness}} & {\large \textit{Contrast}} & {\large \textit{Elastic}} & {\large \textit{Pixelate}} & 
  {\large \textit{JPEG}} \\
  \includegraphics[width=0.2\columnwidth]{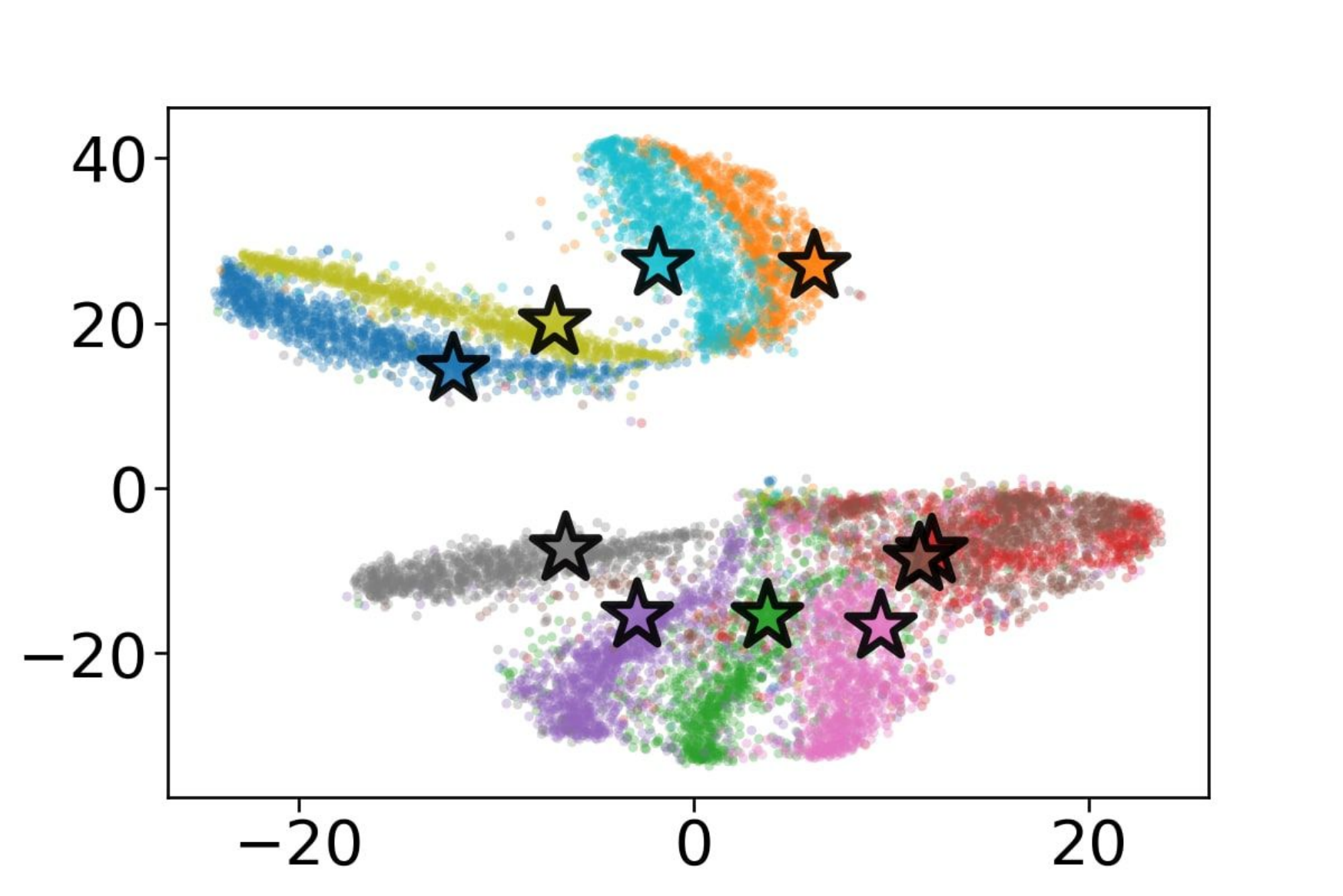} 
  & \includegraphics[width=0.2\columnwidth]{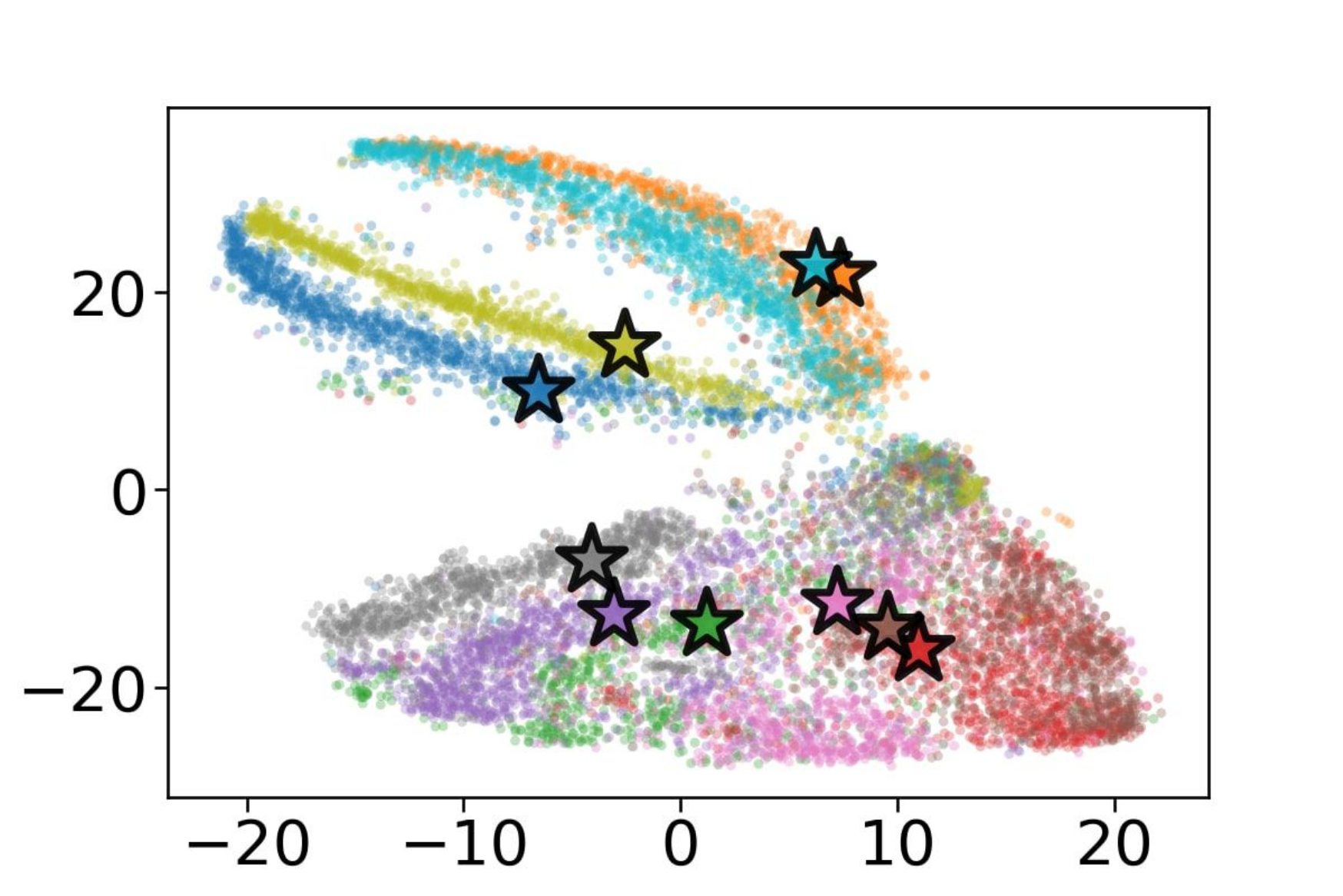}
  & \includegraphics[width=0.2\columnwidth]{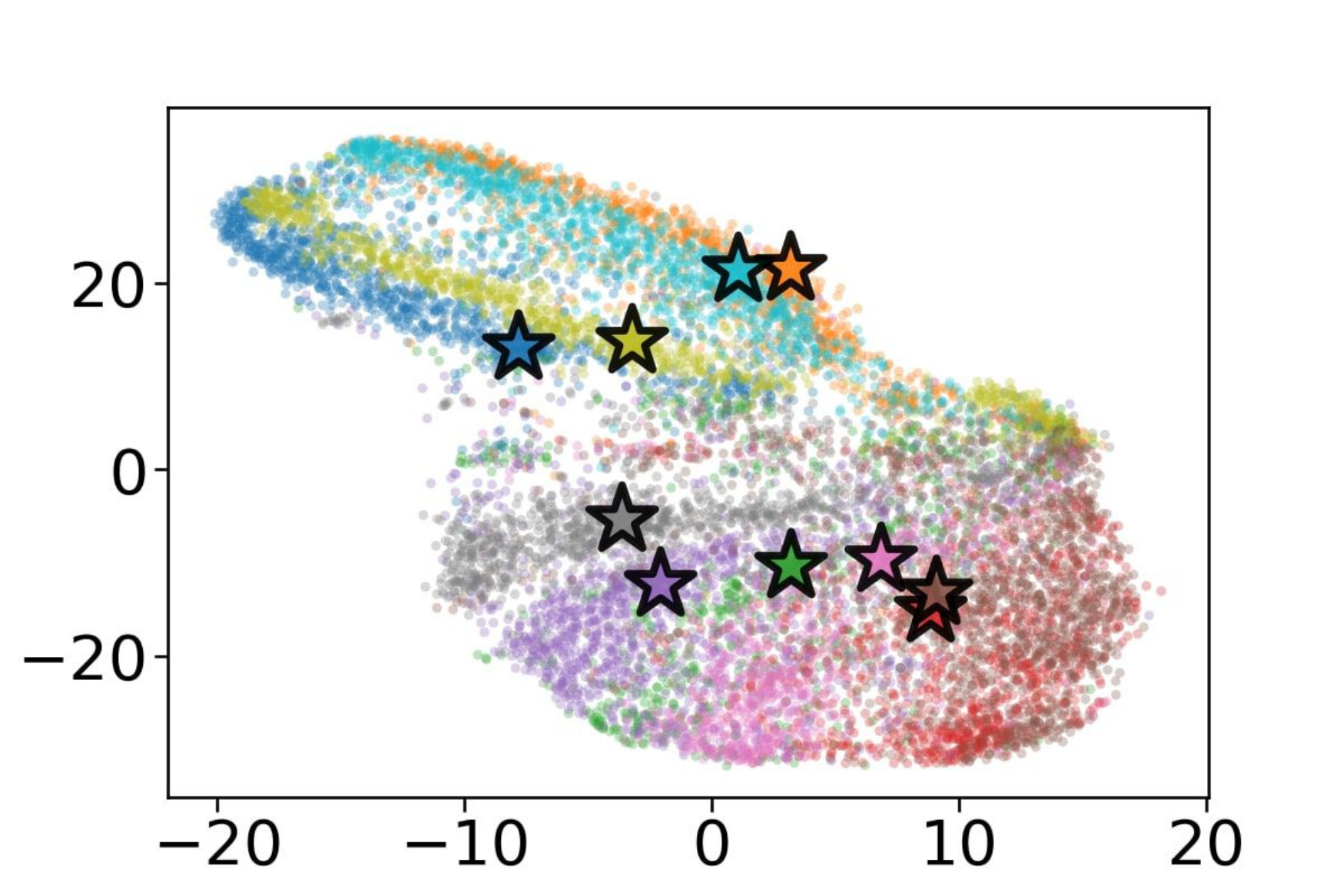}
  & \includegraphics[width=0.2\columnwidth]{figures/ours_vitb16_cifar10/t-SNE_pixelate.pdf}
  & \includegraphics[width=0.2\columnwidth]{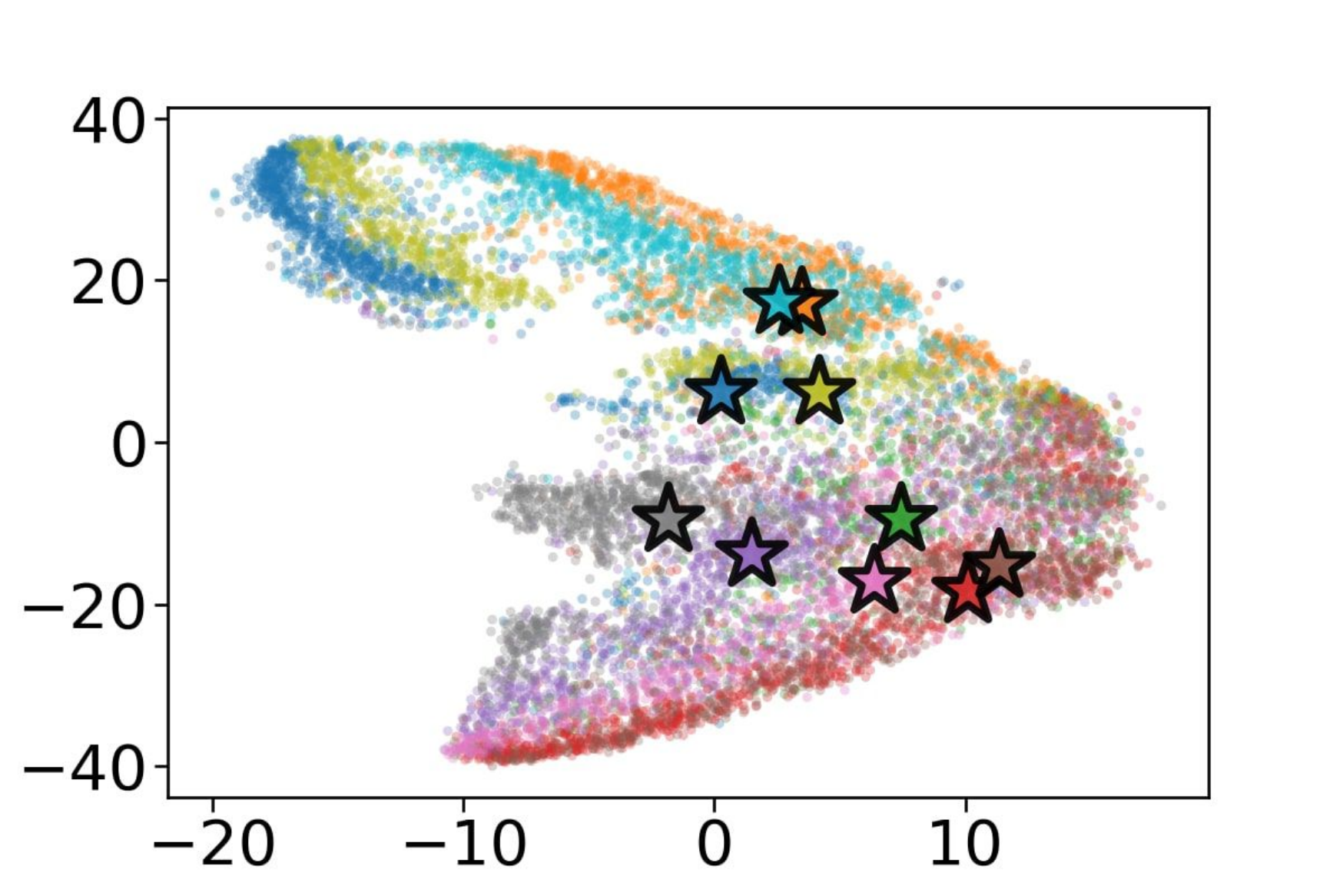}\\
  
\end{tabular}
\caption{\framework\ (w/ ViT-B/16): The t-SNE plots show visual (\(\circ\)) and text ($\bigstar$) features for CIFAR-10C. }
\label{fig: tsne_cifar10_extra}
\end{figure}

\begin{figure}[ht!]
\centering
\setlength{\tabcolsep}{1pt}
\begin{tabular}{ccccc}
  { \large \textit{Gaussian}} & {\large \textit{Shot}} & {\large \textit{Impulse}} & {\large \textit{Defocus}} & 
  {\large \textit{Glass}}  \\
  
  \includegraphics[width=0.2\columnwidth]{figures/source_vitb16_cifar10/t-SNE_gaussian_noise.pdf}   
  & \includegraphics[width=0.2\columnwidth]{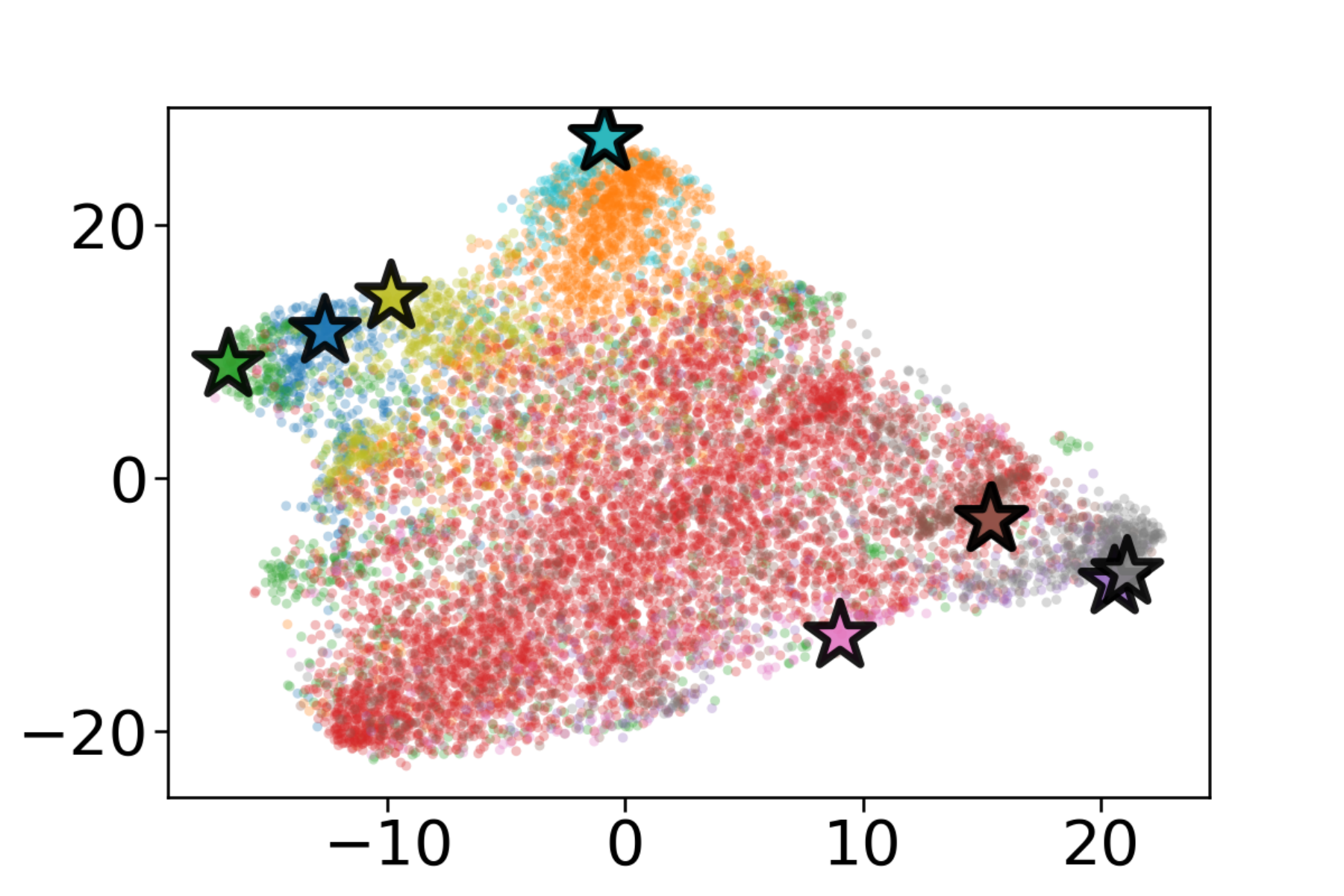}  
  & \includegraphics[width=0.2\columnwidth]{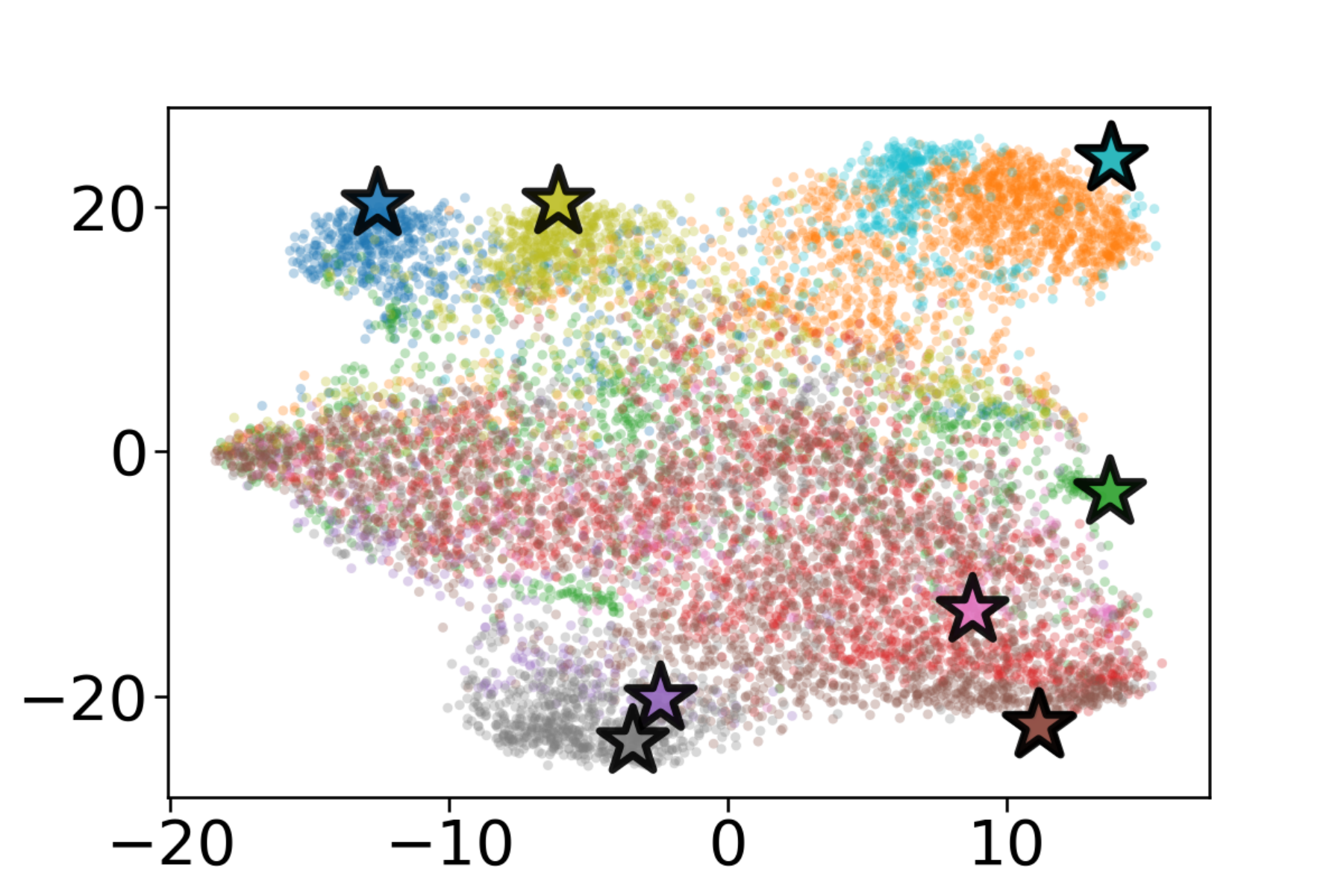}  
  & \includegraphics[width=0.2\columnwidth]{figures/source_vitb16_cifar10/t-SNE_defocus_blur.pdf}
  & \includegraphics[width=0.2\columnwidth]{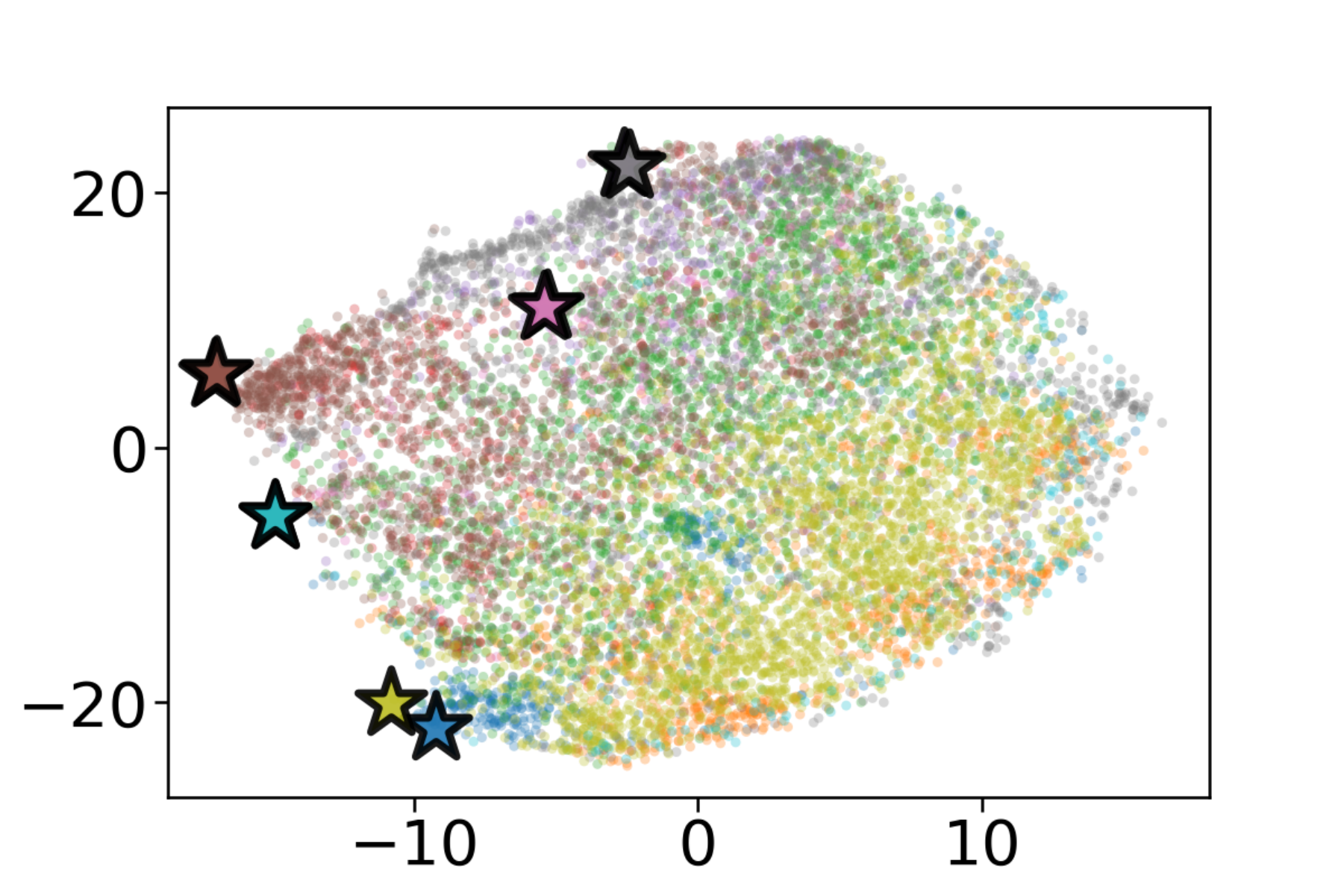}\\
  
  { \large \textit{Motion}} & {\large \textit{Zoom}} & {\large \textit{Snow}} & {\large \textit{Frost}} & 
  {\large \textit{Fog}} \\
  \includegraphics[width=0.2\columnwidth]{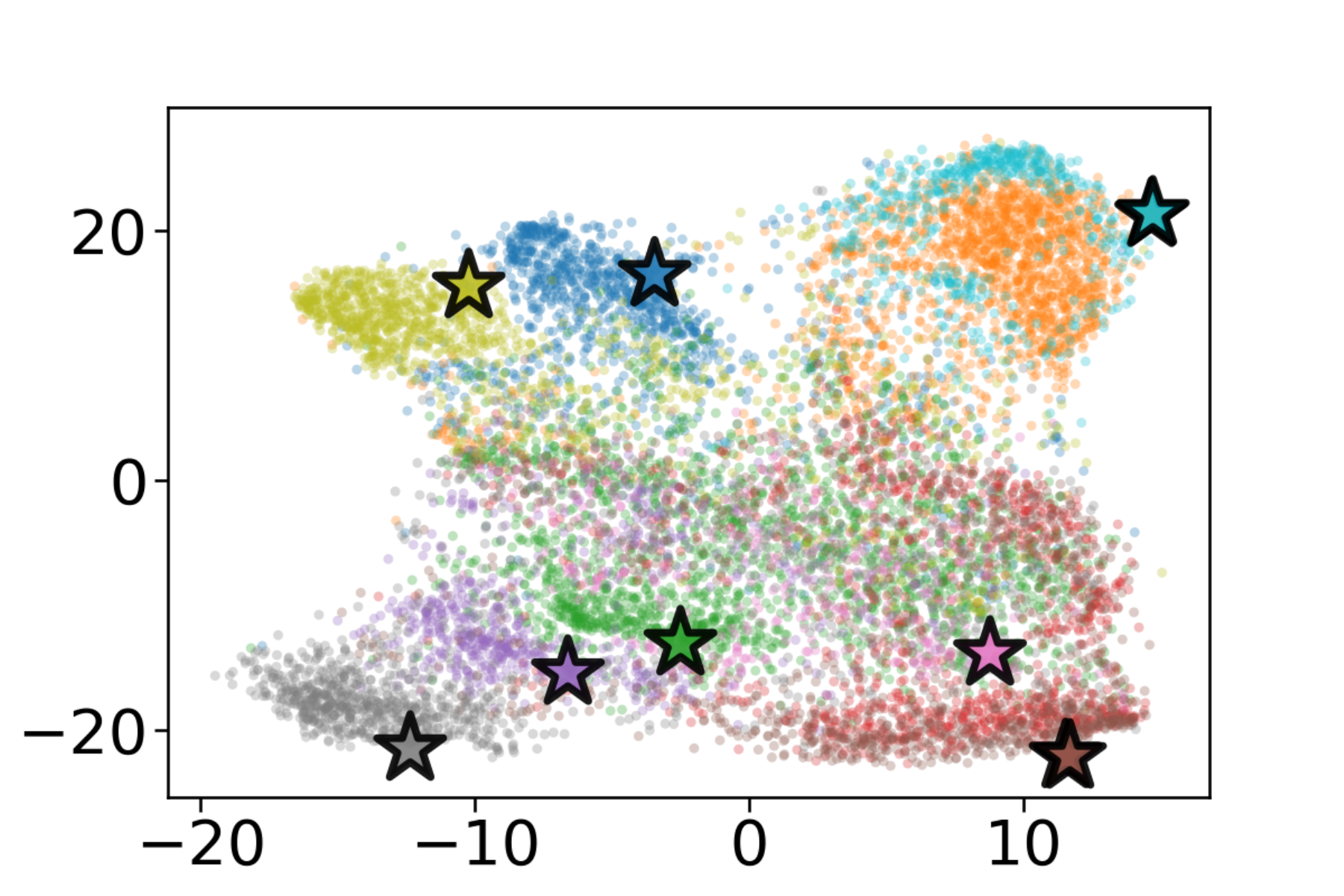} 
  & \includegraphics[width=0.2\columnwidth]{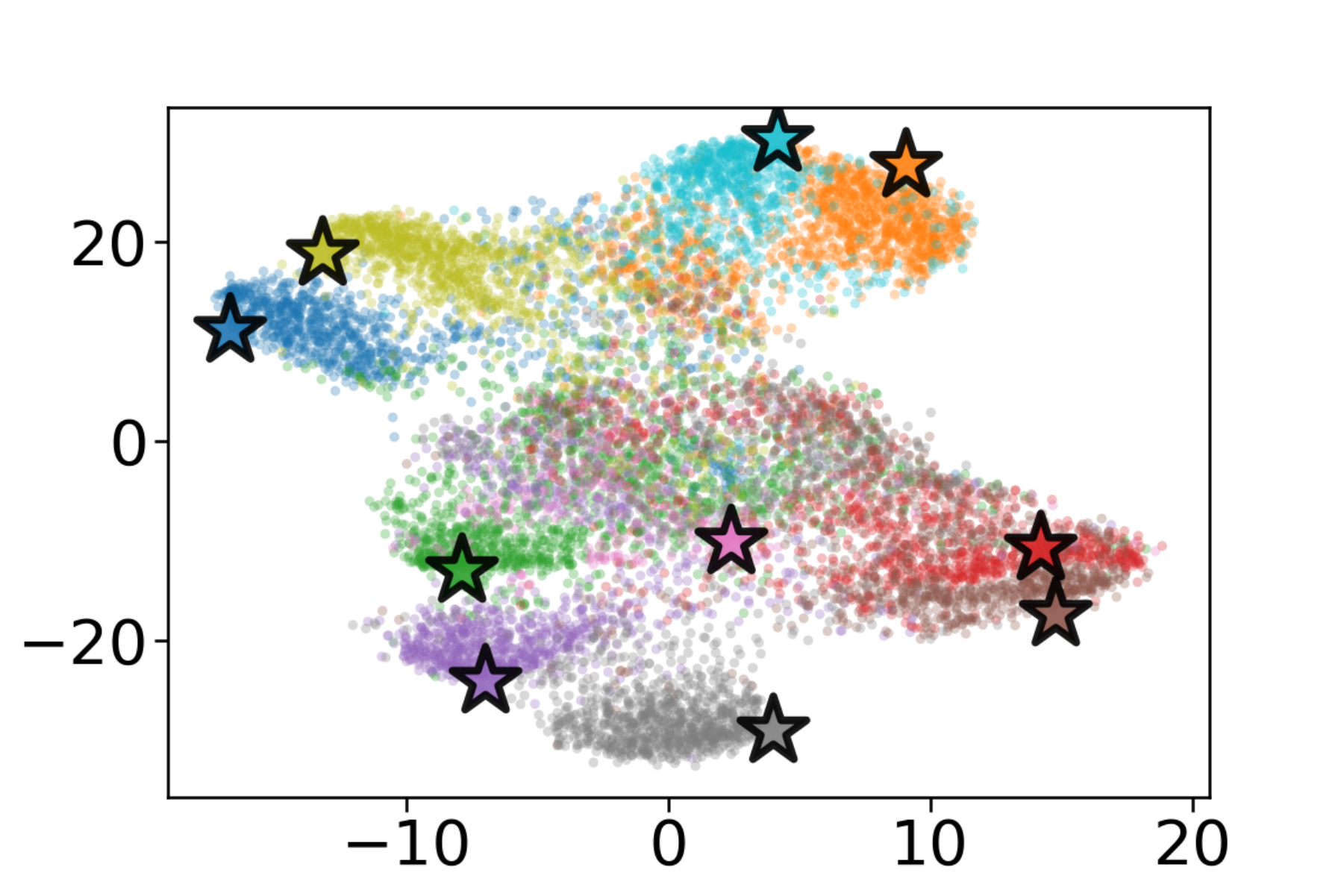}
  & \includegraphics[width=0.2\columnwidth]{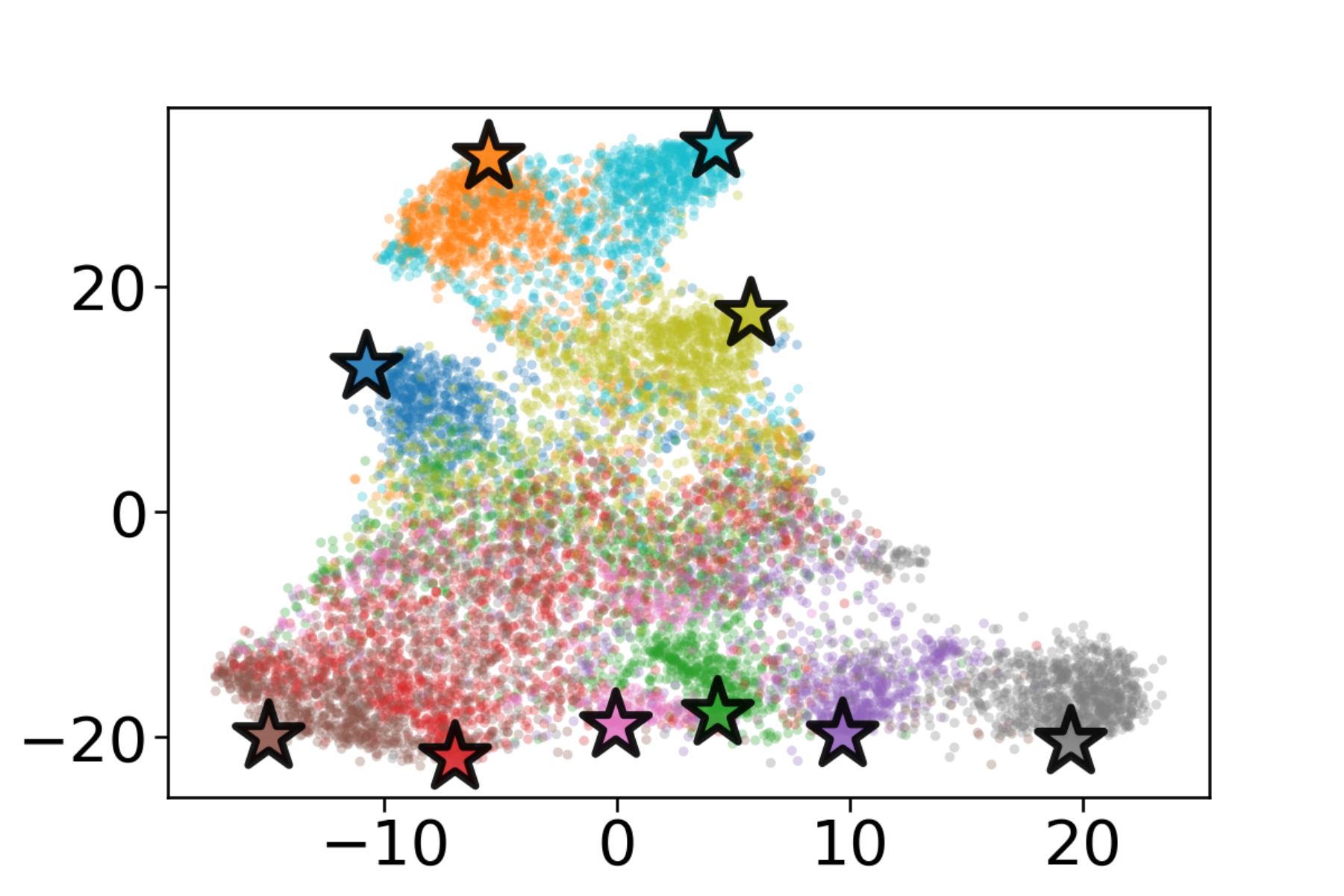}
  & \includegraphics[width=0.2\columnwidth]{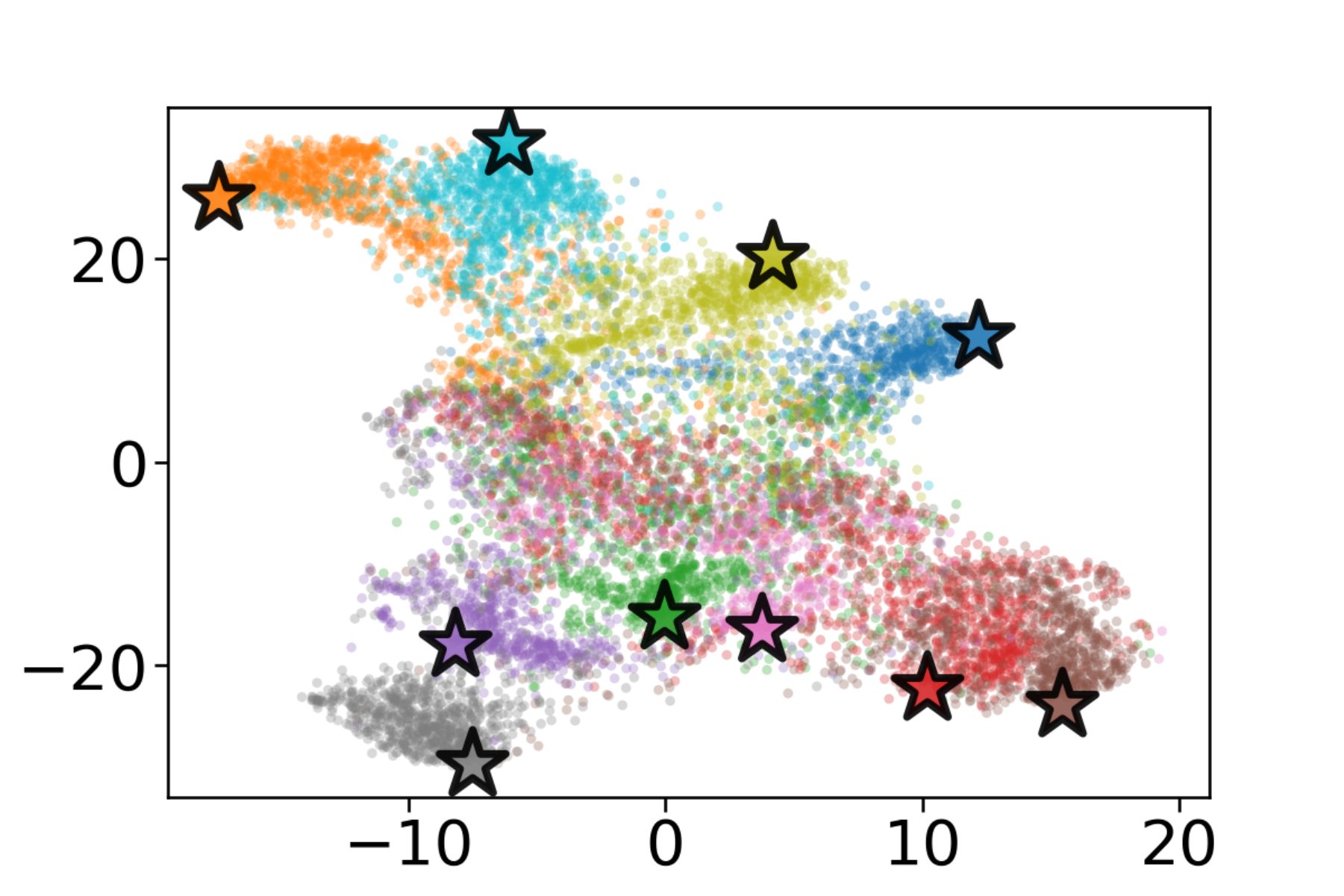}
  & \includegraphics[width=0.2\columnwidth]{figures/source_vitb16_cifar10/t-SNE_fog.pdf}\\

  { \large \textit{Brightness}} & {\large \textit{Contrast}} & {\large \textit{Elastic}} & {\large \textit{Pixelate}} & 
  {\large \textit{JPEG}} \\
  \includegraphics[width=0.2\columnwidth]{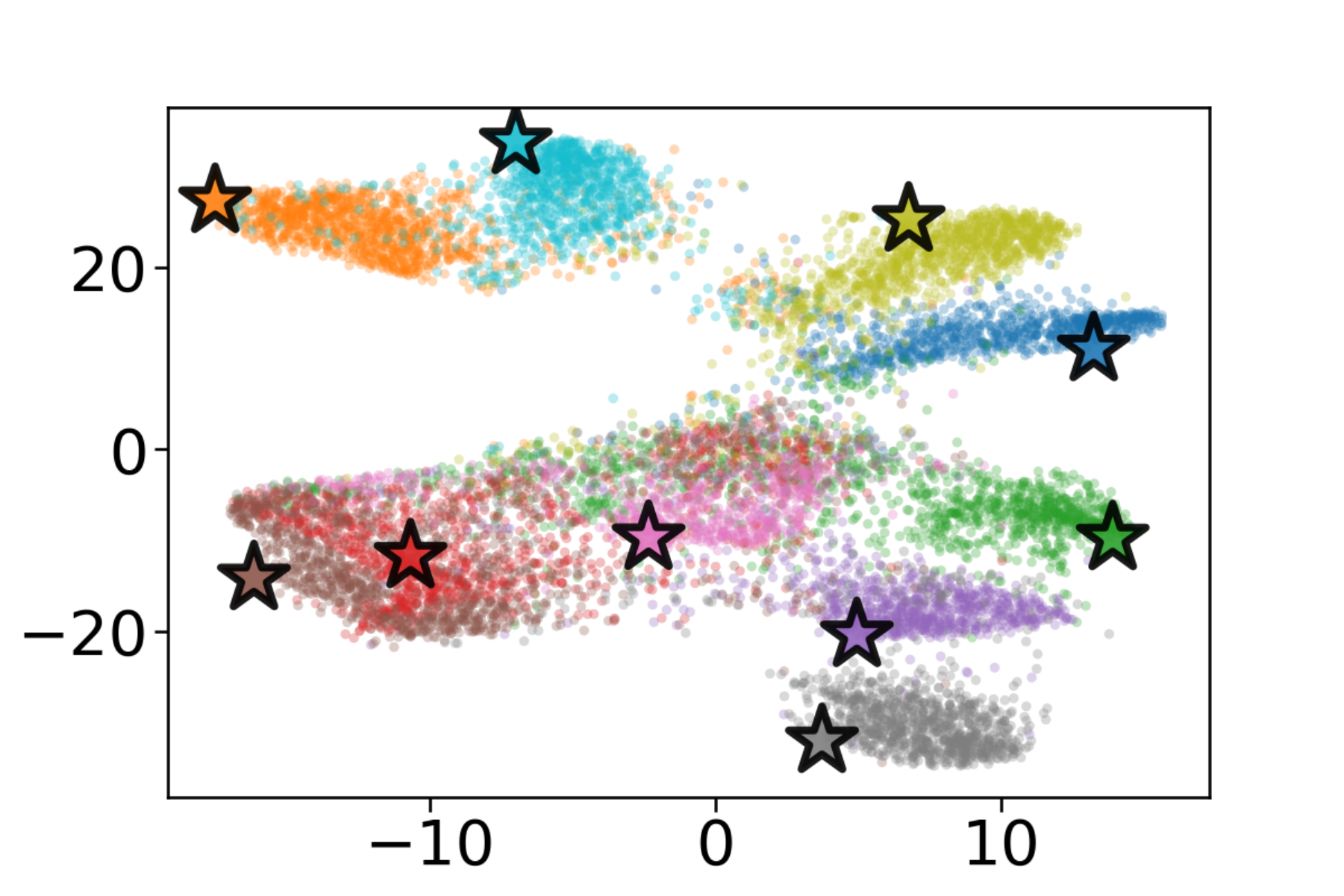} 
  & \includegraphics[width=0.2\columnwidth]{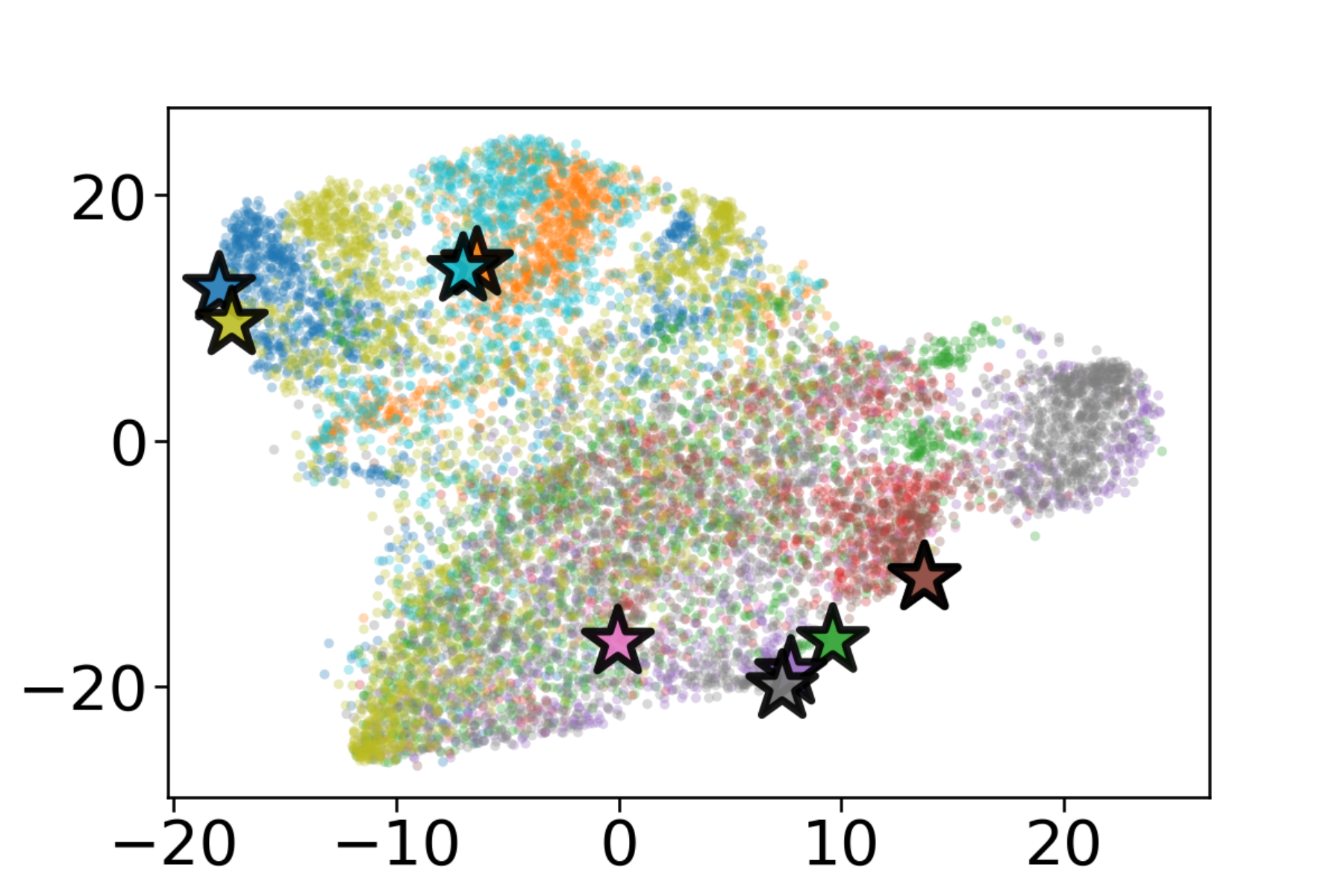}
  & \includegraphics[width=0.2\columnwidth]{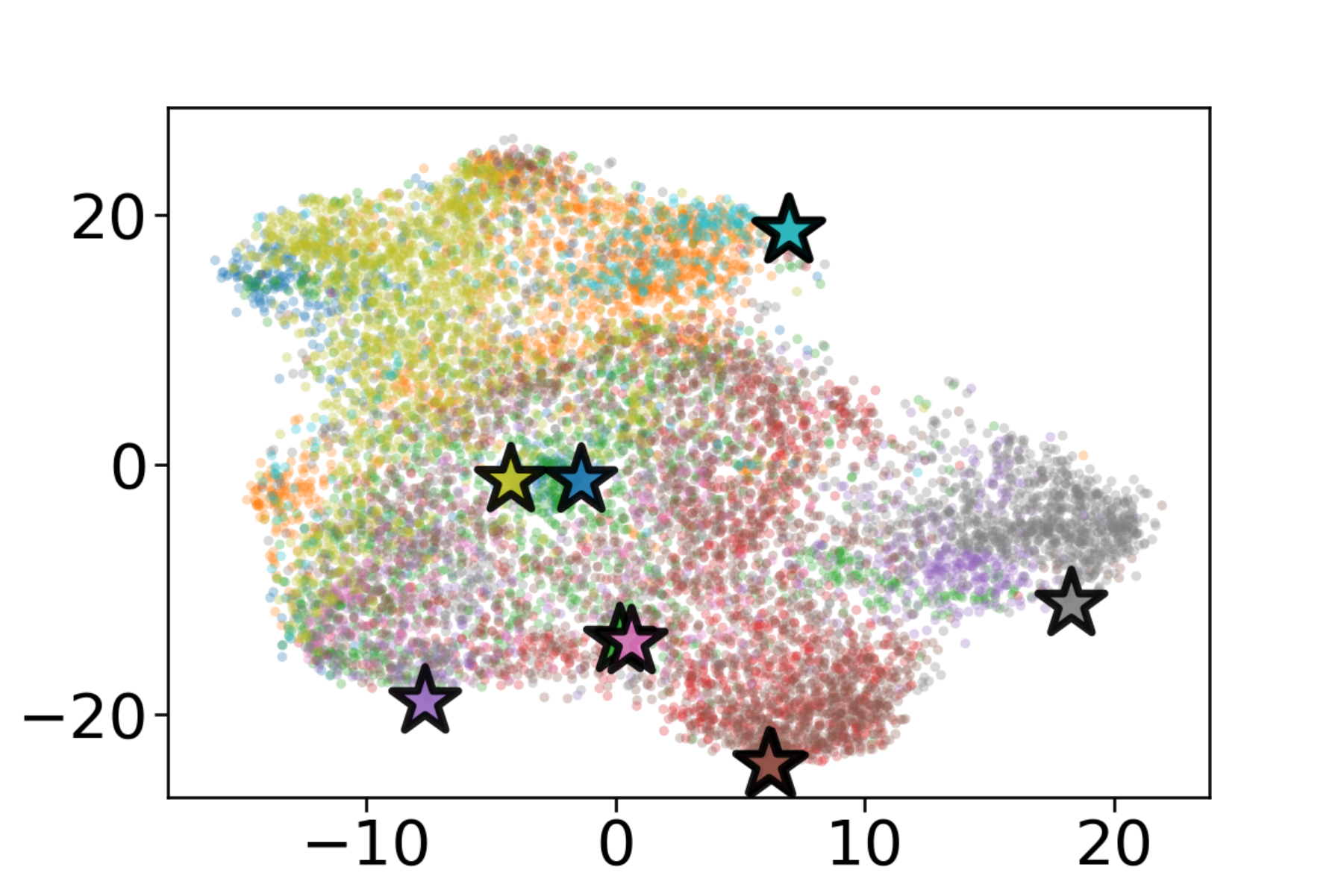}
  & \includegraphics[width=0.2\columnwidth]{figures/source_vitb16_cifar10/t-SNE_pixelate.pdf}
  & \includegraphics[width=0.2\columnwidth]{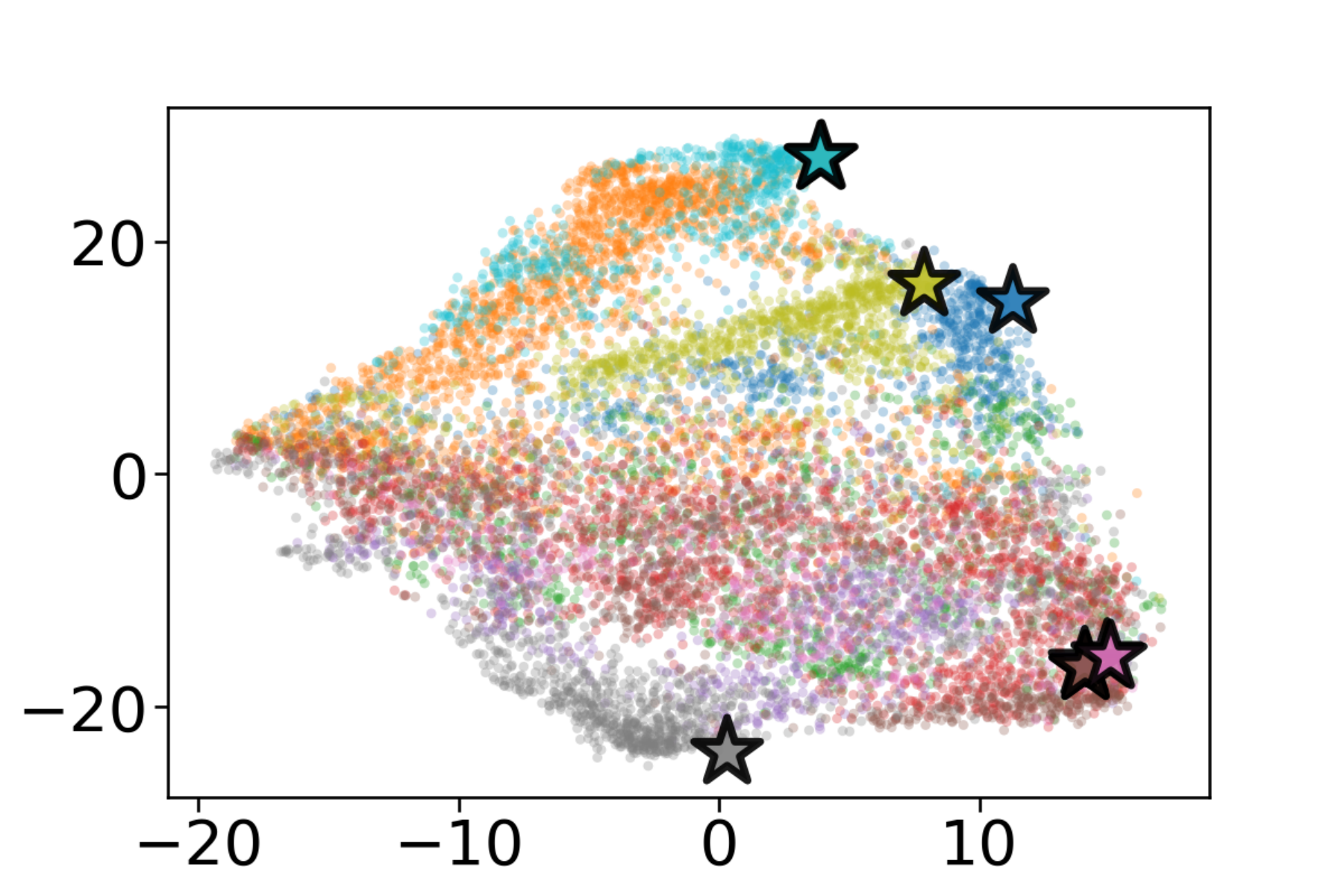}\\
  
\end{tabular}
\caption{Zero-shot ViT-B/16: The t-SNE plots show visual (\(\circ\)) and text ($\bigstar$) features for CIFAR-10C. }
\label{fig: tsne_cifar10_extra_source}
\end{figure}

\begin{figure}[ht!]
\centering
\setlength{\tabcolsep}{1pt}
\begin{tabular}{ccccc}
  { \large \textit{Gaussian}} & {\large \textit{Shot}} & {\large \textit{Impulse}} & {\large \textit{Defocus}} & 
  {\large \textit{Glass}}  \\
  
  \includegraphics[width=0.2\columnwidth]{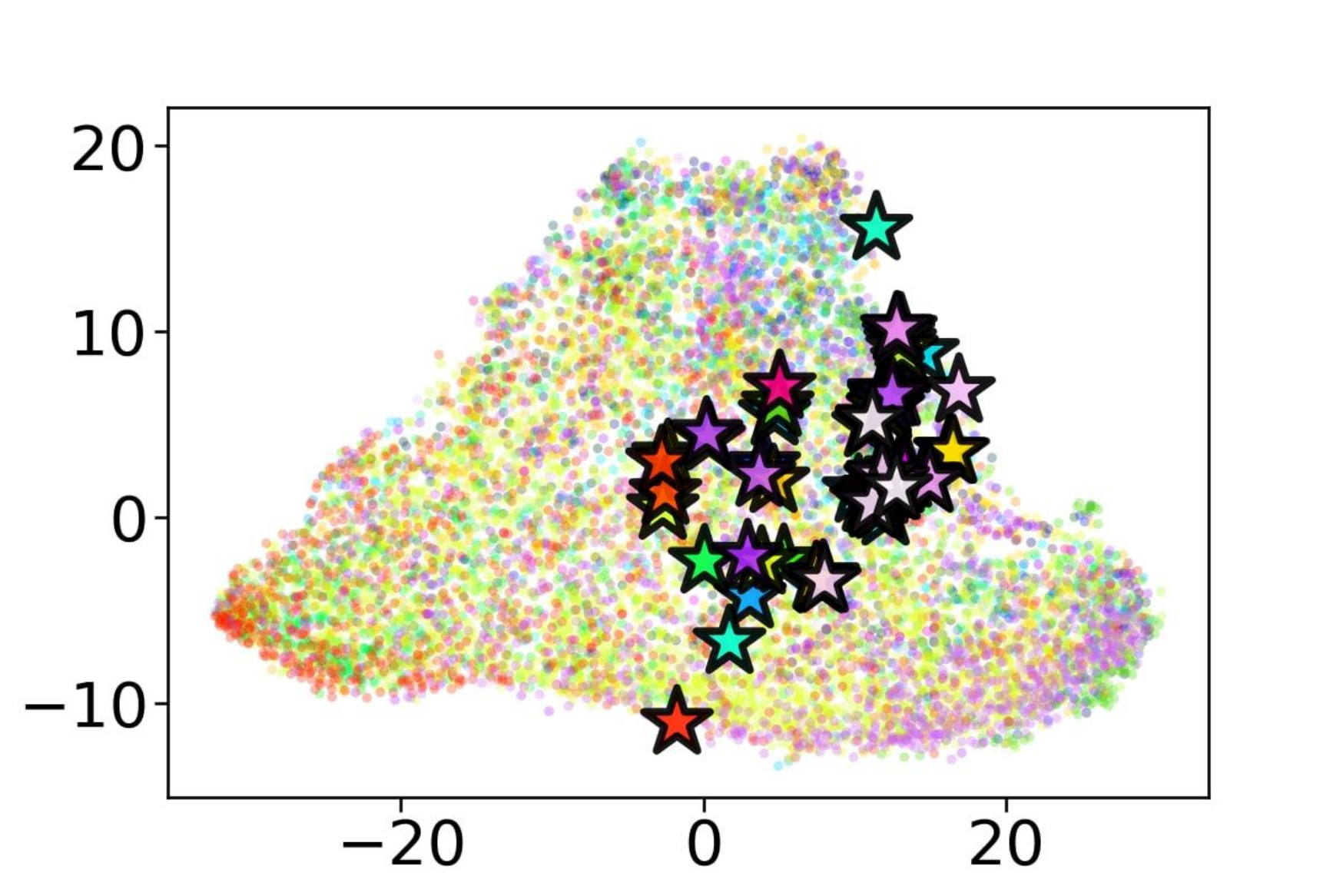}   
  & \includegraphics[width=0.2\columnwidth]{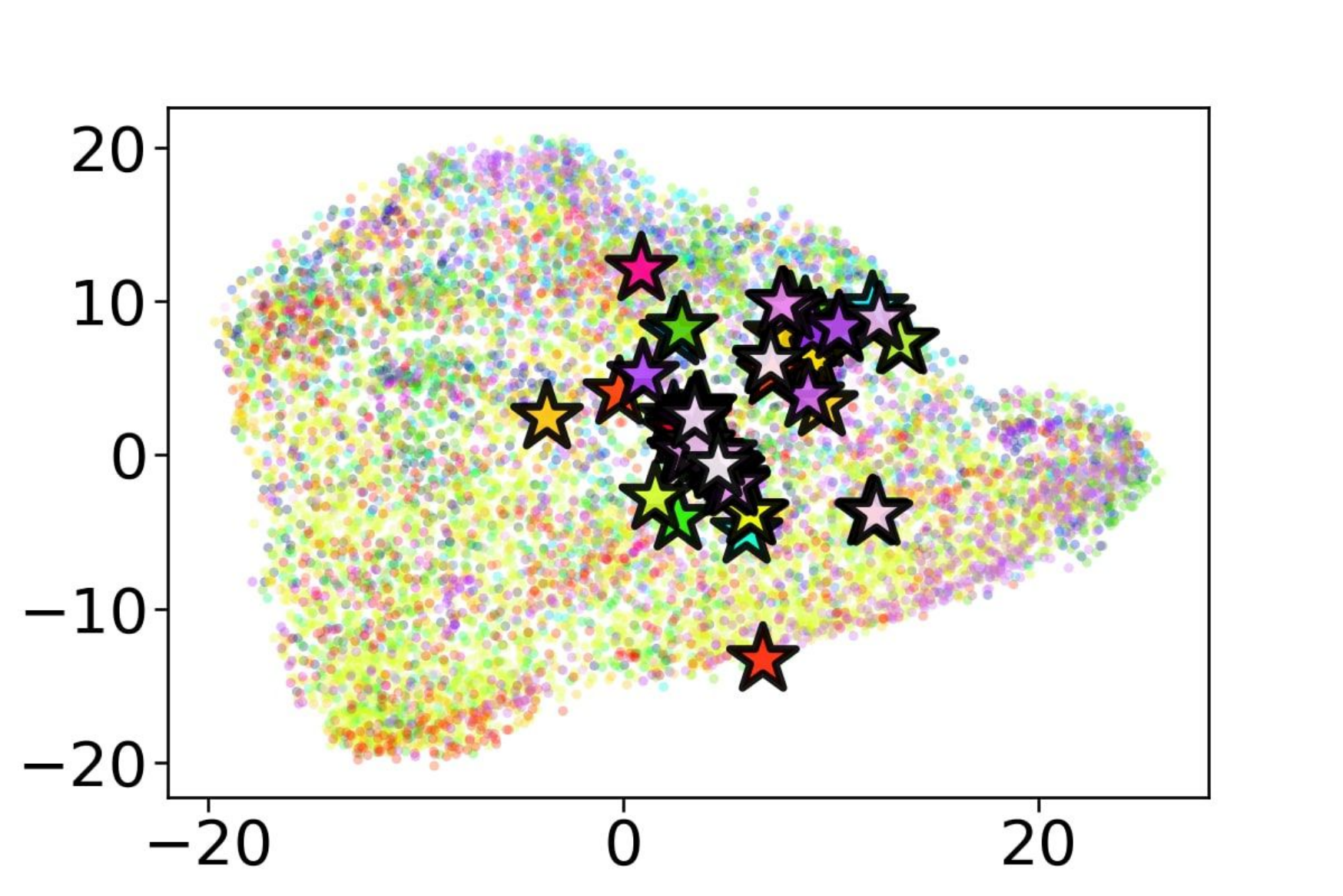}  
  & \includegraphics[width=0.2\columnwidth]{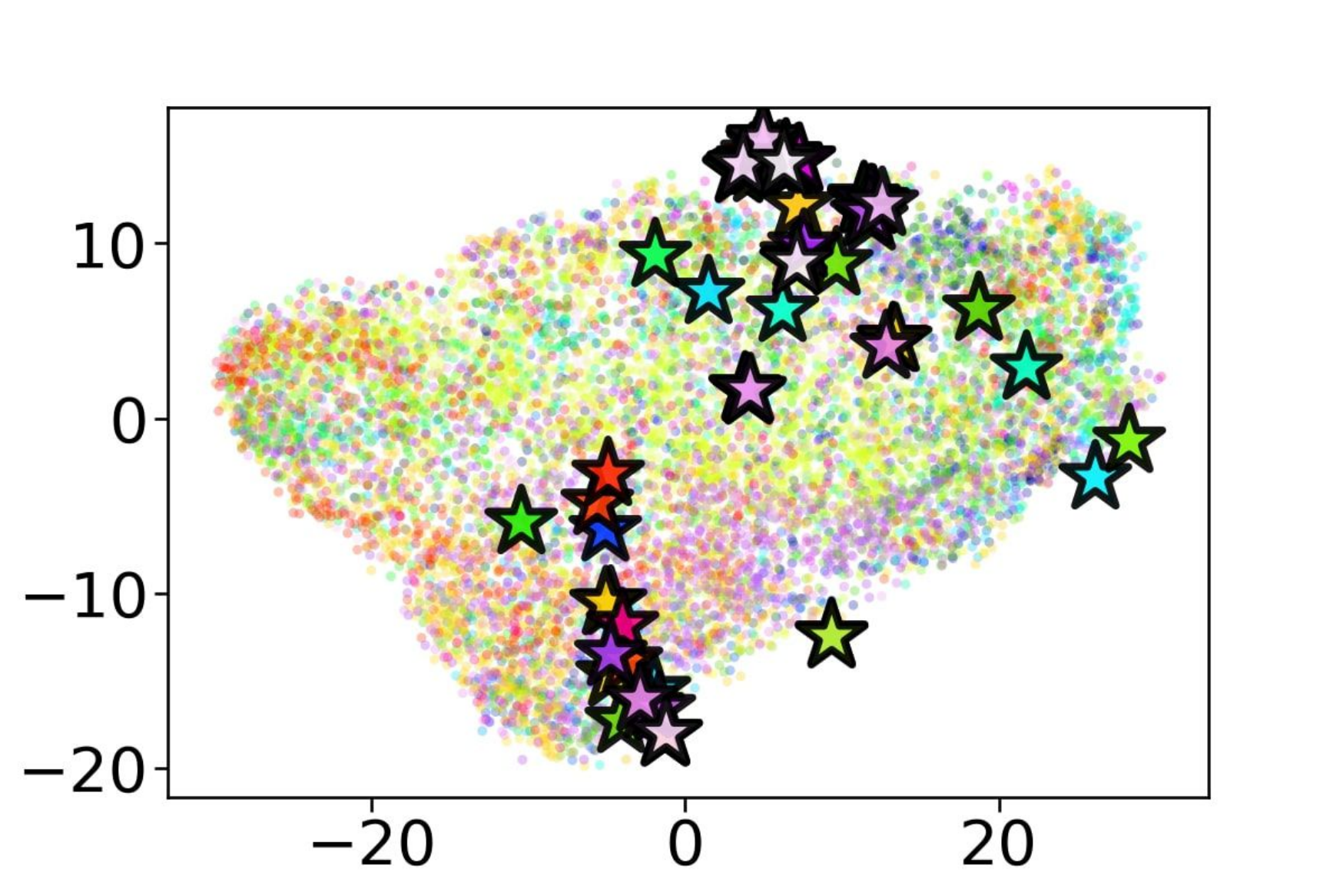}  
  & \includegraphics[width=0.2\columnwidth]{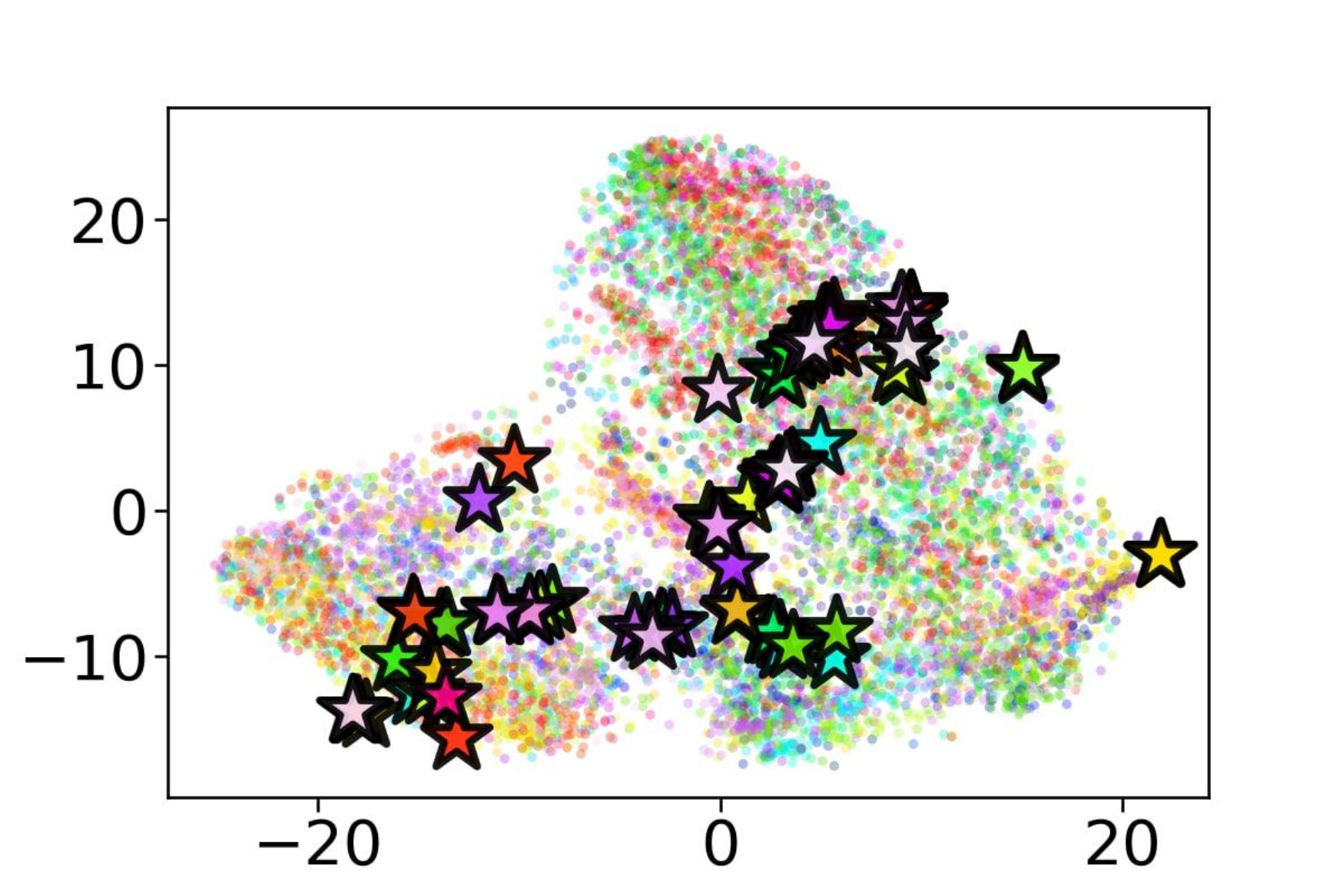}
  & \includegraphics[width=0.2\columnwidth]{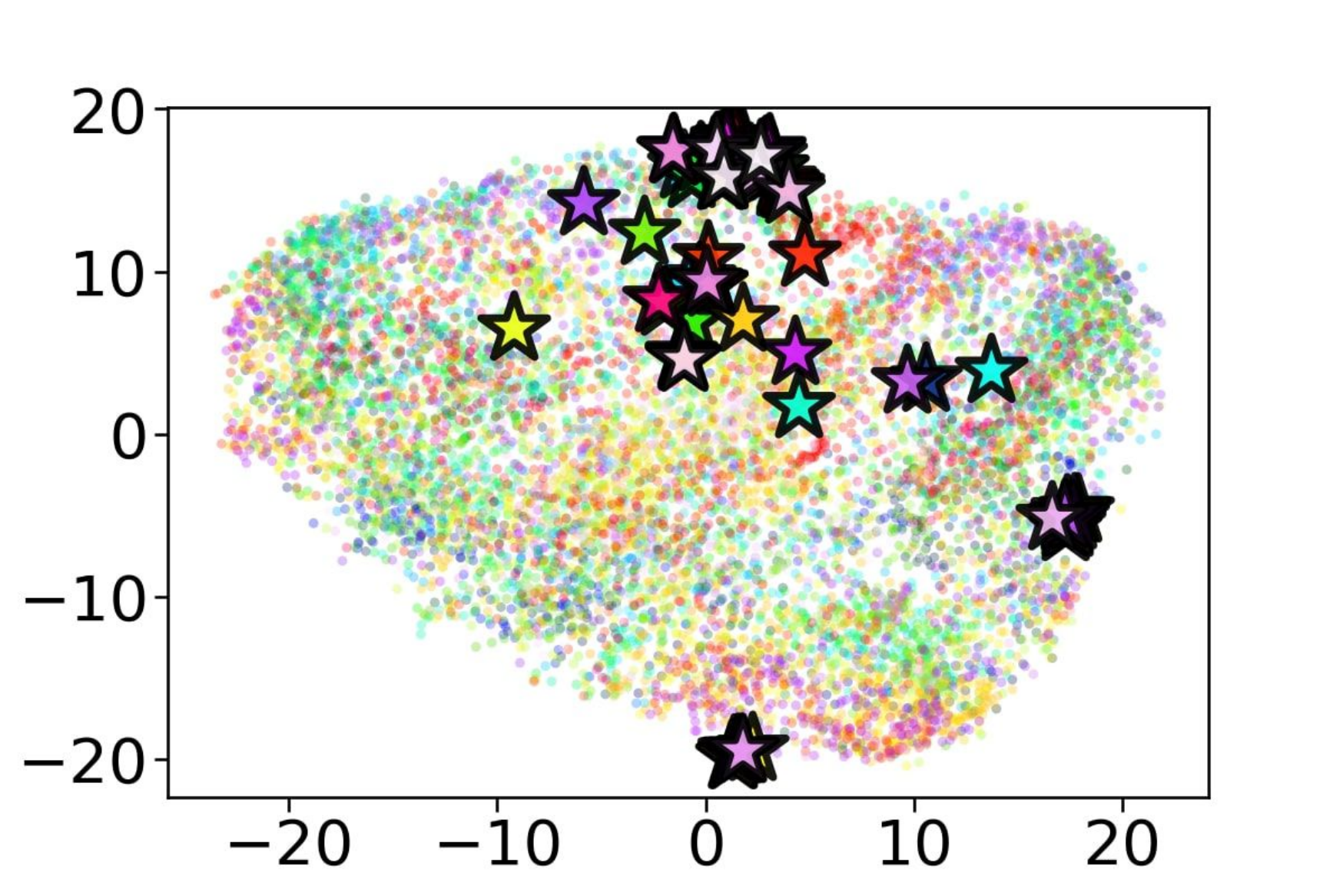}\\
  
  { \large \textit{Motion}} & {\large \textit{Zoom}} & {\large \textit{Snow}} & {\large \textit{Frost}} & 
  {\large \textit{Fog}} \\
  \includegraphics[width=0.2\columnwidth]{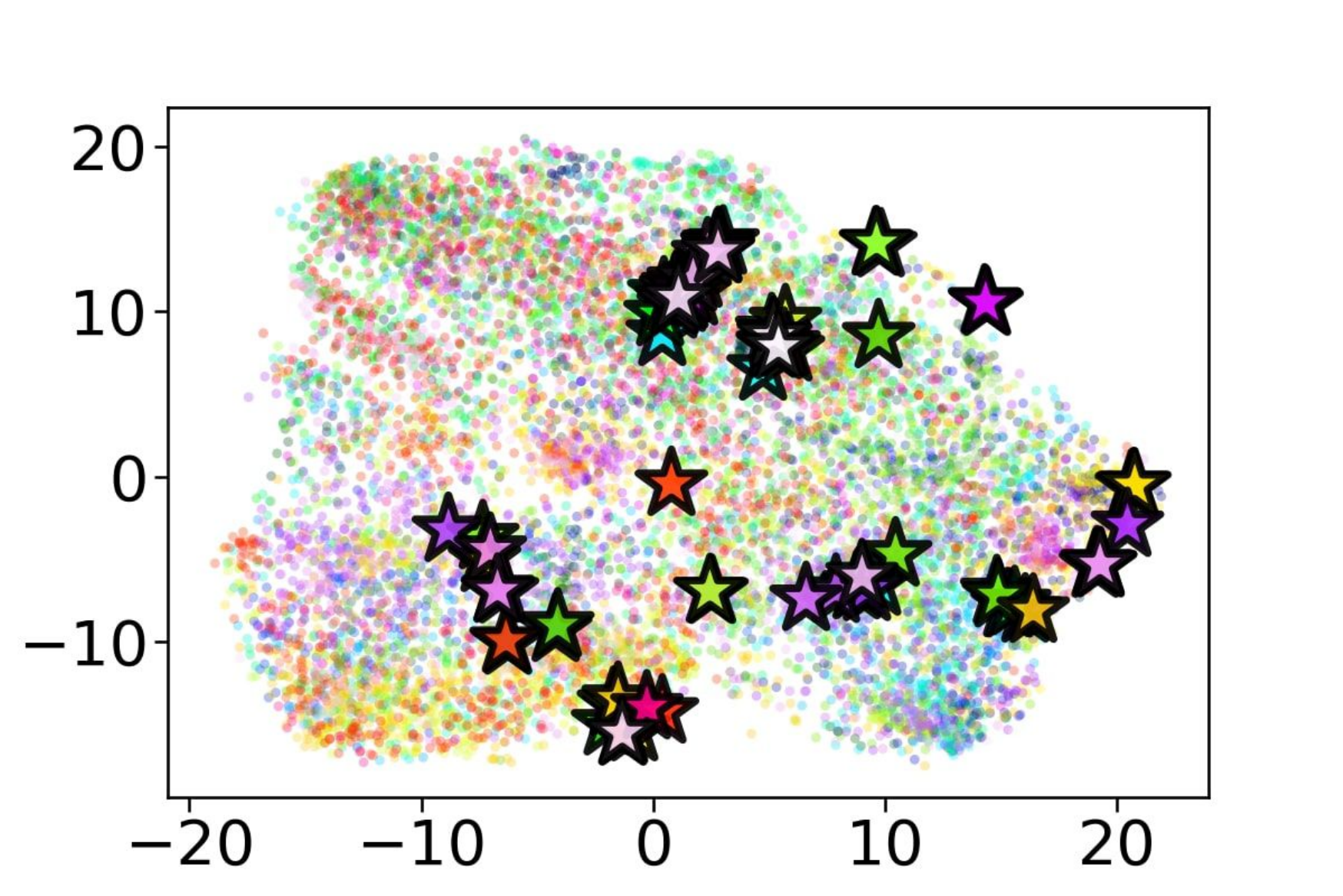} 
  & \includegraphics[width=0.2\columnwidth]{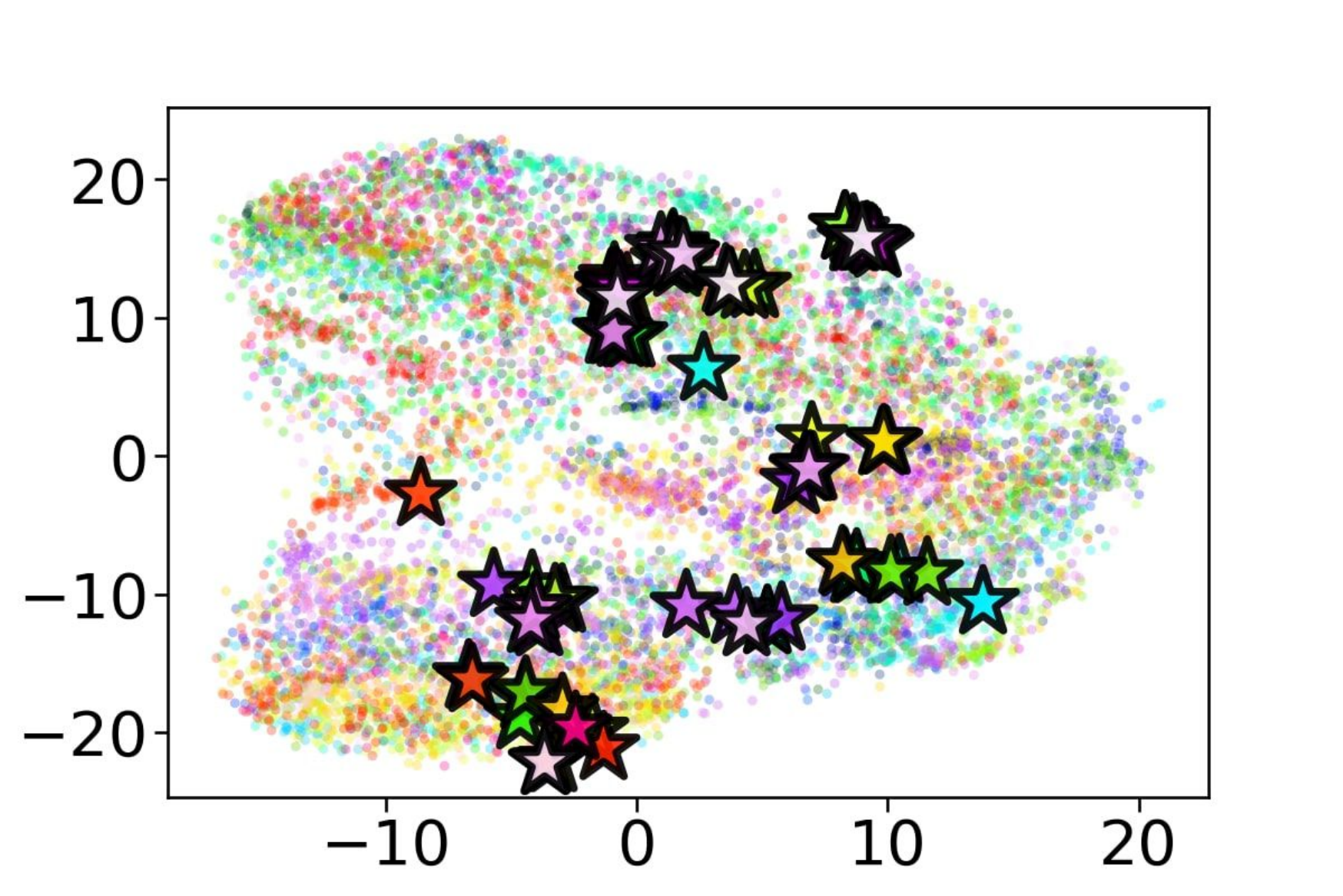}
  & \includegraphics[width=0.2\columnwidth]{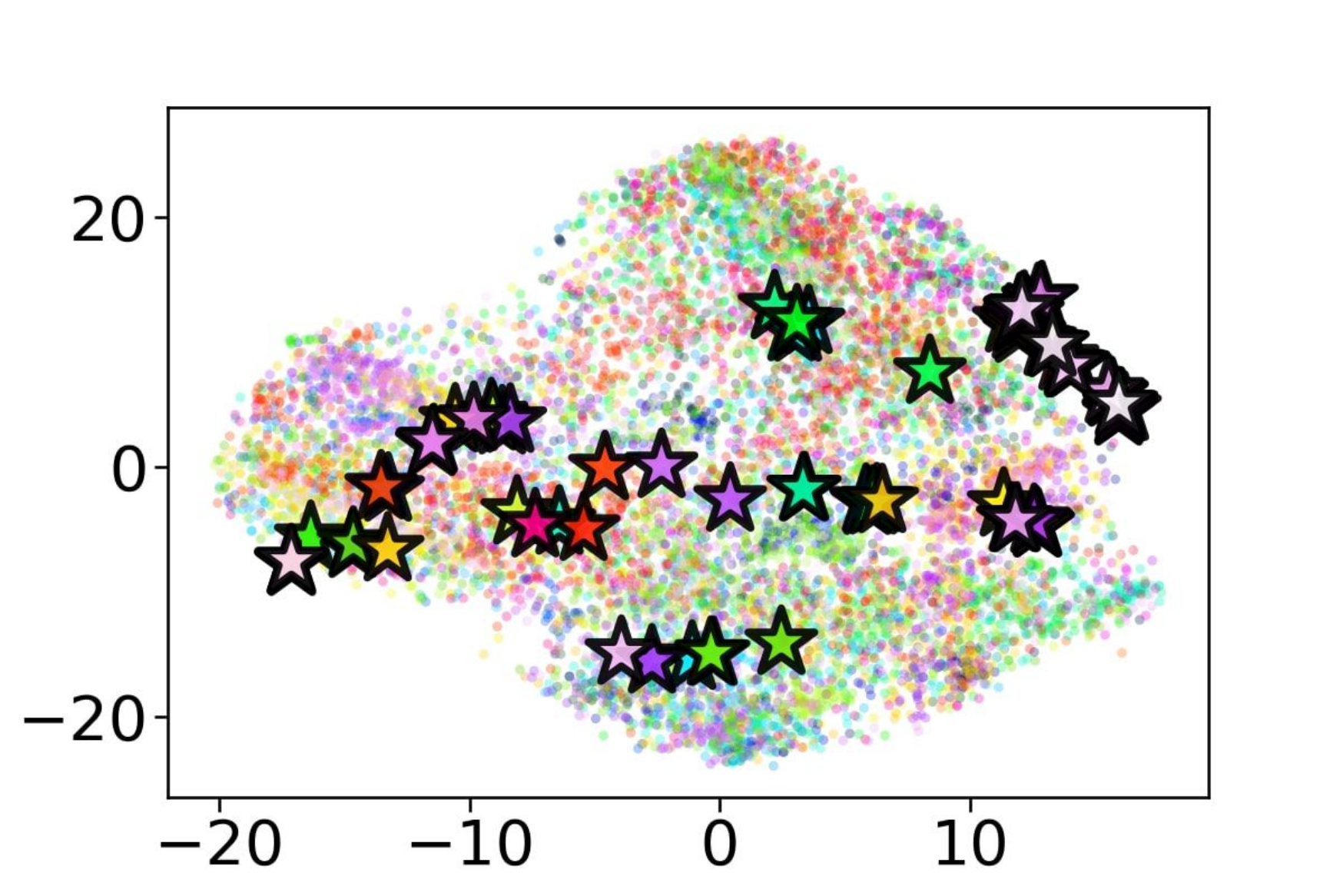}
  & \includegraphics[width=0.2\columnwidth]{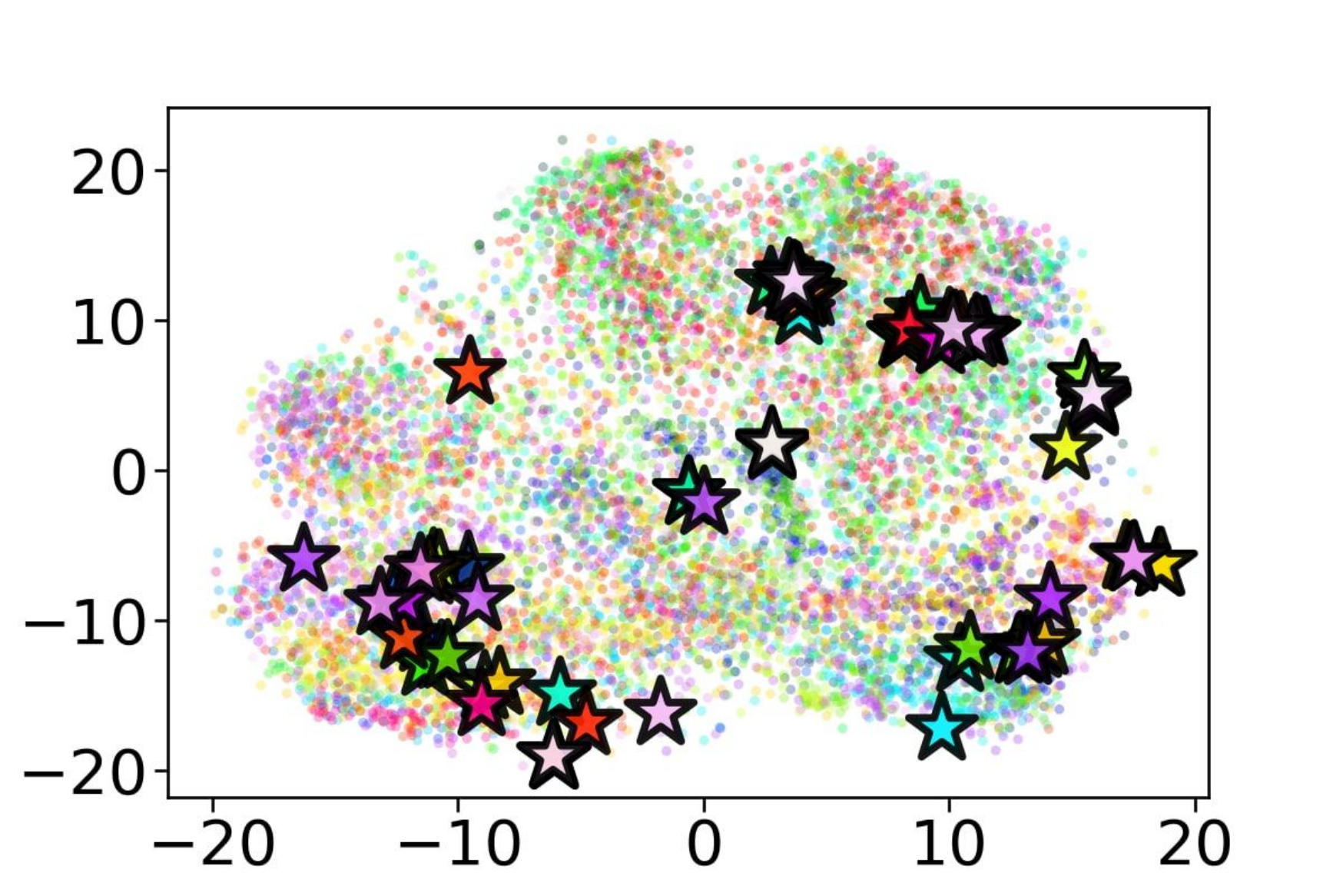}
  & \includegraphics[width=0.2\columnwidth]{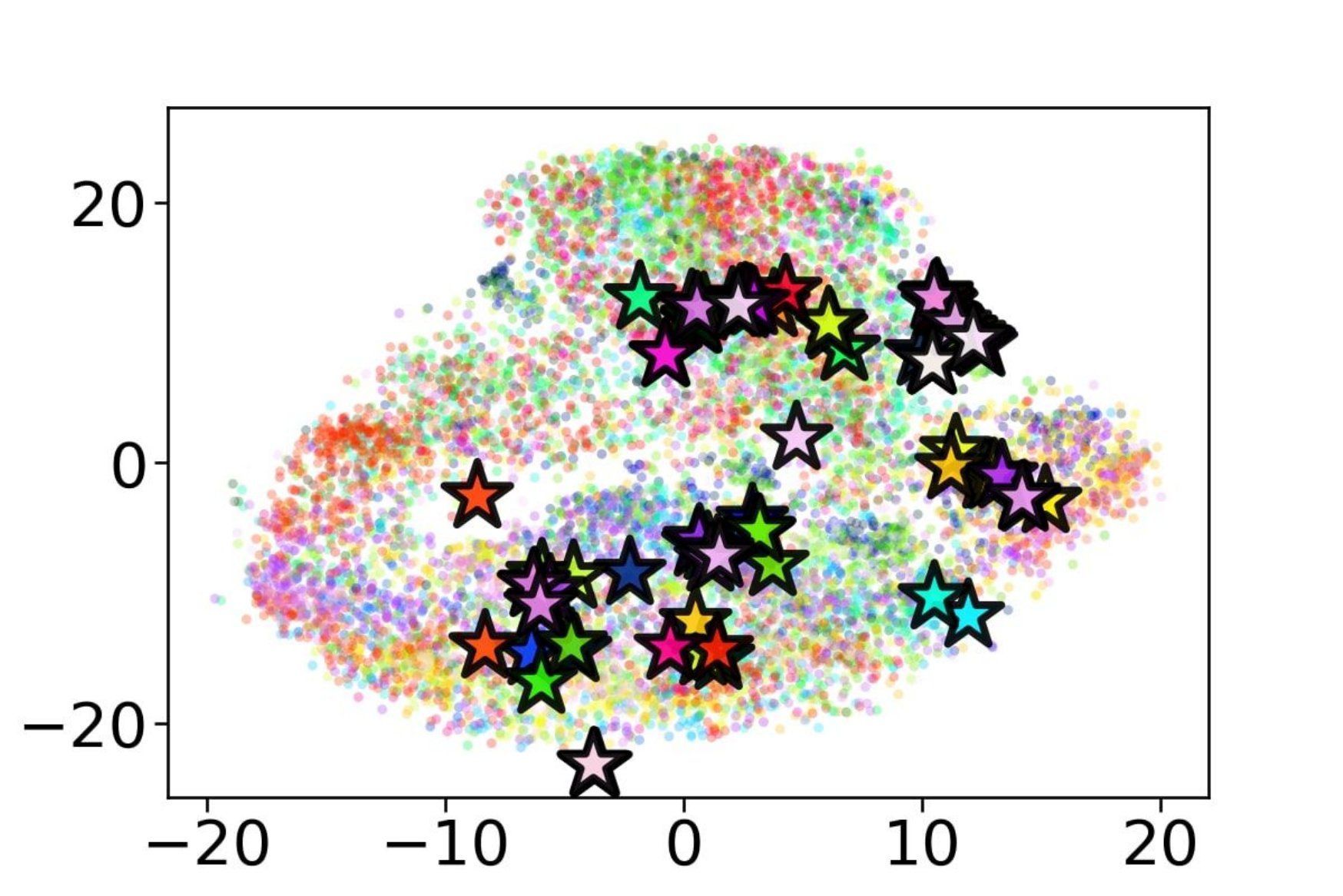}\\

  { \large \textit{Brightness}} & {\large \textit{Contrast}} & {\large \textit{Elastic}} & {\large \textit{Pixelate}} & 
  {\large \textit{JPEG}} \\
  \includegraphics[width=0.2\columnwidth]{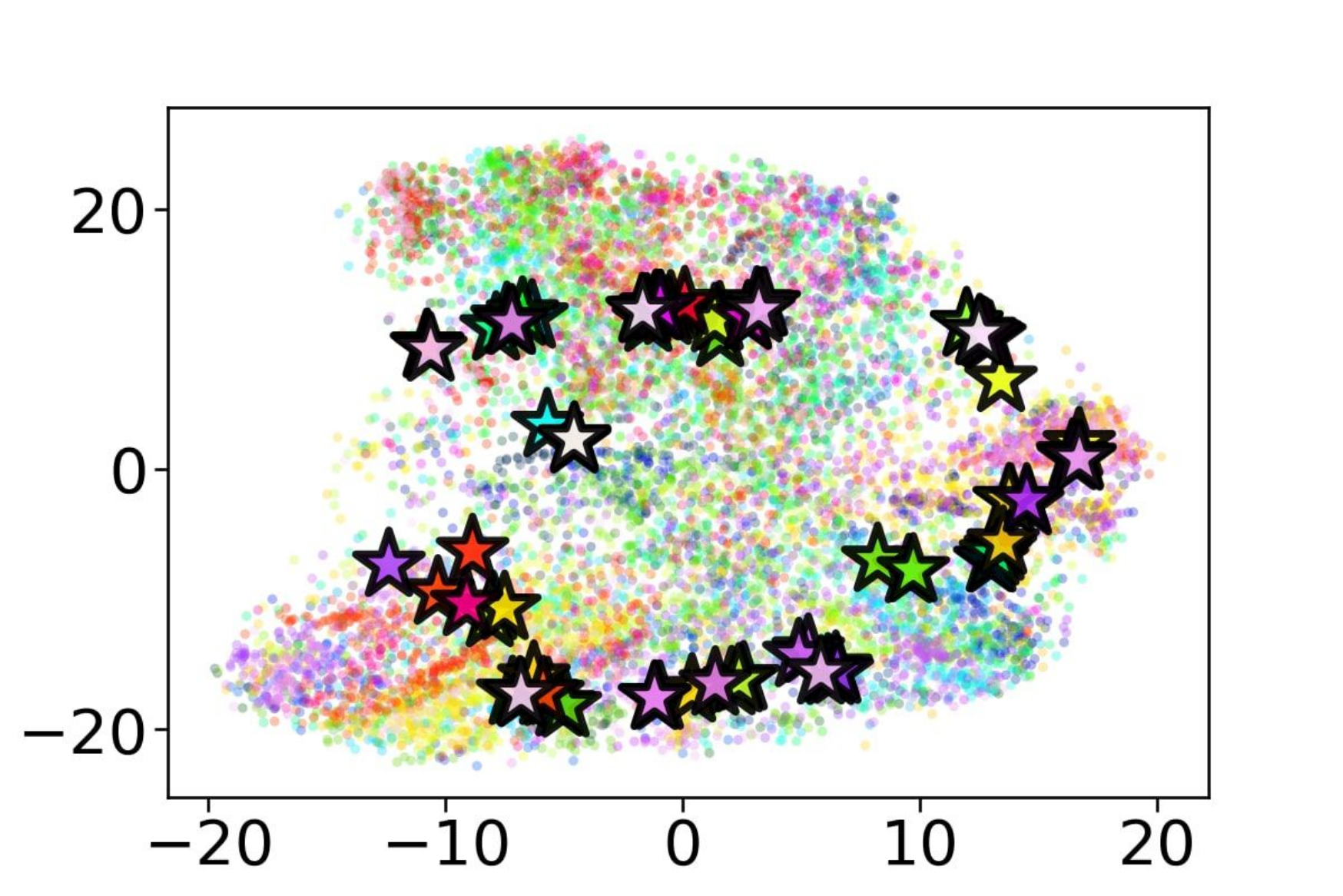} 
  & \includegraphics[width=0.2\columnwidth]{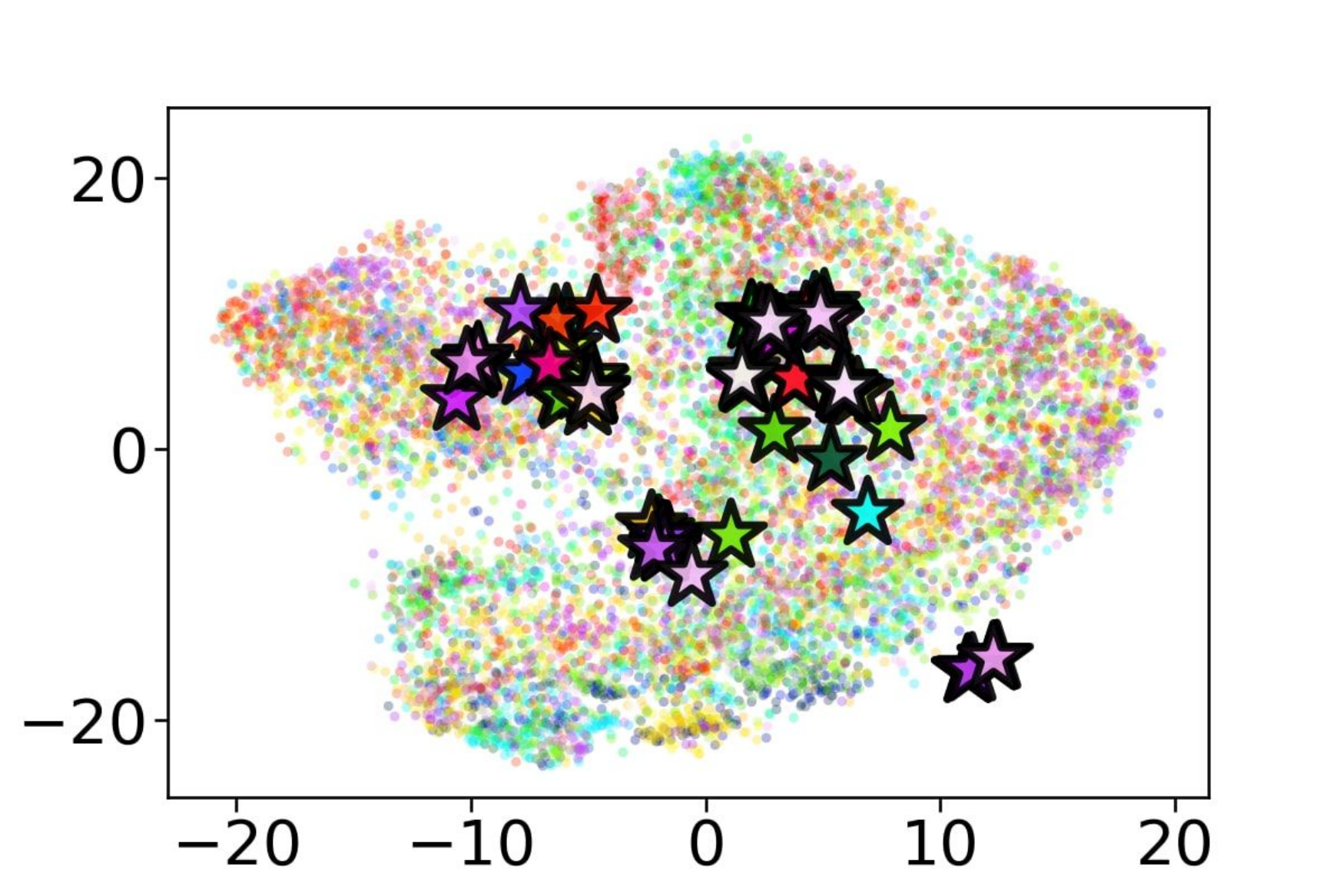}
  & \includegraphics[width=0.2\columnwidth]{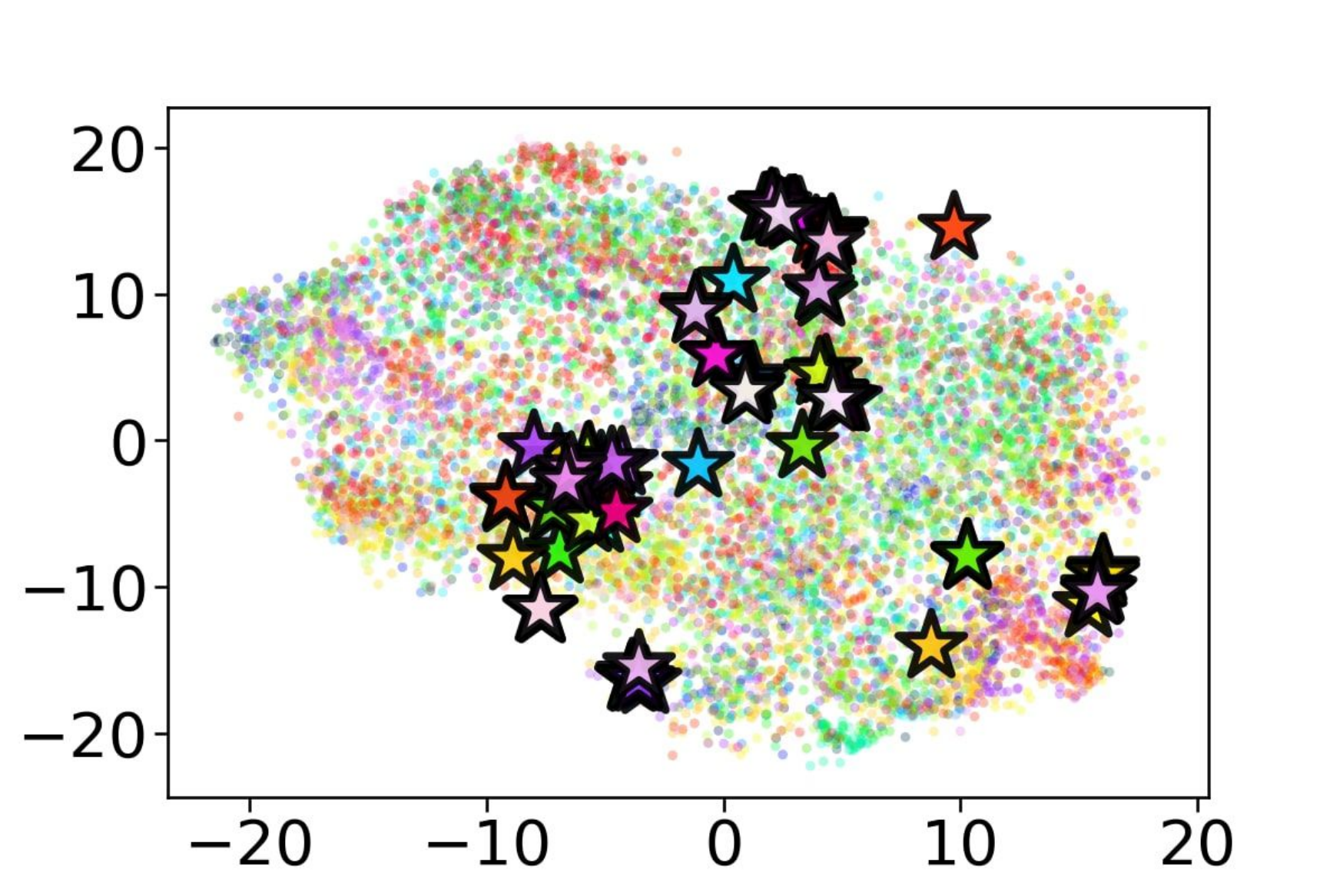}
  & \includegraphics[width=0.2\columnwidth]{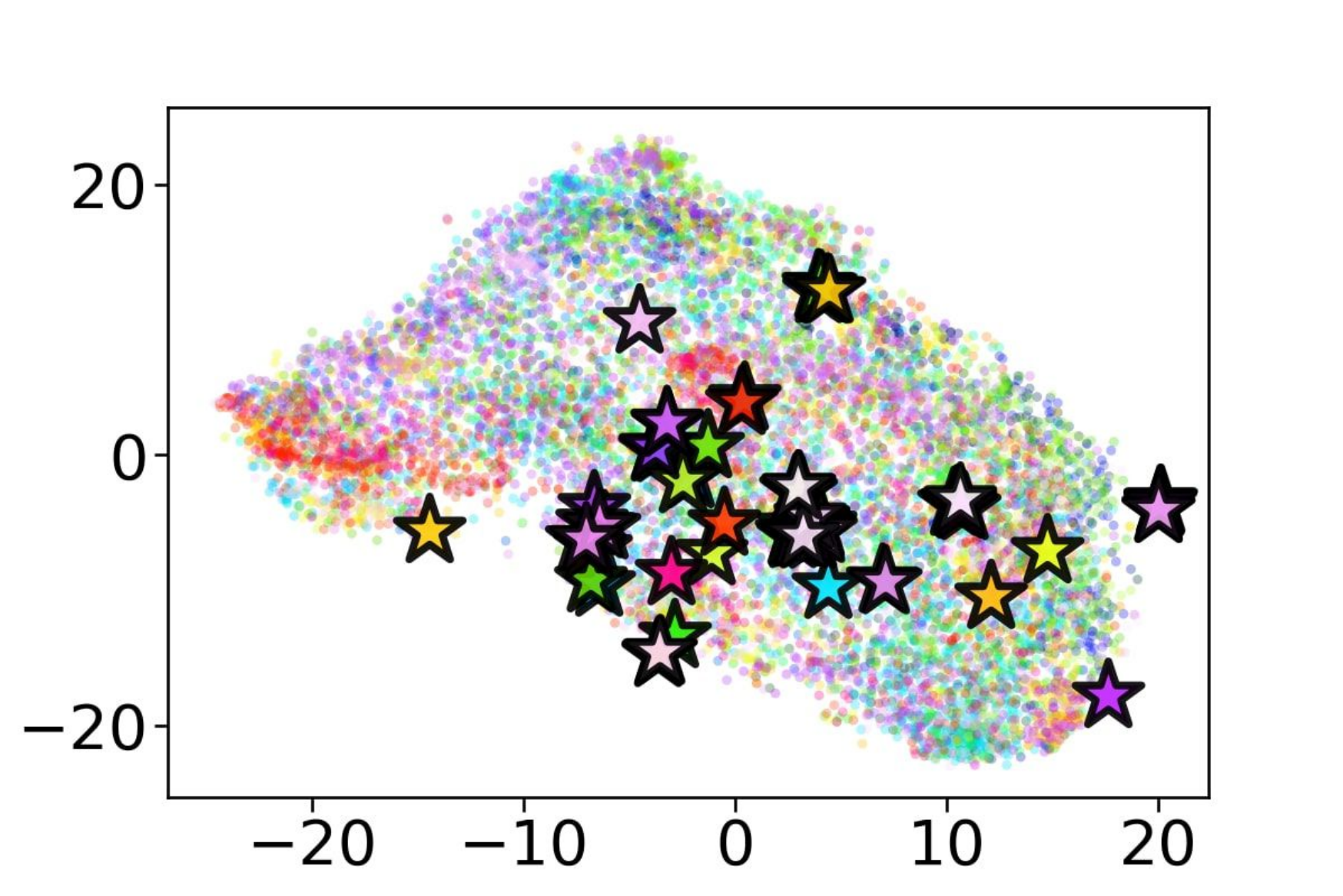}
  & \includegraphics[width=0.2\columnwidth]{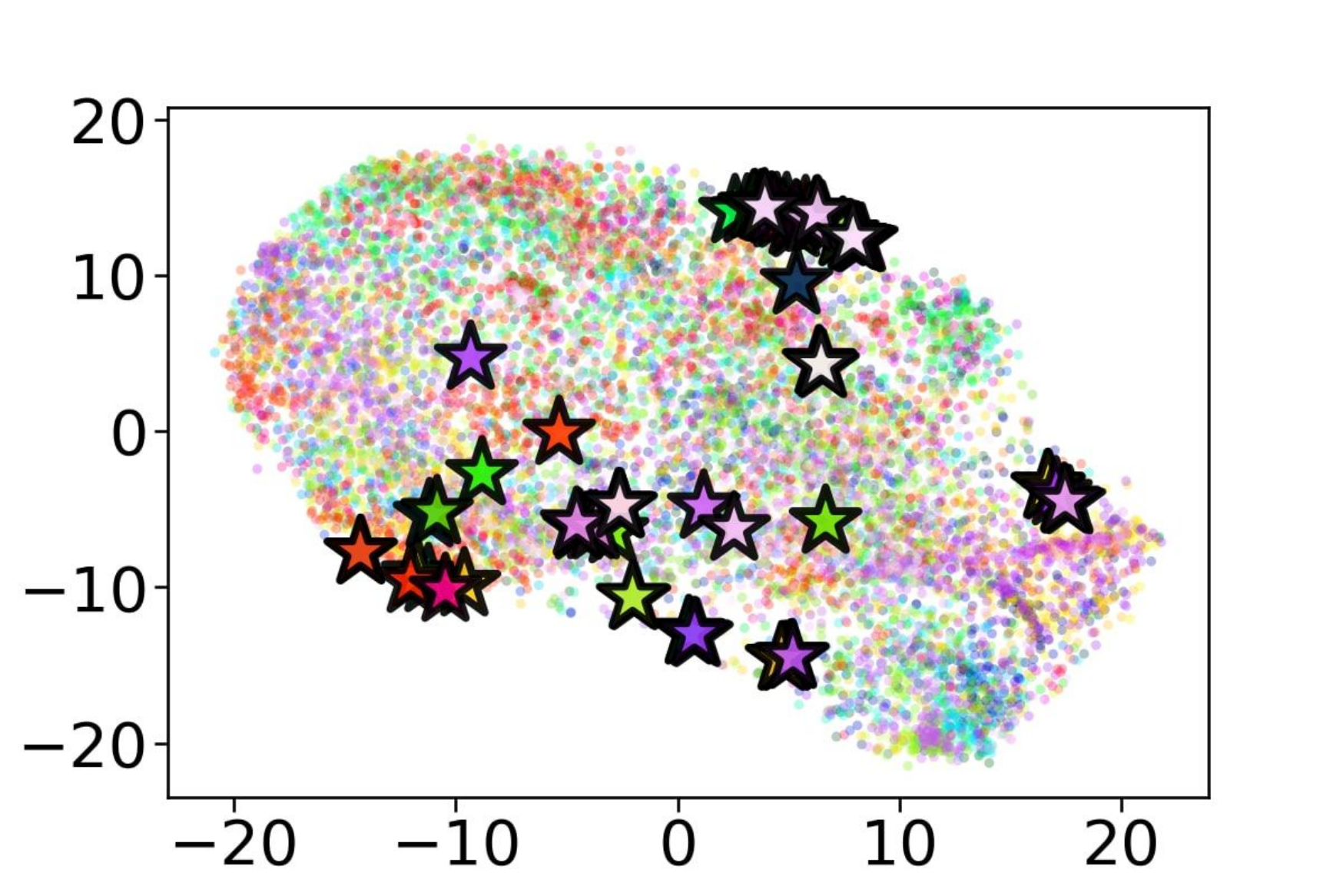}\\
  
\end{tabular}
\caption{\framework\ (w/ ViT-B/16): The t-SNE plots show visual (\(\circ\)) and text ($\bigstar$) features for CIFAR-100C. }
\label{fig: tsne_cifar100_extra}
\end{figure}

\begin{figure}[ht!]
\centering
\setlength{\tabcolsep}{1pt}
\begin{tabular}{ccccc}
  { \large \textit{Gaussian}} & {\large \textit{Shot}} & {\large \textit{Impulse}} & {\large \textit{Defocus}} & 
  {\large \textit{Glass}}  \\
  
  \includegraphics[width=0.2\columnwidth]{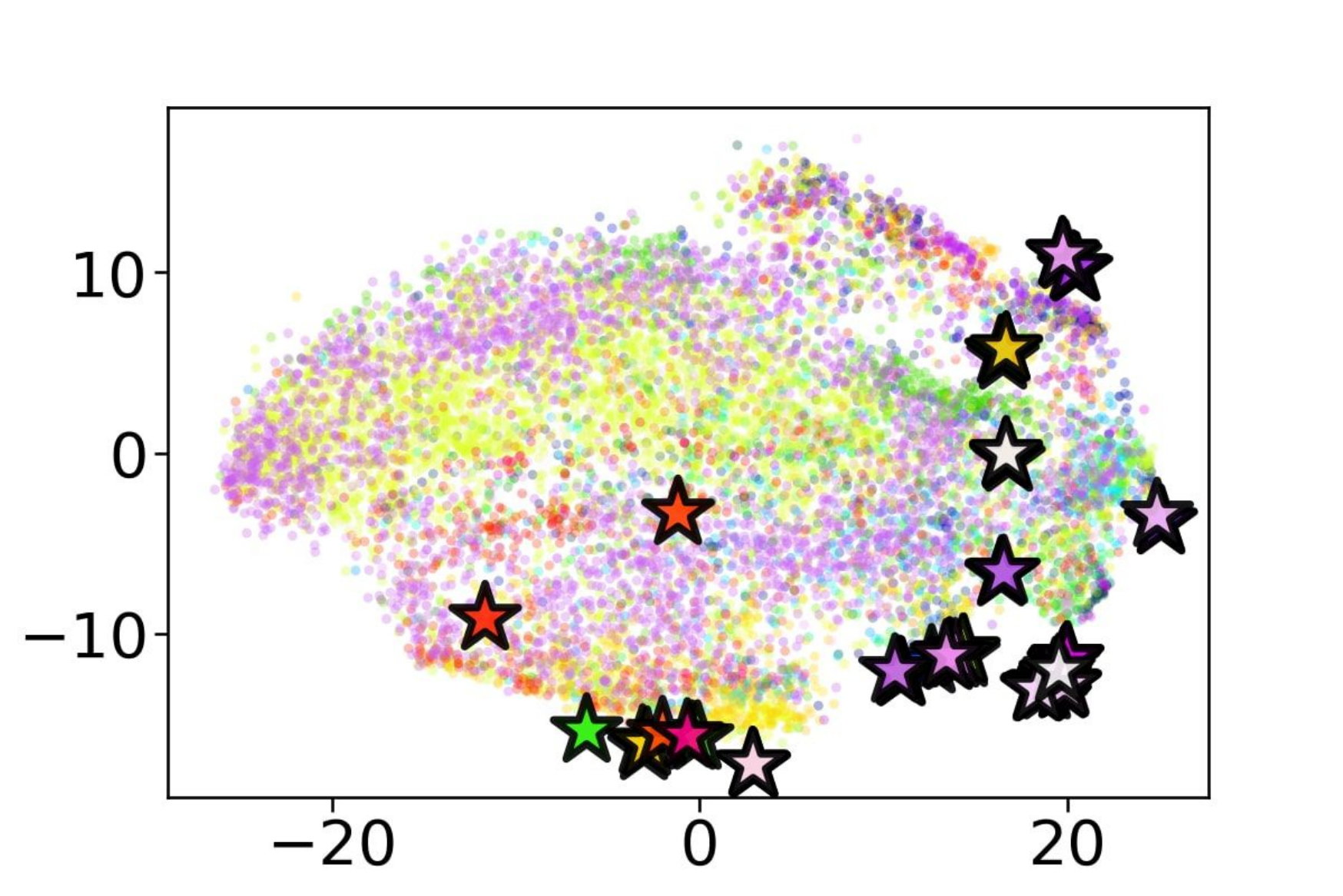}   
  & \includegraphics[width=0.2\columnwidth]{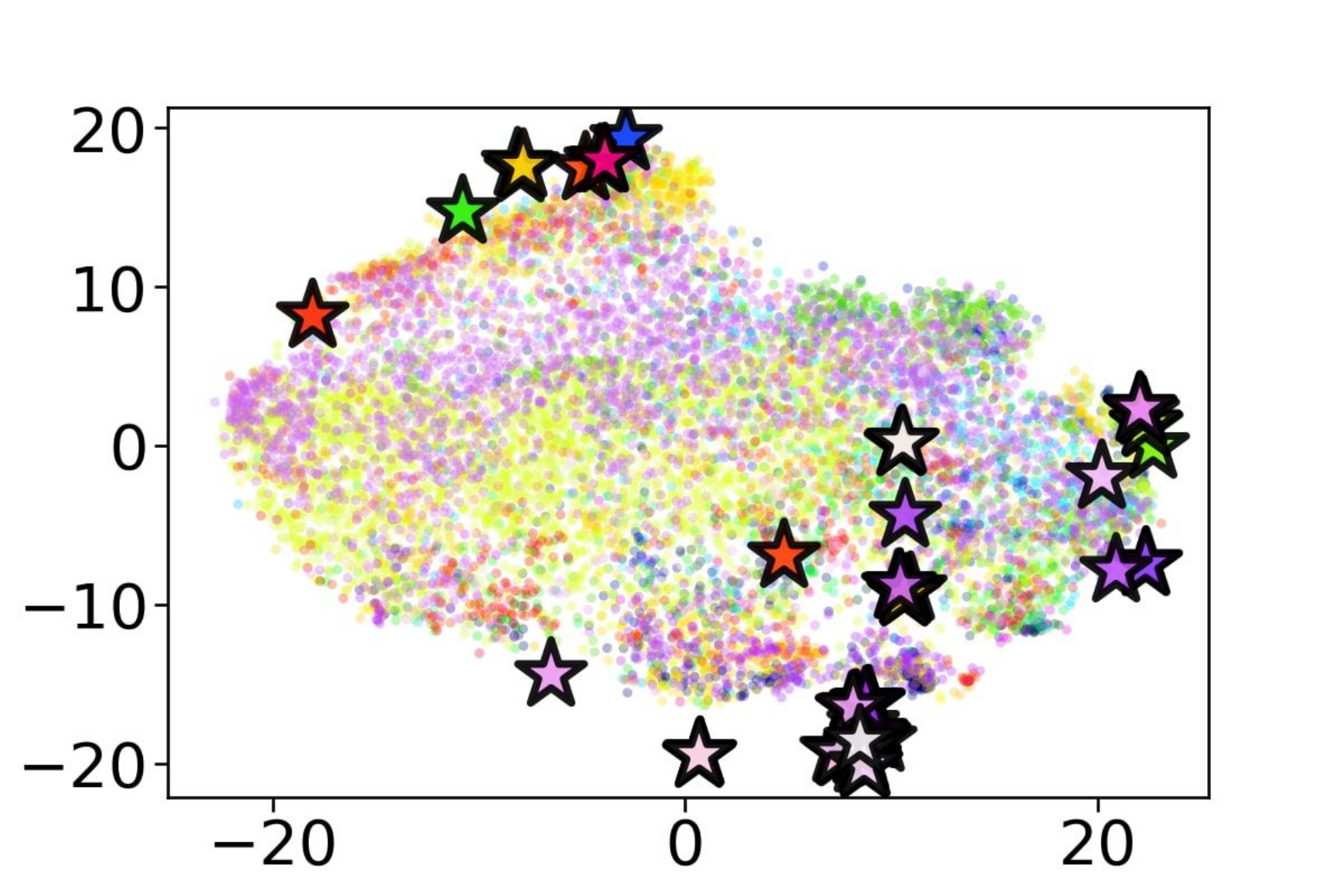}  
  & \includegraphics[width=0.2\columnwidth]{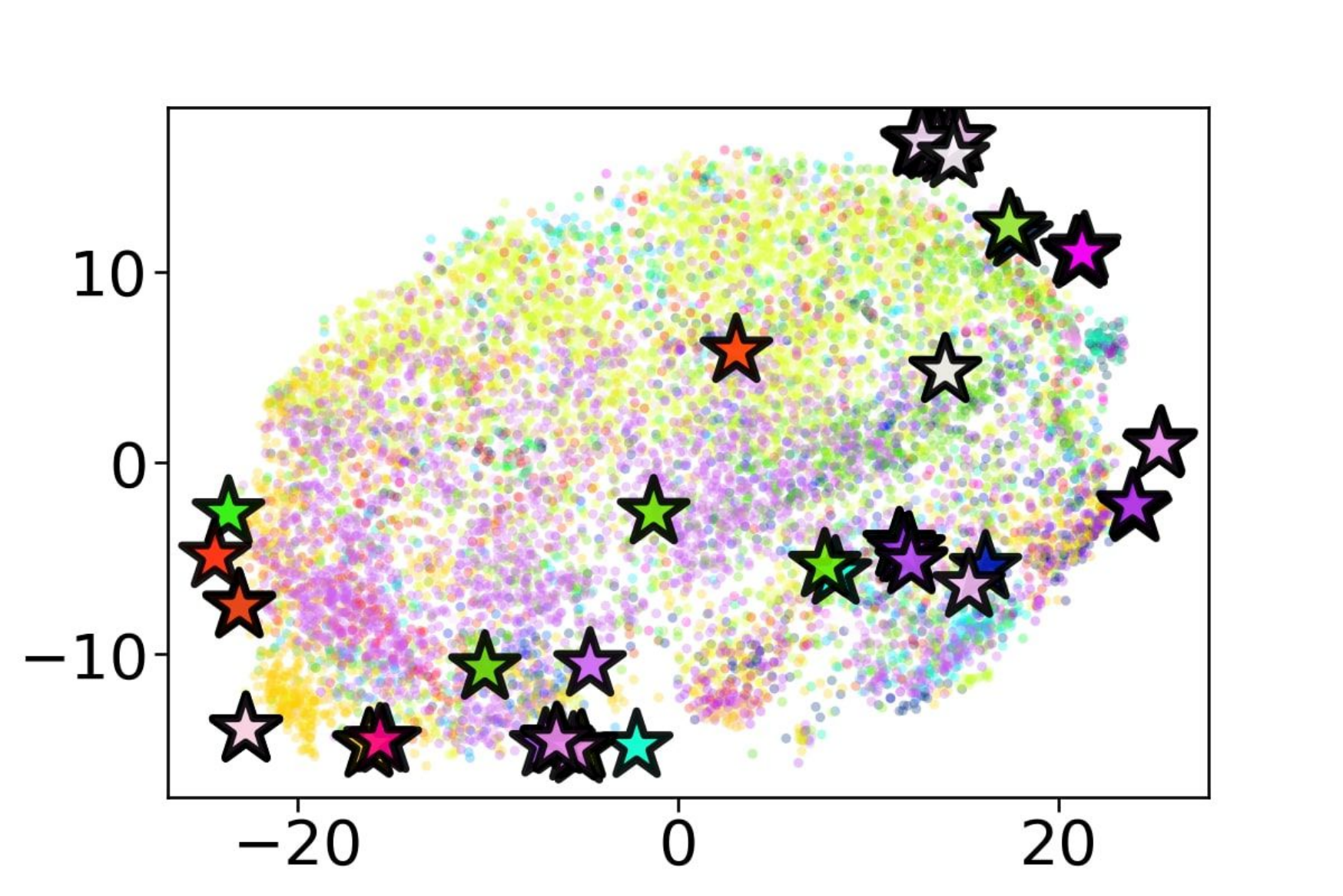}  
  & \includegraphics[width=0.2\columnwidth]{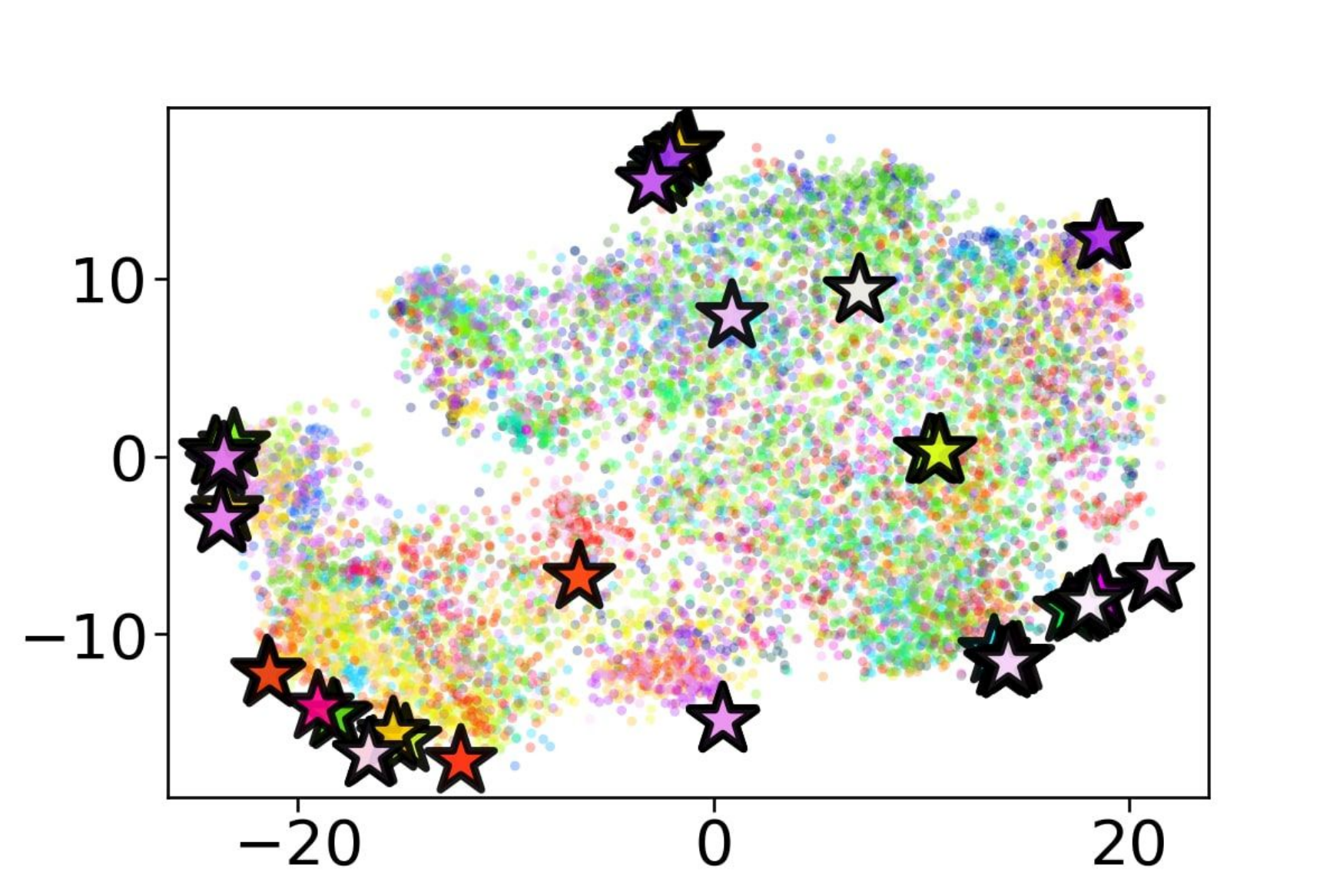}
  & \includegraphics[width=0.2\columnwidth]{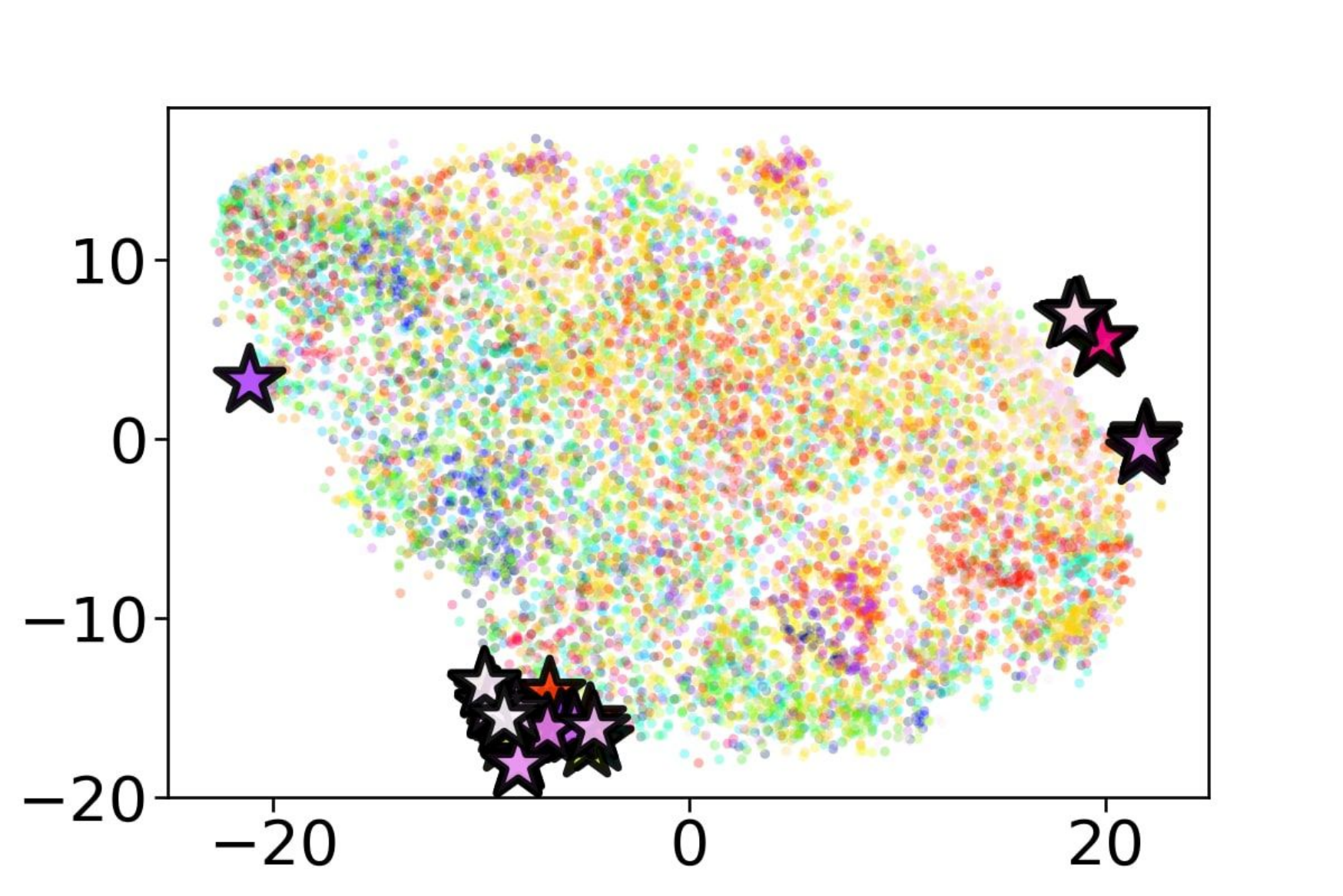}\\
  
  { \large \textit{Motion}} & {\large \textit{Zoom}} & {\large \textit{Snow}} & {\large \textit{Frost}} & 
  {\large \textit{Fog}} \\
  \includegraphics[width=0.2\columnwidth]{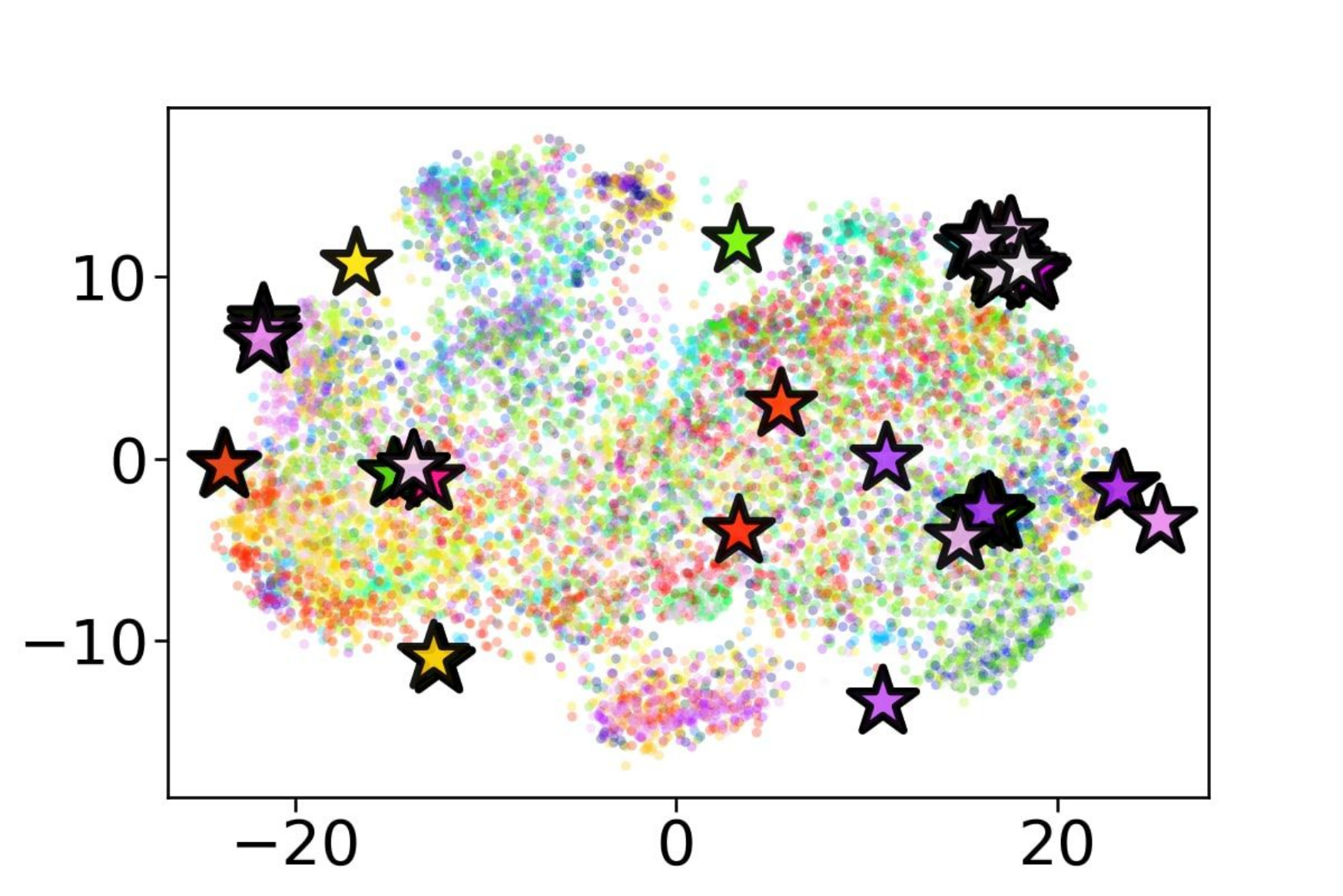} 
  & \includegraphics[width=0.2\columnwidth]{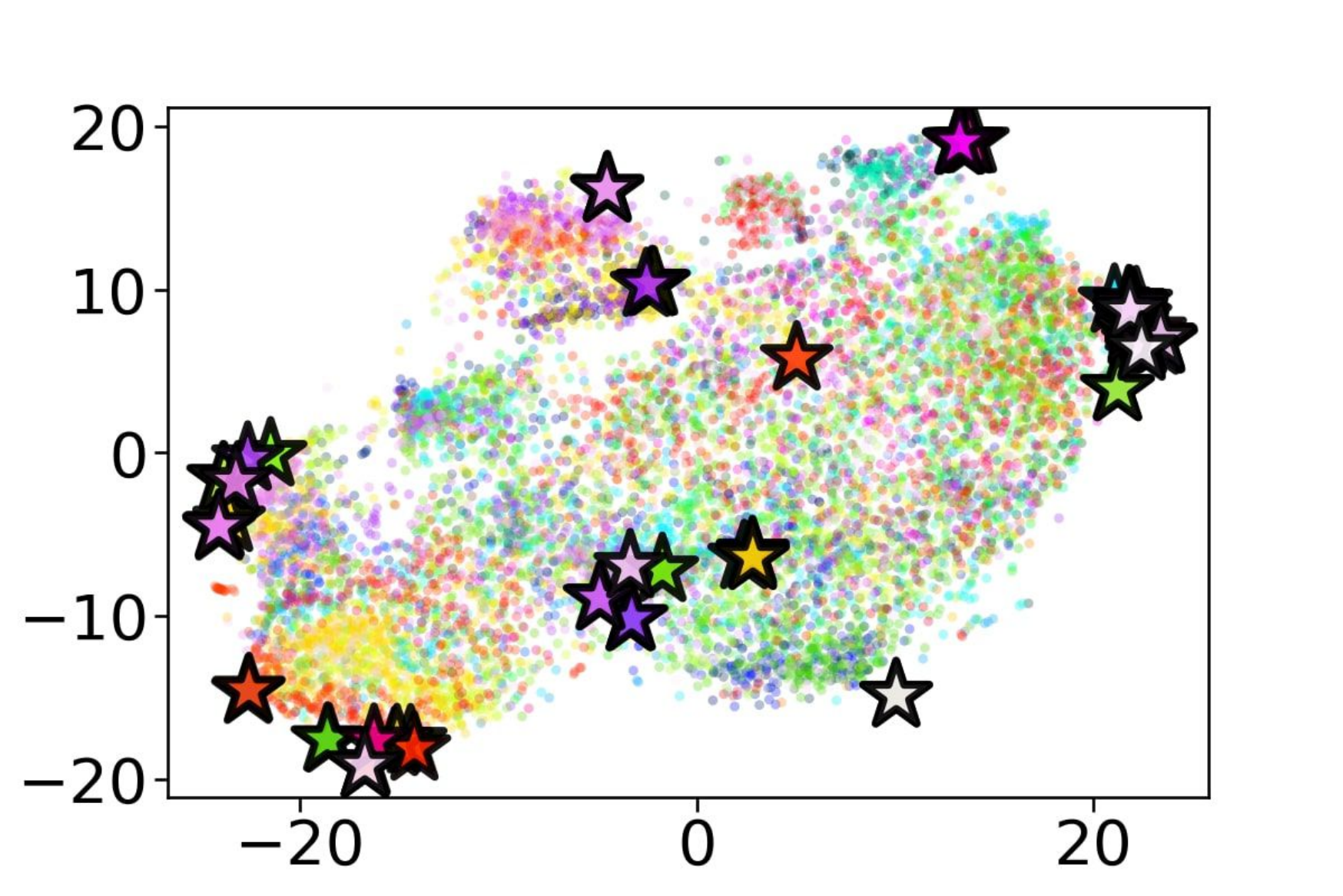}
  & \includegraphics[width=0.2\columnwidth]{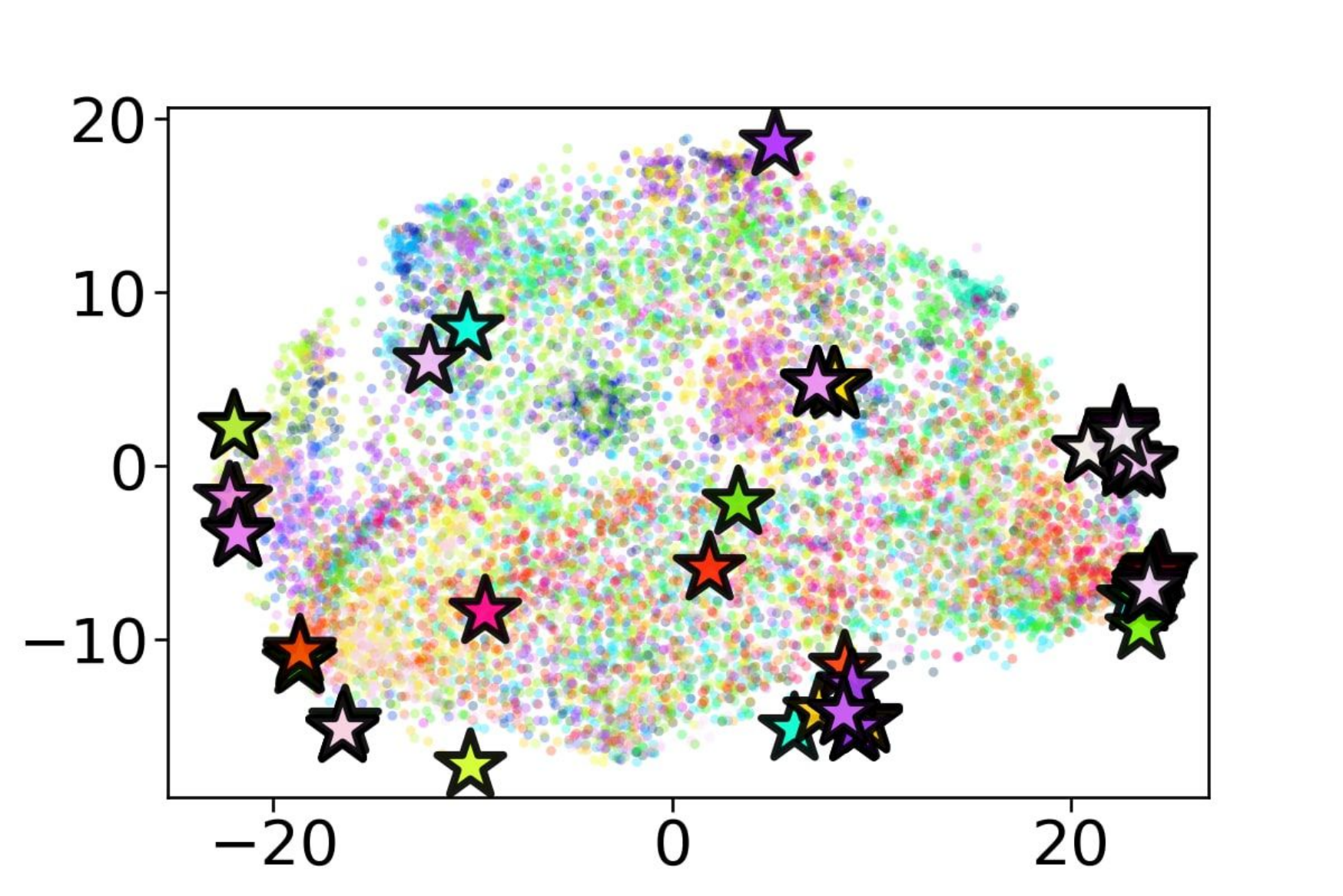}
  & \includegraphics[width=0.2\columnwidth]{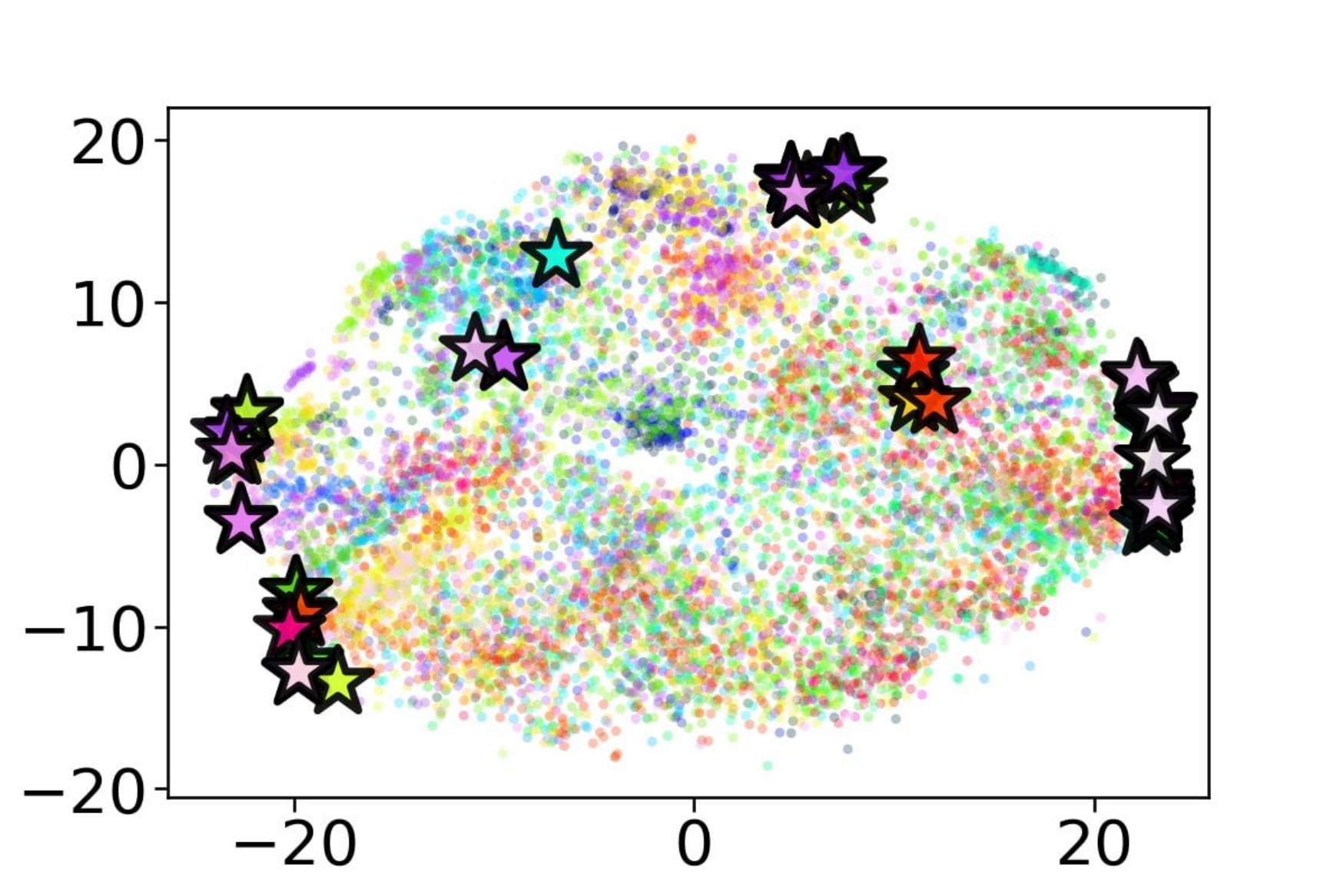}
  & \includegraphics[width=0.2\columnwidth]{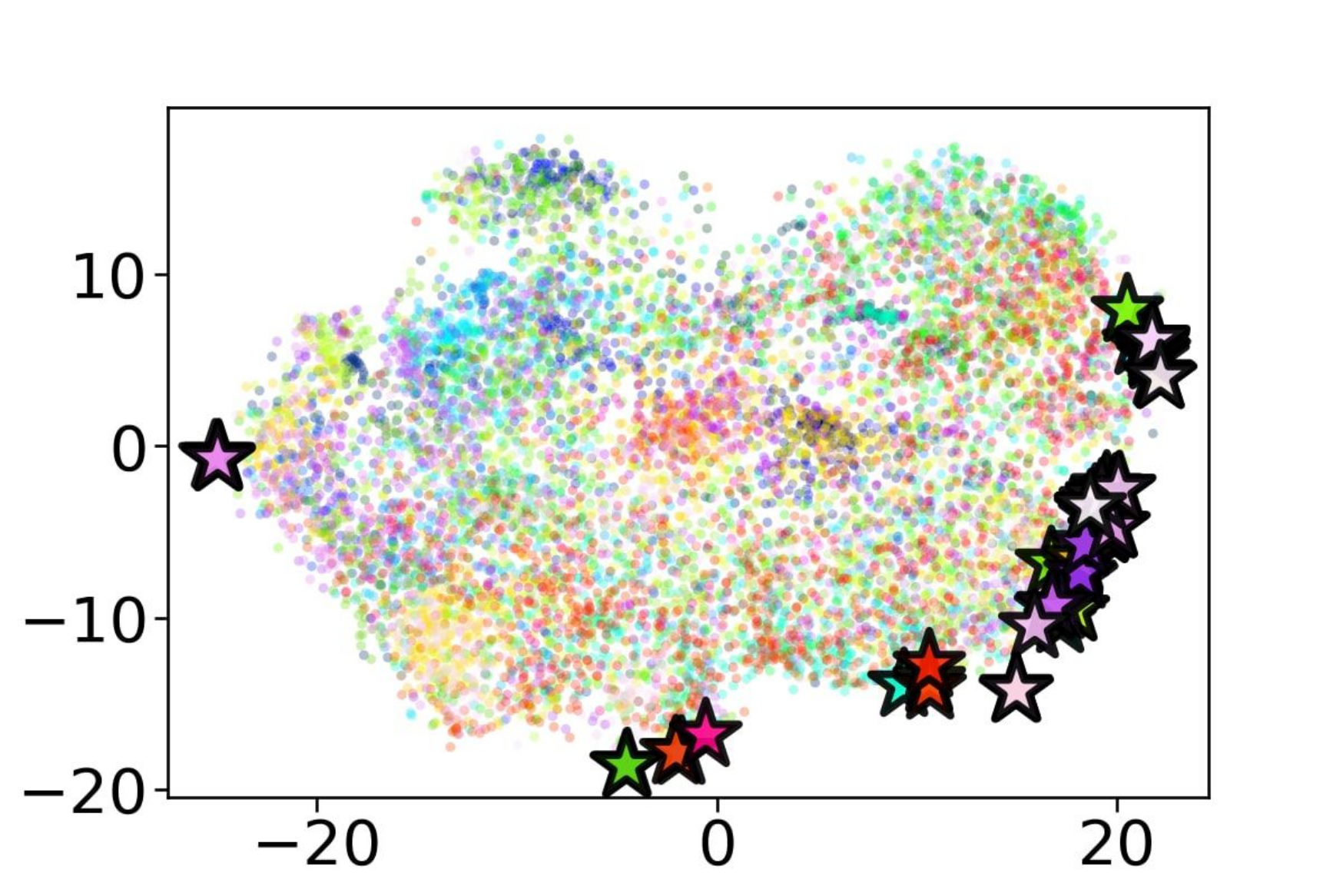}\\

  { \large \textit{Brightness}} & {\large \textit{Contrast}} & {\large \textit{Elastic}} & {\large \textit{Pixelate}} & 
  {\large \textit{JPEG}} \\
  \includegraphics[width=0.2\columnwidth]{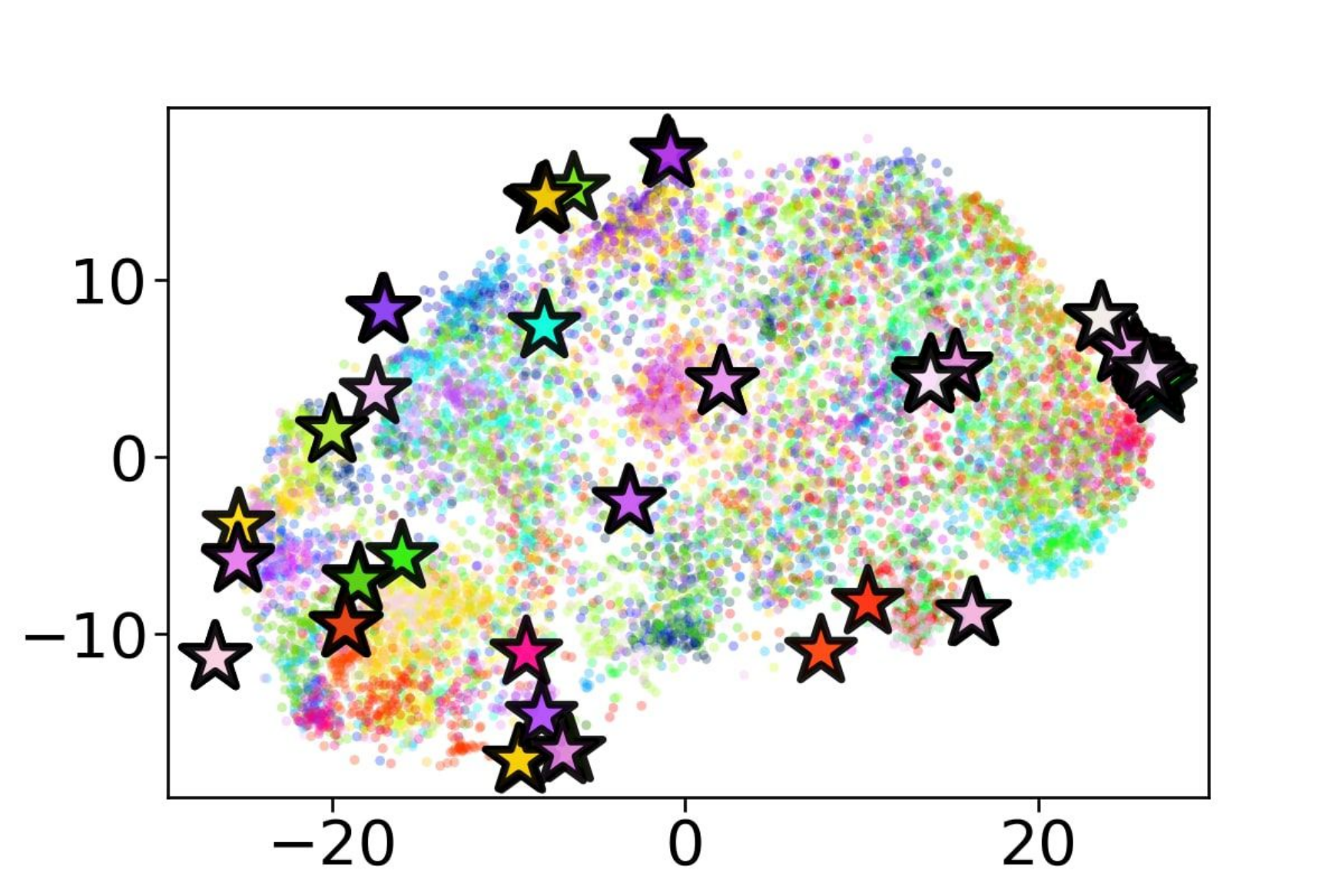} 
  & \includegraphics[width=0.2\columnwidth]{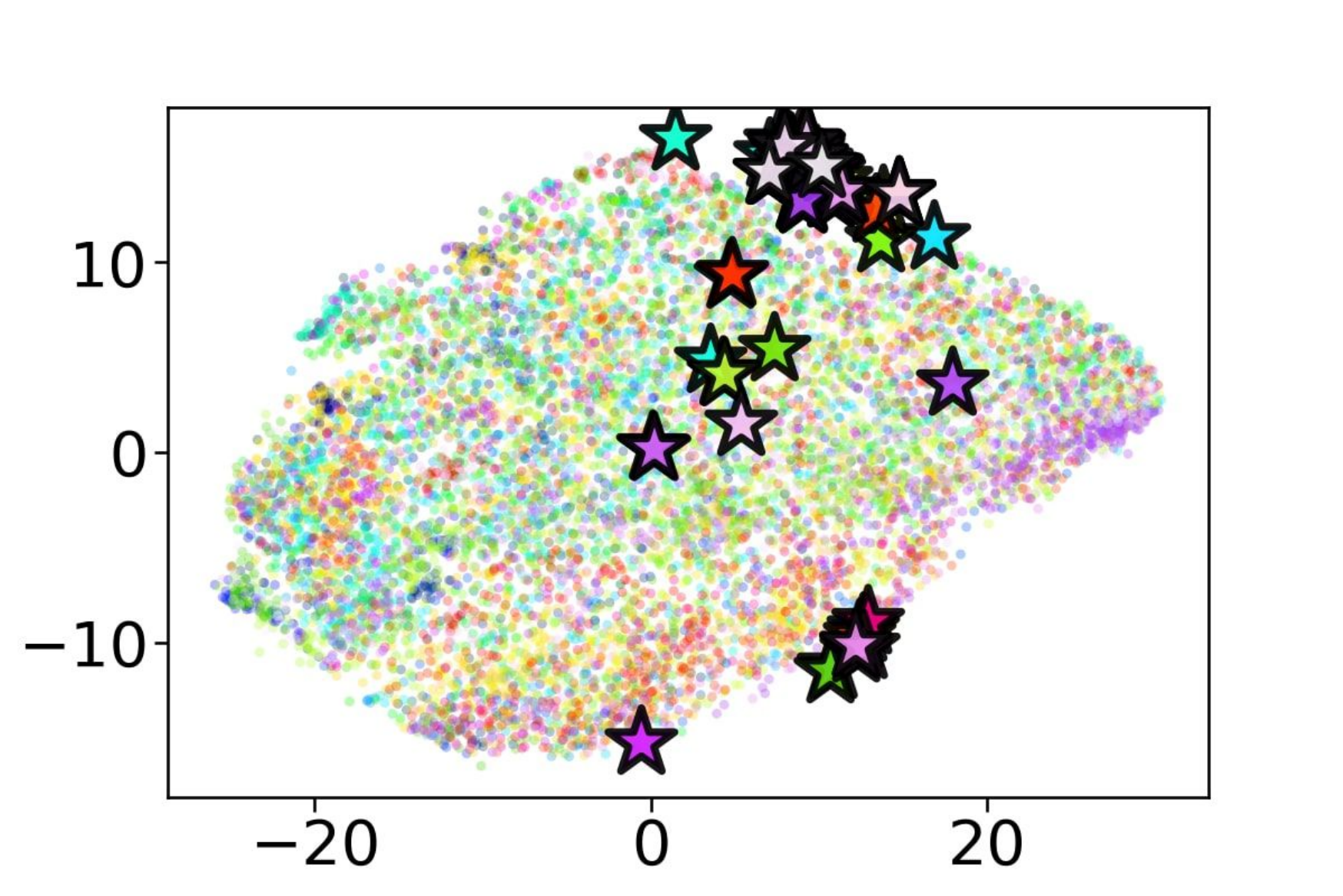}
  & \includegraphics[width=0.2\columnwidth]{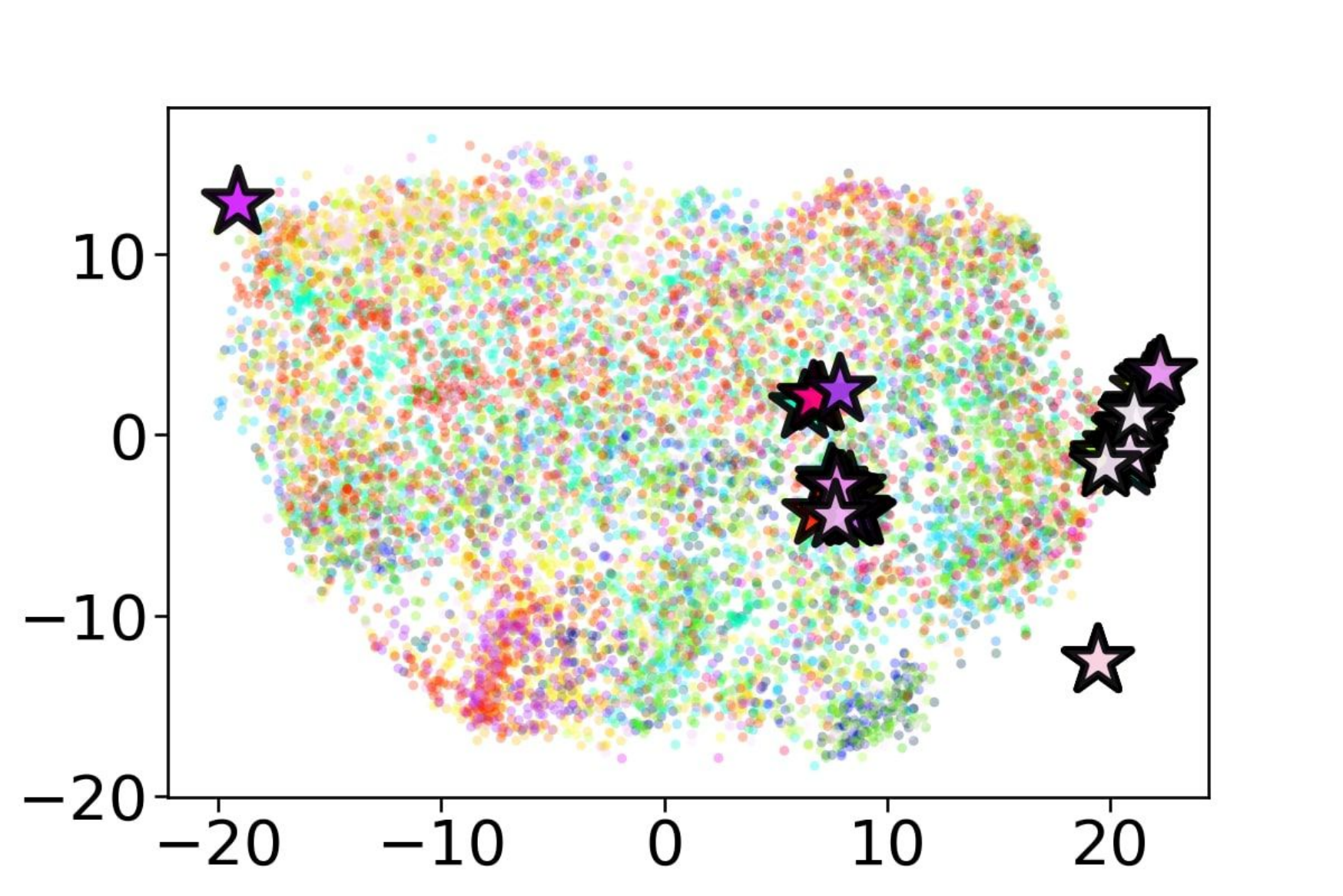}
  & \includegraphics[width=0.2\columnwidth]{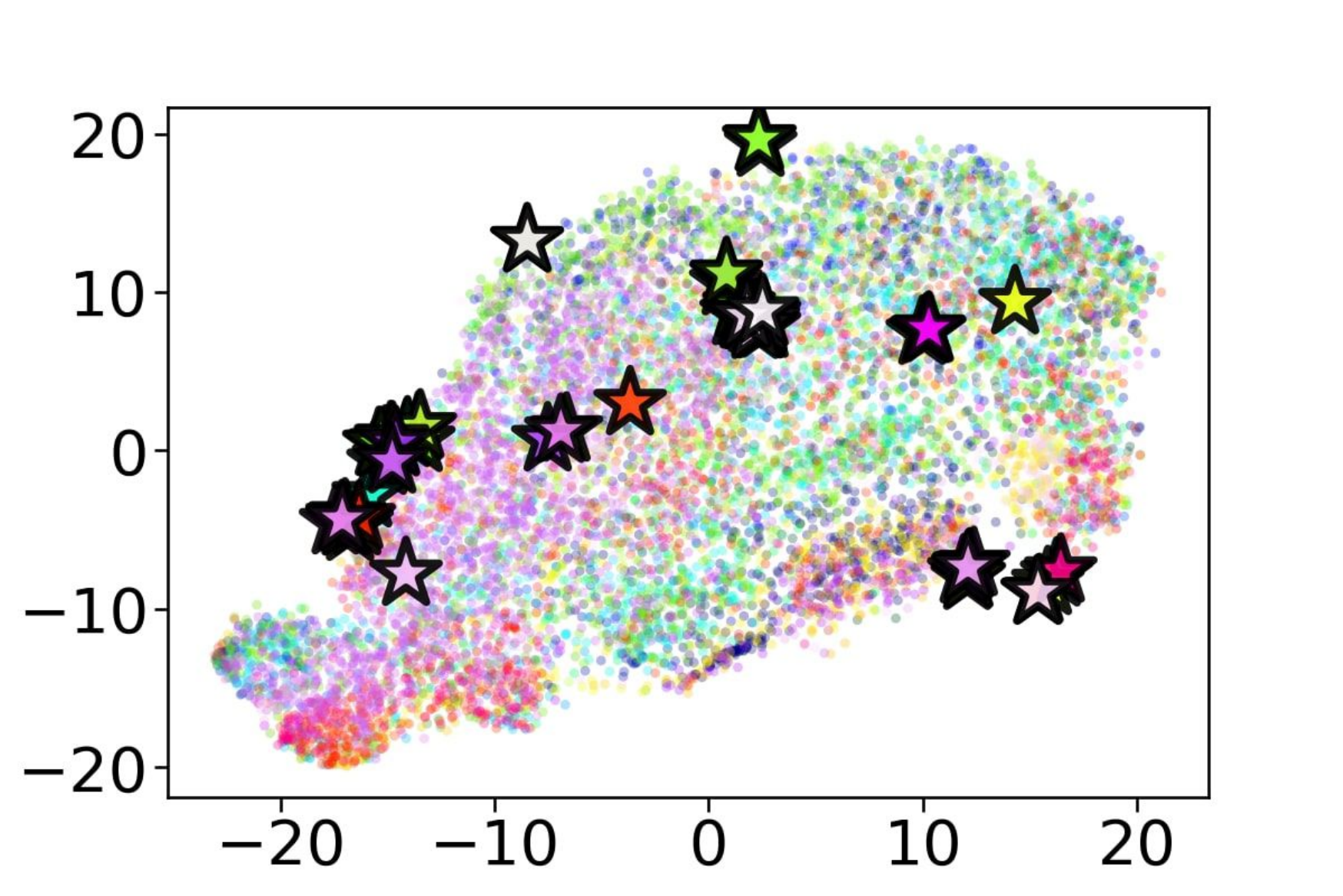}
  & \includegraphics[width=0.2\columnwidth]{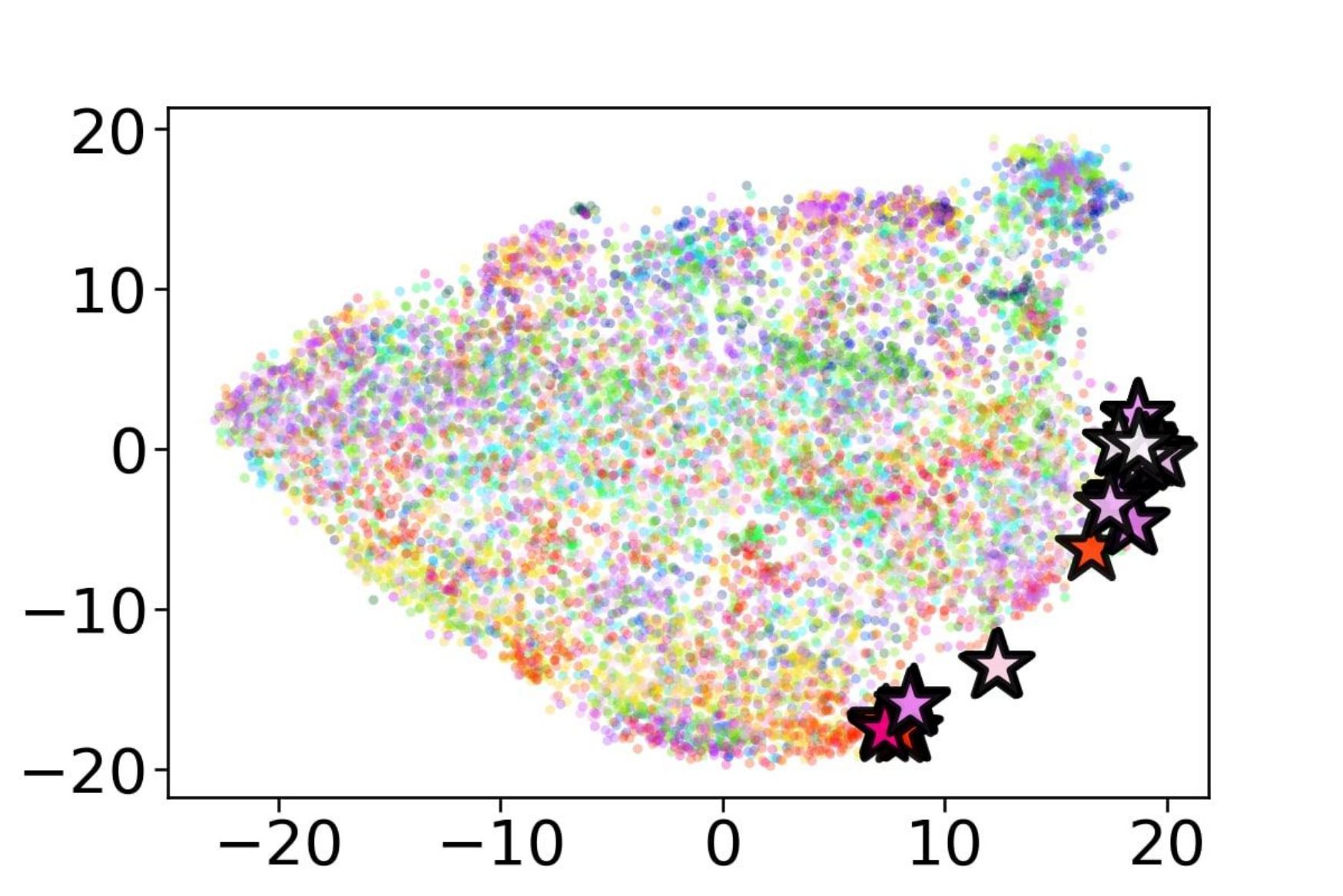}\\
  
\end{tabular}
\caption{Zero-shot ViT-B/16: The t-SNE plots show visual (\(\circ\)) and text ($\bigstar$) features for CIFAR-100C. }
\label{fig: tsne_cifar100_extra_source}
\end{figure}

In this section, we provide illustrations of task-wise t-SNE \cite{van2008visualizing} plots for CIFAR-10C and CIFAR-100C and compare them against zero-shot ViT-B/16. The results are in Figures \ref{fig: tsne_cifar10_extra}, \ref{fig: tsne_cifar10_extra_source}, \ref{fig: tsne_cifar100_extra}, and \ref{fig: tsne_cifar100_extra_source}. Across all corruptions/tasks, \framework\ learns strong discriminative visual features with a strong image-text alignment and class-level separation. ImageNet-C has 1000 classes, so, we do not provide t-SNE plots to avoid complications. However, the analysis and results carry forward.

\newpage

\end{document}